\documentclass[a4paper,fleqn]{cas-dc}

\usepackage[authoryear, sort]{natbib}
\usepackage{textcomp}
\usepackage{multicol,multicap,graphicx}
\usepackage{subfig}
\usepackage{lineno}
\usepackage{arydshln}
\usepackage{algorithm}
\usepackage[noend]{algpseudocode}
\usepackage{color}
\usepackage{xcolor}

\def\tsc#1{\csdef{#1}{\textsc{\lowercase{#1}}\xspace}}
\tsc{WGM}
\tsc{QE}
\begin{document}
\let\WriteBookmarks\relax
\def\floatpagepagefraction{1}
\def\textpagefraction{.001}

% Short title
%\shorttitle{<short title of the paper for running head>}    

% Short author
%\shortauthors{<short author list for running head>}  

% Main title of the paper
\title [mode = title]{\textcolor{black}{A Multi-modal  Garden Dataset and Hybrid 3D Dense Reconstruction Framework Based on Panoramic Stereo Images \textcolor{black}{for a Trimming Robot}}}  

% Title footnote mark
% e.g. \tnotemark[1]
% \tnotemark[<tnote number>] 

% Title footnote 1.
% e.g. \tnotetext[1]{Title footnote text}
%\tnotetext[<tnote number>]{<tnote text>} 

% First author
%
% Options: Use if required
% e.g. \author[1,3]{Author Name}[type=editor,
%       style=chinese,
%       auid=000,
%       bioid=1,
%       prefix=Sir,
%       orcid=0000-0000-0000-0000,
%       facebook=<facebook id>,
%       twitter=<twitter id>,
%       linkedin=<linkedin id>,
%       gplus=<gplus id>]

\author[1,2]{Can Pu}[style=chinese]

% Corresponding author indication
% \cormark[<corr mark no>]

% Footnote of the first author
%\fnmark[<footnote mark no>]

% Email id of the first author
\ead{can.pu@amigaga.com}

% URL of the first author
%\ead[url]{www.amigaga.com}

% Credit authorship
% e.g. \credit{Conceptualization of this study, Methodology, Software}
\credit{conceptualization, methodology, experiment,  writing - original draft, funding acquisition}

% Address/affiliation
\affiliation[1]{organization={Shenzhen Amigaga Technology Co. Ltd.},
            addressline={Building B, U+ Research Center, Gushu 1st Road, Baoan District}, 
            city={Shenzhen},
          citysep={}, % Uncomment if no comma needed between city and postcode
            postcode={518000}, 
%            state={Guangdong},
            country={China}}

\author[1]{Chuanyu Yang}[style=chinese]

% Corresponding author indication
\cormark[1]

% Email id of the second author
%\ead{chuanyu.yang@amigaga.com}

% Credit authorship
\credit{methodology, experiment, review \& editing, funding acquisition}

% Address/affiliation
\affiliation[2]{organization={School of Informatics, University of Edinburgh},
	addressline={1.11 Bayes Centre, 47 Potterrow}, 
	city={Edinburgh},
	citysep={}, % Uncomment if no comma needed between city and postcode
	postcode={EH89BT}, 
	%            state={},
	country={UK}}

\author[1,3]{Jinnian Pu}[]

% Email id of the third author
\ead{jipu0216@uni.sydney.edu.au}

% Credit authorship
\credit{Experiment data analysis and visualization}

% Address/affiliation
\affiliation[3]{organization={The University of Sydney},
	addressline={City Road}, 
	city={Camperdown/Darlington},
	citysep={}, % Uncomment if no comma needed between city and postcode
	postcode={NSW 2006}, 
	%            state={},
	country={Australia}}

\author[2]{Radim Tylecek}[]

% Email id of the third author
\ead{radim.tylecek@gmail.com}

% Credit authorship
\credit{Experiment data analysis and visualization (work done while at University of Edinburgh), review \& editing}

% comment it. Radim did this work in University of Edinburgh rather than his current company

% Address/affiliation
%\affiliation[3]{organization={Boundary Technologies Ltd},
%addressline={}, 
%	city={Edinburgh},
%	citysep={}, % Uncomment if no comma needed between city and postcode
%postcode={EH93FB}, 
%            state={},
%	country={UK}}

\author[2]{Robert B. Fisher}[]

% Email id of the second author
\ead{rbf@inf.ed.ac.uk}

% Credit authorship
\credit{helped with design of theory and experiments, review \& editing, funding acquisition, supervision}

% Corresponding author text
\cortext[1]{ Corresponding author at: Shenzhen Amigaga Technology Co. Ltd. \newline Email address: chuanyu.yang@amigaga.com (C. Yang)}

% Here goes the abstract
\begin{abstract}[SUMMARY]
Recovering an outdoor environment's \textcolor{black}{surface mesh} is vital for an agricultural robot during task planning and remote visualization. 
Image-based dense 3D reconstruction is sensitive to large movements between adjacent frames and the quality of the estimated depth maps.
\textcolor{black}{Our proposed solution for these problems is based on a newly-designed panoramic stereo camera along with a hybrid novel software framework that consists of three fusion modules: \textcolor{black}{disparity} fusion, pose fusion, and volumetric fusion. The panoramic stereo camera with a pentagon shape consists of 5 stereo vision camera pairs to stream synchronized panoramic stereo images for the following three fusion modules.} In the \textcolor{black}{disparity} fusion module, rectified stereo images produce the initial \textcolor{black}{disparity} maps using multiple stereo vision algorithms. Then, these initial \textcolor{black}{disparity} maps, along with the intensity images, are input into a \textcolor{black}{disparity} fusion network to produce refined \textcolor{black}{disparity} maps. Next, the refined \textcolor{black}{disparity} maps are converted into full-view ($360^{\circ}$) point clouds or single-view ($72^{\circ}$) point clouds for the pose fusion module. The pose fusion module adopts a two-stage global-coarse-to-local-fine strategy. In the first stage, each pair of full-view  point clouds is registered by a global point cloud matching algorithm to estimate the transformation for a global pose graph's edge, \textcolor{black}{which effectively implements loop closure}. In the second stage, a local point cloud matching algorithm is used to match single-view point clouds in different nodes. Next, we locally refine the poses of all corresponding edges in the global pose graph using three proposed rules, thus constructing a refined pose graph. The refined pose graph is optimized to produce a global pose trajectory for volumetric fusion. In the volumetric fusion module, the global poses of all the nodes are used to integrate the single-view point clouds into the volume to produce the \textcolor{black}{mesh} of the whole garden. 
The proposed framework and its three fusion modules are tested on a real outdoor garden dataset to show the superiority of the performance. The whole pipeline takes about 4 minutes on a desktop computer to process the real garden dataset, which is available at: \url{https://github.com/Canpu999/Trimbot-Wageningen-SLAM-Dataset}.    
\end{abstract}

% Keywords
% Each keyword is seperated by \sep

\maketitle

% Main text
\section{Introduction}\label{introduction}

\begin{figure*}
	
	\begin{multicols}{2}
		\begin{center}
			\flushleft
			\subfloat[The TrimBot2020 test garden]{\includegraphics[width=1.4\linewidth]{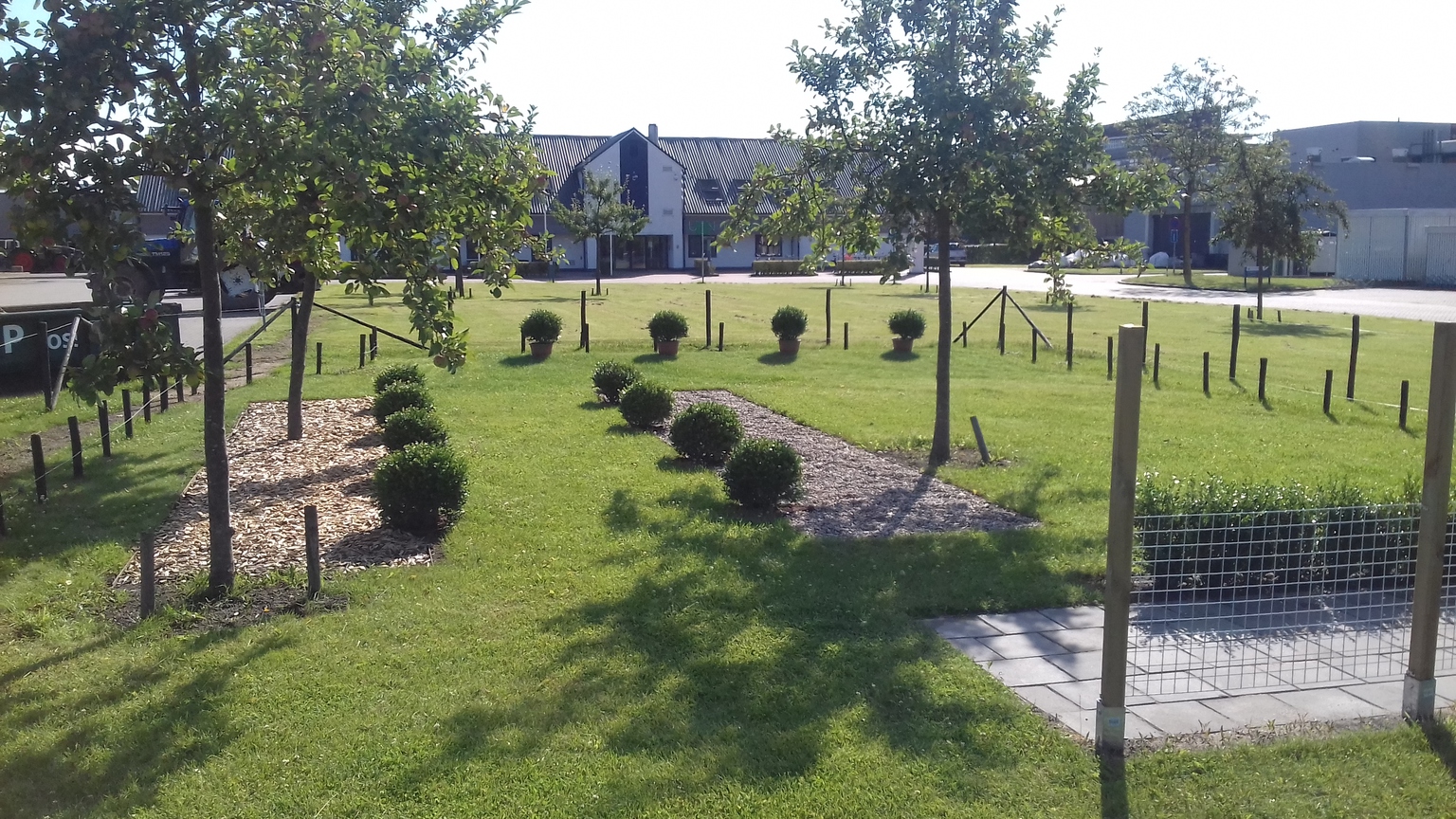}}
		\end{center}
		
		\begin{center}
			\flushright
			\subfloat[\textcolor{black}{A panoramic stereo camera}]{\includegraphics[height=7cm]{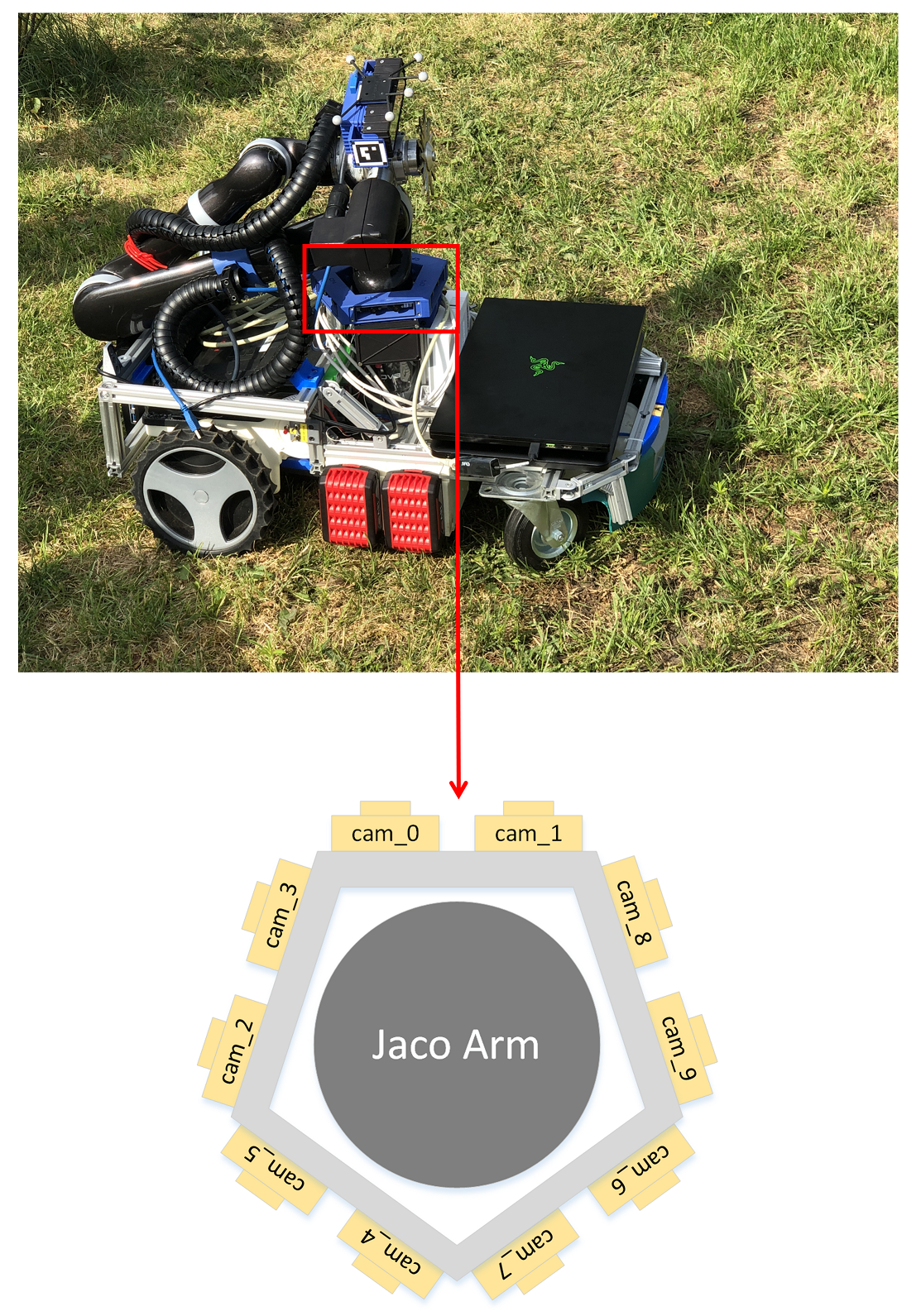}}
		\end{center}
	\end{multicols}
	
	\begin{center}
		\subfloat[\textcolor{black}{10 synchronized images from the panoramic stereo camera consisting of 5 stereo vision cameras (10 image sensors in total)}]{\includegraphics[ width=\linewidth]{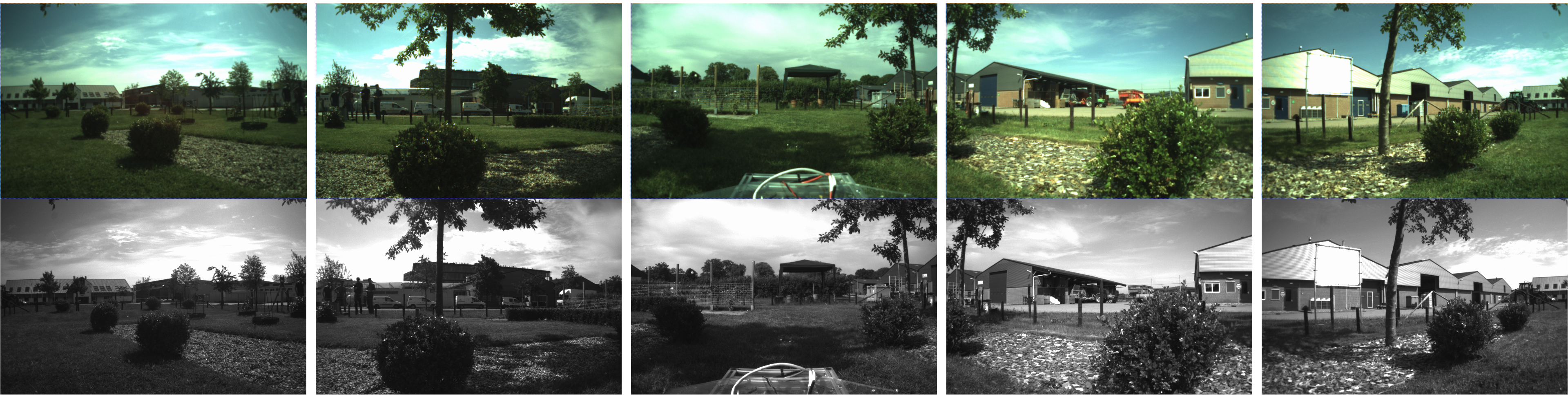}}
	\end{center}
	
	\caption{\textcolor{black}{Figure (a) shows the TrimBot2020 outdoor test garden; Figure (b) shows the newly-designed `panoramic stereo camera' hardware, which is mounted on the trimming robot. The panoramic stereo camera with a pentagon shape consists of 5 stereo vision cameras (10 image sensors, `Cam0' - `Cam9'). Figure (c) shows ten synchronized raw images from the 5 stereo camera pairs (the left image sensors in the stereo configuration are color cameras and the right image sensors are gray scale). The images in this figure are from the Trimbot2020 project. \textcolor{black}{For viewing small details in this figure, readers are recommended to view the electronic version.}}}
	\label{fig:garden-with-robot}
\end{figure*}

An economical but robust online 3D reconstruction approach for outdoor environments is vital for the remote visualization \textcolor{black}{of the scene} and robot task planning. Recovering the dense 3D structure (e.g. mesh) of an outdoor garden with only image input quickly and robustly is challenging because of lighting changes, texture similarity, shadow interference, limited computation and network resources, etc. Figure~\ref{fig:garden-with-robot}~(a) shows a real outdoor garden for our gardening robot Trimbot's\footnote{Trimbot2020 project URL: \url{http://trimbot2020.webhosting.rug.nl/}} navigation and plant pruning. \textcolor{black}{In real applications, there are two big challenges\footnote{\textcolor{black}{For more description about the challenges in the real world, please read the following file: \url{https://github.com/Canpu999/Trimbot-Wageningen-SLAM-Dataset/blob/main/Real-challenges.pdf}}} for image-based dense 3D reconstruction with high fidelity: 1) Movement (rotation or translation) between adjacent frames is big because of e.g. a gardening robot's fast speed (1~$m/s$ translation or 90~$deg/s$ rotation), the temporal downsampling ratio of the frames\footnote{Because of the mobile network speed, we pick one out of every ten frames to transfer to the server for online 3D reconstruction.} (1/10), and the image sensors' low frame rate (12 FPS); 2) \textcolor{black}{Disparity} maps\footnote{In a stereo configuration, disparity and depth are interchangeable measures: $depth = focal \_ length \times baseline
	/ disparity$. When input data is from a depth sensor like Lidar or time of flight sensor, the depth information can be converted into disparity information by using a constant baseline and focal length. Thus, in this paper we regard the two terms as the same and won't distinguish them.} from existing methods  in real outdoor environments are not accurate, dense and robust enough because of texture similarity, lighting changes, and shadows.}

\textcolor{black}{
According to a recent survey~\citep{Survey1-Visual-SLAM}, SLAM\footnote{\textcolor{black}{The abbreviations in this manuscript are listed in Appendix~\ref{Abbreviations} List of Abbreviations. And the frequently-used symbols in this manuscript are listed in Appendix~\ref{symbols} List of Symbols.}} systems are classified by the input data source or the sensors used. Figure~\ref{fig:slam-classification-related-work} shows the SLAM classification results. Given that the input data to our proposed framework is from a newly-designed panoramic stereo camera (see Figure~\ref{fig:garden-with-robot}~(b), Figure~\ref{fig:sensor-set-appendix} or Figure~\ref{fig:panoramic-stereo-cameras-appendix}) and there is no existing similar work as far as we know, we have defined a new branch in the pure visual SLAM class called `Panoramic Stereo SLAM' with abbreviation `PS-SLAM'. \textbf{Panoramic stereo SLAM is a class of pure visual SLAM methods with panoramic stereo images as input  (e.g. our ring of stereo vision cameras to achieve $360^\circ$ perception)}. Our proposed framework belongs to PS-SLAM which is in the pure visual SLAM area. }

\textcolor{black}{
Instead of following the existing main-stream pure visual SLAM technique  [pose recovery by feature extraction and mapping (e.g.~\cite{orbslam3,SFM}); pose recovery by minimizing the pixel-wise photo-metric error (e.g.~\cite{DSO})], we start a new approach (slightly similar to the RGB-D SLAM algorithm Kinectfusion~\citep{Kinect-fusion}) to do pose recovery by using the point clouds rather than the features or pixel intensities (which are sensitive to illumination, scene appearance, and shadows). To guarantee point cloud quality outdoors, a new disparity fusion algorithm is first introduced into the SLAM pipeline, whose outputs are then improved by some practical techniques. To deal with fast motion or rotation of the robot, an innovative multi-stage pose trajectory estimation method with joint information (Algorithm~\ref{fig:algorithm-MPTEJI}) is developed based on loop closure (LC), view switching, and global \& local information transition. The integration of multi-level fusion modules, various supporting algorithms and different innovative strategies make the proposed hybrid framework unique and able to cope with the real challenges mentioned above, on which the traditional SLAM frameworks (e.g. Orbslam3~\citep{orbslam3}, Open3D reconstruction system~\citep{Open3d}, and the commercial software `ContextCapture') perform badly.     
}

More specifically, to solve the two big challenges above in a real garden for the trimming robot, \textcolor{black}{the TrimBot2020 project team designed a new hardware configuration called the `panoramic stereo camera' along with a novel 3D reconstruction software framework containing three fusion modules to compute accurate \textcolor{black}{disparity} maps, estimate relative pose, and geometrically integrate the maps. Figure~\ref{fig:garden-with-robot} (b) shows the panoramic stereo camera which is mounted on the TrimBot2020 robot, and which is primarily used for navigation and visual servoing when the vehicle is near to plants to be trimmed. The diagram in Figure~\ref{fig:garden-with-robot} (b) shows the panoramic stereo camera with 5 stereo vision cameras (10 image sensors `Cam0' - `Cam9') arranged in a pentagon shape. 
The panoramic stereo camera streams the synchronized panoramic stereo images (see Figure~\ref{fig:garden-with-robot} (c)) from the 10 image sensors (`Cam0' - `Cam9') for the following three modules to deal with.} 
First, in the \textcolor{black}{disparity} fusion module, rectified stereo images are combined to compute the initial \textcolor{black}{disparity} maps by multiple stereo vision algorithms. Then the initial \textcolor{black}{disparity} maps along with the image information are input into a \textcolor{black}{disparity} fusion network to produce a refined \textcolor{black}{disparity} map. Next, the refined \textcolor{black}{disparity} map is converted into a full-view ($360^{\circ}$) point cloud or a single-view ($72^{\circ}$) point cloud for the pose fusion module (see Algorithm~\ref{fig:algorithm-MPTEJI}). In the first stage of pose fusion, each two $360^{\circ}$ local point clouds are registered by a global point cloud matching algorithm to get the corresponding transformation for the global pose graph's edge, \textcolor{black}{which realizes loop closure (LC) essentially}. The global pose graph is then optimized to produce a coarse global pose trajectory of the robot's path through the garden. In the second stage, a refined pose graph is computed based on the coarse global pose graph. Local point cloud registration \textcolor{black}{along with the coarse global pose trajectory from the first stage jointly update}  all the available edges in the refined pose graph, which is then optimized to estimate an accurate global sensor pose trajectory. Lastly, in the volumetric fusion module, the global poses of all the nodes in the refined pose graph are used to integrate the corresponding depth maps or point clouds into a volume to produce \textcolor{black}{the surface mesh} of the whole garden, which could be used for task planning and the remote visualization.  Figure~\ref{fig:visual-pipeline} gives an  overview of the whole fusion pipeline.

\textcolor{black}{
In conclusion, there are three major contributions which could be regarded as the foundation of the PS-SLAM 
approach. The \textbf{major contributions} are :
}  

\textcolor{black}{
(1)	First real garden dataset (Figure~\ref{fig:garden-dataset}) for future PS-SLAM research, which contains the ground truth of the fully-dense depth maps, the semantic maps, the global poses, the rectified stereo images, sparse Lidar scans, and the semantic 3D model;}

\textcolor{black}{
(2) First hybrid 3D dense reconstruction framework based on panoramic stereo images, which could be regarded as the initial baseline framework (Figure~\ref{fig:visual-pipeline}) for future PS-SLAM research; }

\textcolor{black}{
(3) First two-stage full-view-to-single-view global-coarse-to-local-fine pose trajectory estimation method (Algorithm~\ref{fig:algorithm-MPTEJI}), which is robust to fast or large transformations between adjacent frames}.

\textcolor{black}{
Additionally, there are three notable minor contributions to solve the related problems or improve the related performance in this paper: }

\textcolor{black}{
	1) Theoretical proof (Appendix~\ref{Appendix-Formular-derivation-loss-function}) that the Frobenius-norm-based transformation difference loss function (Equation~\eqref{eq: graph-optimization-global}) is a special case of the maximum likelihood loss function when applied to the pose graph optimization problem;}

\textcolor{black}{
	2) Two practical strategies (in Section~\ref{section:depth-fusion}) with the theoretical proof (Appendix~\ref{Appendix-Formular-derivation-maximum-distance}) to improve the \textcolor{black}{disparity} fusion accuracy by setting the maximum distance of interest (denoted by `Maximum Distance') and up-and-down resolution transformation (denoted by `High Definition');  }

\textcolor{black}{
	3) Three rules (in Section~\ref{section:refined-global-graph-reconstruction}) to optimize the edge set which constrains the pose graph's loss function, boosting the estimated pose's accuracy}.

The remainder of this paper is structured as follows. Section~\ref{related-works} presents previous research about SLAM classification, influence factors, and the dataset. Section~\ref{methodology} presents the proposed multi-level fusion framework including the \textcolor{black}{disparity} fusion module, the pose fusion module, and the volumetric fusion module. Section~\ref{Experiment} describes the real garden dataset and demonstrates the performance of the fusion framework including the \textcolor{black}{disparity} fusion module, the pose fusion module and the volumetric fusion module on the real garden dataset. Section~\ref{Conclusion} presents a discussion and summary of the work.

\section{Related Works}\label{related-works}
\textcolor{black}{
This section reviews the existing SLAM classification and  positioning the proposed new SLAM framework within it. Secondly, we analyse factors which influence the framework's performance. Lastly, the outdoor datasets used for visual SLAM are reviewed. 
}

\begin{figure*}
	\centering
	\includegraphics[width=0.95\linewidth]{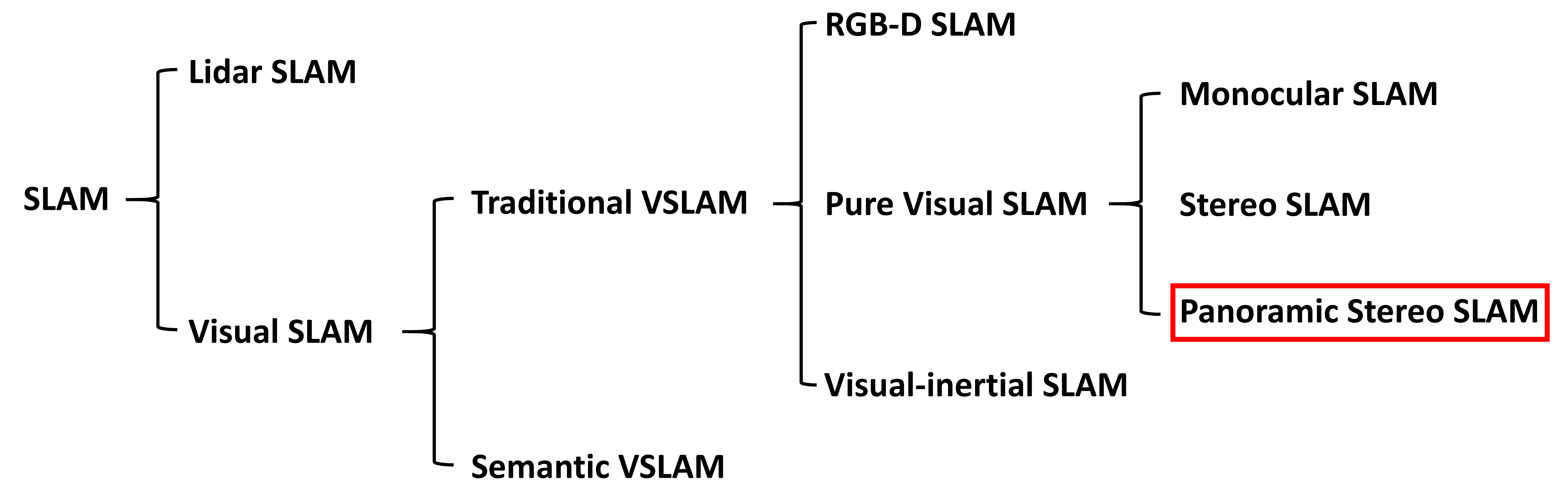}
	\caption{\textcolor{black}{The figure shows SLAM classification~\citep{Survey1-Visual-SLAM} based on the input data source or the used sensors. As far as we know, our proposed framework is the first work in the branch `Panoramic Stereo SLAM'.}}
	\label{fig:slam-classification-related-work}
\end{figure*}

\textcolor{black}{	
\subsection{SLAM Classification}  \label{section:slam-classification-related-work}
According to a recent survey~\citep{Survey1-Visual-SLAM}, SLAM systems are classified by the input data source or the sensors used. Figure~\ref{fig:slam-classification-related-work} shows the classification of different SLAM systems. 
SLAM systems have been divided into two main categories: Lidar SLAM (e.g. Lego-Loam~\citep{Lego-Loam}) and visual SLAM. Within the visual SLAM category, there are two sub-categories: semantic visual SLAM (e.g. Blitz-slam~\citep{Blitz-slam}) and traditional visual SLAM. RGB-D SLAM (e.g. Kinectfusion~\citep{Kinect-fusion}, Elasticfusion~\citep{Elasticfusion}), pure visual SLAM and visual-inertial SLAM (e.g. Vins-mono~\citep{Vins-mono}) constitute the traditional visual SLAM family. Monocular SLAM with a single image sensor (e.g. Colmap~\citep{SFM}), stereo SLAM with a stereo vision camera (e.g. Stereo LSD-SLAM~\citep{Stereo-LSD-SLAM}), and our proposed framework with a panoramic stereo camera belong to the pure visual SLAM class.       
}

\textcolor{black}{
While the newly-designed panoramic stereo camera (a ring of 5 synchronized stereo vision cameras mounted in a pentagon shape) extends the category of stereo SLAM, there are significant differences which stem from the specifics of the panoramic arrangement of multiple cameras. Thus, we define a new traditional pure visual SLAM branch `Panoramic Stereo SLAM' (PS-SLAM), which uses a panoramic stereo camera - a ring of synchronized stereo vision cameras - to input a $360^\circ$ view. 
Although there is a stereo panoramic vision system~\citep{PS-SLAM-fake1} that uses a stereo vision camera containing two panoramic vision sensors with a wide field of view (FOV) , that approach still largely follows the classic stereo SLAM concept with some improvements to the stereo SLAM framework. As far as we know, our proposed framework (Figure~\ref{fig:visual-pipeline}) is the first true PS-SLAM research.
}

\textcolor{black}{
Compared with mainstream panoramic SLAM algorithms~\citep{PanoramicSLAM7,PanoramicSLAM6,PanoramicSLAM5,PanoramicSLAM4,PanoramicSLAM3,PanoramicSLAM2,PanoramicSLAM1}, one difference between ours and theirs is that they 
cannot provide dense global depth information because there is only one monocular camera at each viewpoint inside their panoramic camera whereas in our case multiple stereo images from different perspectives create the panoramic image. 
The second difference is that they still follow the typical visual SLAM pipeline (e.g. SFM~\citep{SFM}, Orb-slam~\citep{Orbslam1}): feature extraction and mapping, pose estimation by triangulation, loop closure detection, and global optimization, which makes them sensitive to lighting changes unlike our proposed method. Additionally, these algorithms only produce a sparse reconstruction of the scene based on matched feature points as compared to our dense reconstruction.}

\textcolor{black}{Meanwhile, some researchers~\citep{RPV-SLAM,HDPV-SLAM} projected the point clouds from a Lidar scanner to the image plane of a panoramic image from a panoramic camera to form a panoramic RGB-D image first. Then, the panoramic RGB-D images were input into OpenVSLAM~\citep{OpenVSLAM}, an open-source third-party library containing commonly used visual SLAM algorithms (e.g. RGB-D SLAM algorithm in Orb-slam framework~\citep{orbslam2, orbslam3}). Compared with our solution, their Lidar sensor is expensive and does not produce a dense point cloud. In  summary, the existing panoramic SLAM algorithms follow the traditional visual SLAM pipeline and ours has a different theoretical framework. Based on our newly-designed panoramic stereo camera, we  provide an economic dense reconstruction solution which is robust to light 
changes and scene appearance. }

\textcolor{black}{      
Different from previous frameworks~\citep{Survey-visual-slam,Survey1-Visual-SLAM,Survey-rgbd-slam,Survey-Lidar-SLAM}, the proposed framework is designed specifically for a panoramic stereo camera set and concentrates on depth quality improvement under some challenging conditions (illumination changes, similar texture, etc.) and accurate global pose trajectory estimation while coping with the robot's fast motion or rotation. Thus, some new features are proposed to enhance the SLAM framework, such as the disparity fusion module, the field of view switching ($360^\circ$ versus $72^\circ$), the novel multi-stage global pose trajectory estimation algorithm~\ref{fig:algorithm-MPTEJI}. 
}

\textcolor{black}{
To summarize, the proposed framework is the first to do 3D dense reconstruction in the PS-SLAM research subfield, and so is a baseline to facilitate the progress of PS-SLAM. Compared with the popular SLAM frameworks (e.g. OrbSLAM3~\citep{orbslam3}, Open3D reconstruction system~\citep{Open3d}, and the commercial software `ContextCapture'), the integration of multi-level fusion modules, various supporting algorithms, and different innovative strategies makes the proposed framework both unique and capable of performing well even given the two real challenges facing any real outdoor robot: depth data quality and fast robot motion.  
}

\textcolor{black}{
\subsection{Performance Factors}  \label{section:influence-factors-related-work}
}

\textcolor{black}{
\subsubsection{Depth Quality}
}

Compared with Lidar scanners (expensive and their point cloud is sparse) and ToF sensors (sensitive to infrared light outdoors), etc., image-based depth estimation methods (e.g. stereo vision algorithms) are economical and produce dense depth map indoors and outdoors robustly. \textcolor{black}{In our proposed framework, the stereo vision algorithms estimate the raw disparity maps and the disparity fusion algorithm is used to refine the raw disparity maps from the stereo vision algorithms to get a refined disparity map.} 

The most well-known classical stereo vision algorithm is the semi-global matching method~\citep{SGM}, which conducts pixel-wise matching using mutual information with a global smoothness approximation. With the rise of deep neural networks, Flownet~\citep{Flownet} is the first to use an end-to-end convolutional neural network to estimate the disparity map between two images. A recent survey~\citep{Survey_stereo-vision} gives an overview of the latest progress of stereo vision algorithms. \textcolor{black}{Although stereo vision algorithms have made huge progress recently, a single stereo vision algorithm still has different advantages and disadvantages, and fails to estimate disparity maps accurately at all pixels in all scenes.} \textcolor{black}{Disparity} fusion is a good method for refining the initial raw disparity maps (from the same viewpoint) from several individual \textcolor{black}{disparity} estimation algorithms to estimate a more accurate and robust \textcolor{black}{disparity} map based on their complementary properties. The majority of classical \textcolor{black}{disparity} fusion methods~\citep{Reliable-stereo-tof-fusion,Confidence-stereo-tof-fusion,Guided-stereo-tof-fusion} share the same pipeline: estimate the \textcolor{black}{disparity} map and a confidence map from different sensors, and then use a specific fusion method to fuse the \textcolor{black}{disparity} maps using the confidence maps as weights. Because it is hard to estimate the confidence map and \textcolor{black}{disparity} distribution accurately, these classical methods have a lower precision compared with deep-learning-based methods~\citep{sdf-man, Multi-sensor-fusion, UDFNet}.
To highlight, Sdf-man~\citep{sdf-man} is the first to input multiple initial disparity maps with auxiliary information (e.g. RGB, gradients) into the refiner network to produce a refined \textcolor{black}{disparity} map.
It used a discriminator to classify the refined \textcolor{black}{disparity} map and the ground truth \textcolor{black}{disparity} map as real or fake to improve the refined \textcolor{black}{disparity} map's accuracy. 

\textcolor{black}{In existing SLAM system surveys~\citep{Survey-visual-slam,Survey1-Visual-SLAM,Survey-rgbd-slam,Survey-Lidar-SLAM}, the emerging concept of `disparity fusion' was not mentioned and 
we saw no mention of using a disparity fusion algorithm in the SLAM system to improve the 3D dense reconstruction accuracy. 
We are the first to encode the disparity fusion part in the front end of our proposed SLAM system based on Sdf-man~\citep{sdf-man}. Although Sdf-man achieved state-of-art real-time performance in an outdoor garden, its error rate still lies at \textasciitilde10~$cm$ level. In this paper, we propose two new practical strategies (Section~\ref{section:depth-fusion}) along with a proof (Appendix~\ref{Appendix-Formular-derivation-maximum-distance}) to improve the disparity fusion accuracy, as demonstrated by experiments in Section~\ref{exp: depthfusion}. }

\subsubsection{Pose Accuracy}
\textcolor{black}{Compared with image feature matching (e.g. SIFT~\citep{SIFT}, ORB~\citep{ORB}) to estimate the 6D pose between different views, estimation based on point clouds can produce a more reliable and accurate result.}  
Currently, there are three classes of point cloud matching algorithms to estimate the relative 6D pose. A recent survey~\citep{survey-point-cloud-matching-2021} has an overview.
The first class of algorithms (e.g. ~\cite{GICP, ICP-1-2022, ICP-2-2022}) is derived from ICP~\citep{ICP1992}, which calculates the relative 6D pose between two point clouds by finding the closest corresponding points in two point clouds and minimizing their Euclidean distance. 
Exactly corresponding points seldom exist in the real cases, so the ICP-based methods have low accuracy (and initialization issues).
The second class is feature-based algorithms~\citep{3DMatch-2017, point-cloud-matching-descriptor-2021,point-cloud-matching-descriptor2-2021,Fast-global-registration}. They extract local descriptors from two point clouds first and then do feature matching to estimate the relative 6D pose between the two point clouds. This class is sensitive to noisy and sparse point clouds which may lead to inaccurate local descriptors and could even make the algorithm collapse when the density is too sparse or the noise is too strong. 
The third class~\citep{CPD, DUGMA, CPD2, GMM-point-cloud-matching} treats point cloud registration as a probability matching problem. They use probabilistic models to describe the geometric distribution of the two point clouds first and then maximize the likelihood of two probabilistic models overlap to calculate the relative 6D pose of the two point clouds.
This class of algorithms aligns point clouds more accurately and robustly compared with the previous two classes, but is slow because of their computational complexity. 

\textcolor{black}{
Pairwise point cloud registration algorithms only compute the relative 6D pose between two local segments within a whole pose trajectory, and do not guarantee estimation of the global optimum of the whole global pose trajectory. That is, 
}
registering point clouds sequentially produces a sensor pose trajectory which inherently drifts over time because of the accumulated error. Building an optimized pose graph~\citep{Tutorial-pose-graph, ICP-based-pose-graph, SFM-pose-graph} could reduce the accumulated error and give an optimized global solution. 
Ordinarily, loop closure helps resolve this issue, but here the 360 degree point clouds allow many overlapping point sets.
\textcolor{black}{
Based on the full-view and single-view point clouds, we are the first to develop a two-stage full-view-to-single-view global-coarse-to-local-fine pose trajectory estimation method (Algorithm~\ref{fig:algorithm-MPTEJI}), which can cope well with the fast motion of the real robot outdoors.	
}

\subsubsection{Volumetric Fusion}
With a range of depth maps and their corresponding global poses, volumetric fusion methods~\citep{Volumetric-fusion, Scalable-volumetric-fusion} integrate the surface geometry information into a volume that represents the 3D space in the world coordinate system. Using volumetric integration to build the 3D model of the garden and the marching cube technique~\citep{Marching-cube-new, Marching-cube} to extract the surface mesh and its corresponding point cloud is a good option to remove outliers and noise, which could result in good quality when reconstructing the 3D garden model. 
\textcolor{black}{We use an existing volumetric fusion technique (Section~\ref{section:volumetric-fusion}) for completeness and visualization purposes and do not claim a contribution for this part.}

\textcolor{black}{
\subsection{Dataset\label{section:dataset-related-work}}
There are multiple outdoor datasets for visual SLAM research, such as Kitti~\citep{Kitti1, Kitti2} and Cityscapes~\citep{Cityscape} for autonomous driving in the city. Besides our dataset, some other datasets (e.g. ~\cite{LettuceMOT,TobSet,Sugar-Beets-Dataset,BLT-dataset}) for agricultural robots have recently been announced. For example, LettuceMOT~\citep{LettuceMOT} and TobSet~\citep{TobSet} have only semantic information for lettuce, tobacco crop, weed detection and tracking. The Sugar Beets Dataset~\citep{Sugar-Beets-Dataset} contains the data from an RGB-D sensor (Kinect v2), a 4-channel multi-spectral camera (JAI AD-130GE), two on-board Lidar scanners (Velodyne VLP-16 Puck) and two GPS sensors (Leica RTK GPS and Ublox GPS) as well as wheel encoders to facilitate the research relevant to plant classification, localization and mapping in a sugar beet field. The BLT Dataset\footnote{BLT dataset: \url{https://lcas.lincoln.ac.uk/wp/research/data-sets-software/blt/}}~\citep{BLT-dataset} contains the data from two RGB-D sensors (ZED), an IMU (RSX-UM7), a 2D Lidar scanner (SICK MRS1000) and a 3D Lidar  scanner (Ouster OS1-16) for long-term mapping and localization in a vineyard.} 

\textcolor{black}{
Compared with all the previous datasets, the obvious difference in our dataset is the inclusion of fully dense ground truth depth maps, a fully dense ground truth semantic 3D model and the synchronized joint panoramic stereo information (including RGB \& intensity, fully-dense depth, semantic labels, sparse laser scan and global pose) for the first time. Our released dataset could facilitate multiple research topics in SLAM, including sensor calibration, depth estimation, semantic segmentation, pose estimation, and all types of SLAM frameworks (Lidar SLAM, semantic visual SLAM, and traditional visual SLAM). To the best of our knowledge, this is the first public dataset (Section~\ref{exp: dataset}) in the panoramic stereo SLAM domain.
}

\section{Methodology}\label{methodology}

\begin{figure*}
	\includegraphics[width=1\linewidth]{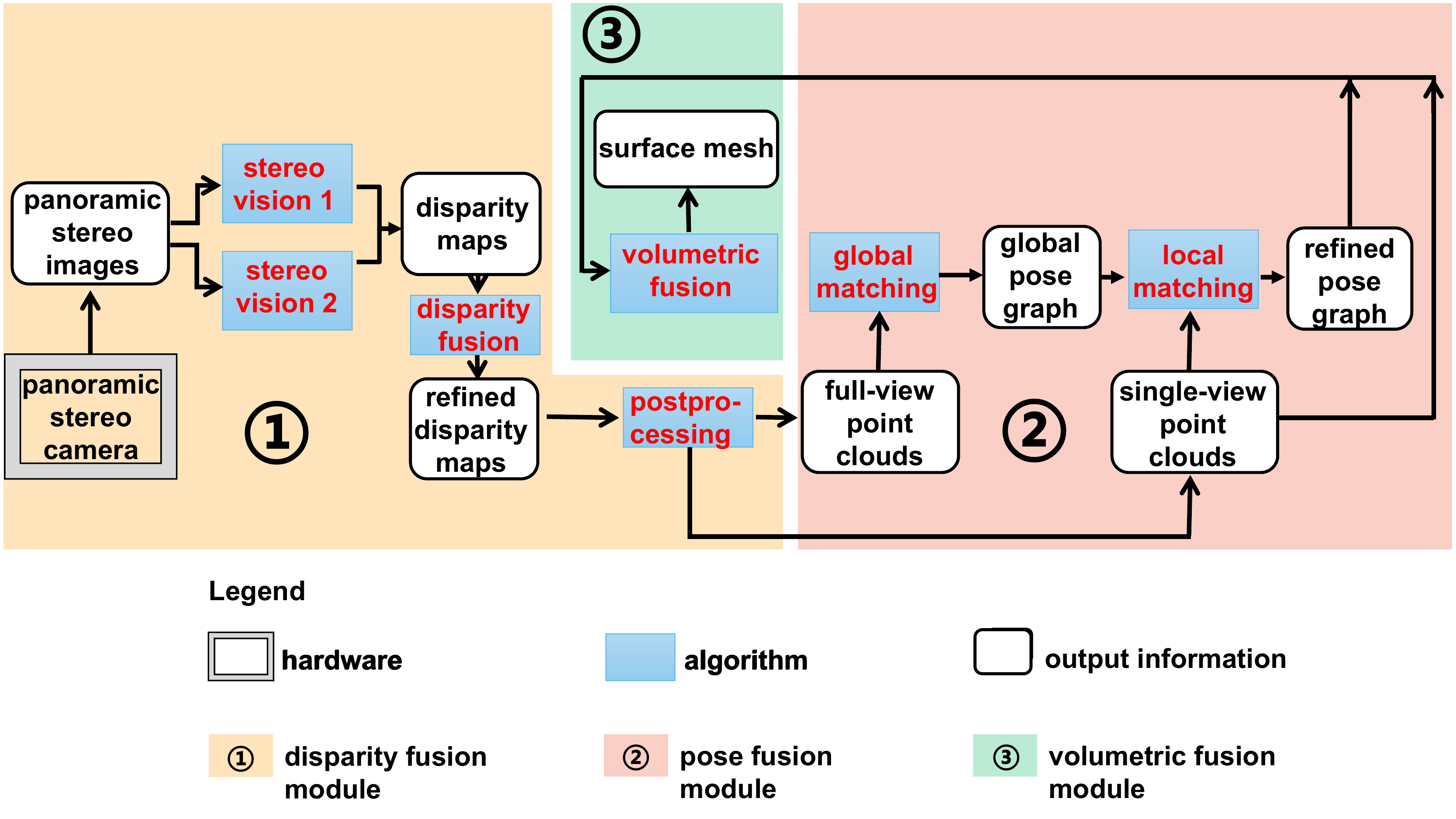}
	\caption{\textcolor{black}{The proposed hybrid multi-level fusion framework based on panoramic stereo images for 3D dense reconstruction}}
	\label{fig:visual-pipeline}
\end{figure*}

Figure~\ref{fig:visual-pipeline} shows \textcolor{black}{the proposed hybrid 3D dense reconstruction framework based on  images from a panoramic stereo camera rig}. 
 Five calibrated binocular cameras \textcolor{black}{inside the panoramic stereo camera (each with a FOV of $72^\circ$)} stream five pairs of rectified stereo images from the related left and right cameras synchronously into the \textcolor{black}{disparity} fusion module. The stereo image pairs are fed into two separate stereo vision algorithms to compute disparity maps in the same view. The disparity maps are used by the disparity fusion algorithm to produce the refined disparity map, which can be converted into the corresponding full-view (FOV = $360^{\circ}$) or single-view (FOV = $72^{\circ}$) point cloud. \textcolor{black}{The point clouds' outliers are removed} by post-processing. 
 
In the pose fusion module, the full-view point clouds from different frame times are first registered with each other by a global point cloud matching algorithm for a coarse global pose estimate, which becomes the corresponding edge's transformation in the global pose graph.
Then, the global pose graph is optimized to produce a coarse global sensor pose trajectory. 
In the second pose fusion stage, the refined pose graph is initialized by the global pose graph. Each pose graph edge's transformation is an input into q local point cloud registration algorithm to obtain a more accurate global pose, which will be used to update each edge's transformation in the refined pose graph. 
Finally, the refined pose graph is optimized to output a more accurate global pose trajectory, which transforms each single-view point cloud (or depth map) into the global coordinate system in the volumetric fusion module to create a mesh of the whole garden.

In the following, Section~\ref{section:depth-fusion-module} introduces the \textcolor{black}{disparity} fusion module including the stereo vision algorithms, the disparity fusion algorithm, and the post-processing step. Section~\ref{section:pose-estimation} introduces the pose fusion module, including the global pose graph and refined pose graph. Section~\ref{section:volumetric-fusion} introduces the volumetric fusion module.

\subsection{\textcolor{black}{Disparity} Fusion Module} \label{section:depth-fusion-module}
\subsubsection{\textcolor{black}{Disparity} Estimation} \label{section:depth-estimation}
In this stage, stereo vision algorithms with complementary properties estimate the initial disparity maps from the stereo images. Based on common sense and experience, classical stereo vision algorithms (e.g.~\cite{SGM}) perform better at the edges and small objects while the methods based on deep learning (e.g.~\cite{Dispnet}) perform better at other aspects \textcolor{black}{(e.g. flat planes, close shots)}. 
We have chosen DispNet~\citep{Dispnet} and Semi-global matching~\citep{SGM} as suitable representatives to compute the initial disparity maps in our project, but other stereo vision algorithms can be used as well. 
In the following, the initial disparity maps and auxiliary information (\textcolor{black}{intensity and gradient information}) are fed into the disparity fusion network to get a refined disparity map. 

\subsubsection{\textcolor{black}{Disparity} Fusion} \label{section:depth-fusion}
\textcolor{black}{In order to obtain a more accurate \textcolor{black}{disparity} map robustly, fusing \textcolor{black}{disparity} maps from multiple sources is a good solution considering cost and performance, under the assumption that the initial \textcolor{black}{disparity} inputs are from the same viewpoint at the same time. Fusing multiple input \textcolor{black}{disparity} maps to get a refined \textcolor{black}{disparity} map output is called \textcolor{black}{disparity} fusion, and we base it on Sdf-man~\citep{sdf-man} with some small differences, motivated by a machine learning ensemble approach. As demonstrated in~\citep{sdf-man}, the \textcolor{black}{disparity} fusion algorithm Sdf-man can refine the initial \textcolor{black}{disparity} inputs effectively and produce a more accurate \textcolor{black}{disparity} map robustly compared with its initial \textcolor{black}{disparity} inputs, even where the \textcolor{black}{disparity} inputs are inaccurate on their own. 
%That is the reason why we include the \textcolor{black}{disparity} fusion algorithm in the proposed framework rather than a single \textcolor{black}{disparity} estimation method merely. Please notice that the \textcolor{black}{disparity} fusion (whose input must include the initial \textcolor{black}{disparity} information) is different from the \textcolor{black}{disparity} estimation (whose input has no \textcolor{black}{disparity} information). Thus, readers could use more advanced \textcolor{black}{disparity} estimation methods along with a \textcolor{black}{disparity} fusion algorithm rather than the mere \textcolor{black}{disparity} estimation methods in the proposed framework.
}

%\begin{figure}
%	\centering
%	\includegraphics[width =0.95\linewidth]{figure//Theory//fusion_definition.png}
%	\caption{\textcolor{black}{Depth fusion of multiple initial input depth maps to get a more accurate fused output depth map. }} 
%	\label{fig:depth_fusion_definition}
%\end{figure}

\textcolor{black}{Similar to the GAN approach, Sdf-man~\citep{sdf-man} consists of two adversarial networks (refiner and discriminator) to perform a mini-max two-player game strategy to make the refiner network output a more accurate \textcolor{black}{disparity} map.
However, unlike standard GANs~\citep{Standard_GAN}, the input is the initial disparity maps plus intensity and gradient information rather than random noise, and its output is deterministic during inference.} 

\textcolor{black}{
For the sake of readability, we summarize the Sdf-man~\citep{sdf-man} method; more background and details can be found in the original paper.}

\begin{figure*}
	\centering
	\subfloat[]{\includegraphics[width = 9cm]{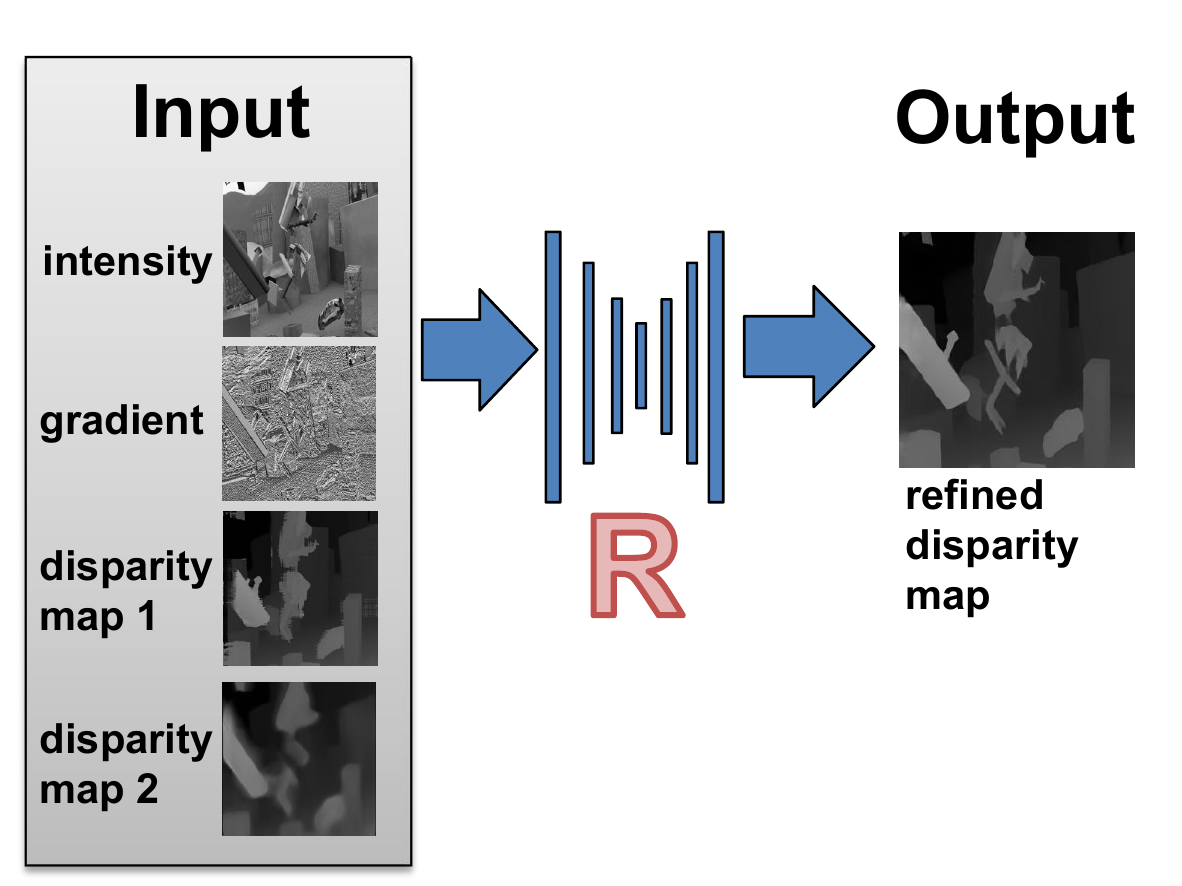}}
	
	\subfloat[]{\includegraphics[width = 7cm]{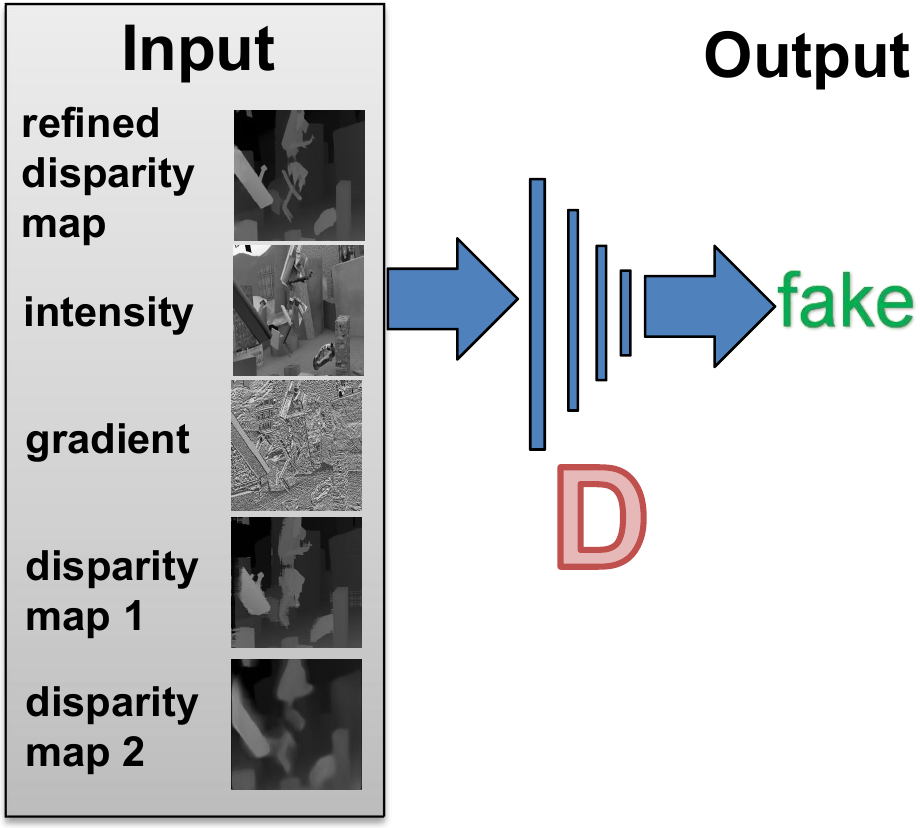}} 
	\hspace{1cm}
	\subfloat[]{\includegraphics[width = 7cm]{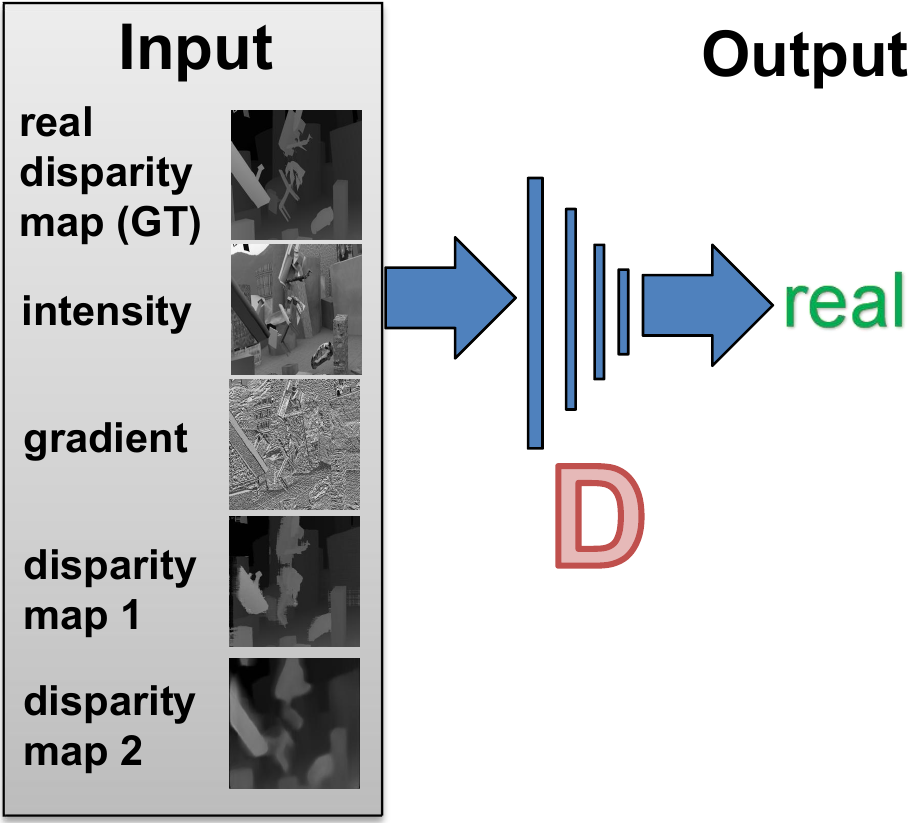}} 
	\caption{\textcolor{black}{Overview
		of the disparity fusion algorithm Sdf-man. The refiner network~\textit{R} is trained to map initial disparity maps (\textit{disparity map 1} and \textit{disparity map 2}) from two stereo vision algorithms to the ground truth disparity map based on the corresponding image information (intensity, gradient). The refiner network \textit{R} attempts to produce a refined disparity map, which is closer to the ground truth. The discriminator network \textit{D} attempts to discriminate whether its input is the real (\textcolor{black}{real disparity map (GT)} from the ground truth) or a fake (\textcolor{black}{refined disparity map} from~\textit{R}). The refiner network and discriminator network are updated alternately. (\textbf{a})~Refiner: a network to produce a refined disparity map; (\textbf{b})~Negative examples (fake): a discriminator network with refined disparity maps as input; (\textbf{c})~Positive examples (real): a discriminator network with real disparity maps (GT) as input.} \textcolor{black}{For small details in this figure, readers are recommended to view the electronic version.}} 
	\label{fig:Whole_diagram}
\end{figure*}

The refiner neural network $R$ (which is similar to the generator $G$ in~\citep{Standard_GAN}) is trained to output a refined disparity map that is not classified as ``fake'' by the discriminator network $D$. The discriminator network $D$ is trained simultaneously to conclude that the input disparity map from the ground truth is real and the input disparity map from the refiner network $R$ is fake. With a minimax two-player game strategy, it leads the output distribution from the refiner to approximate the real \textcolor{black}{disparity} data distribution. The full system pipeline is shown in Figure~\ref{fig:Whole_diagram}.

\unskip 

To train the refiner network and discriminator network, the following loss function is used in a fully supervised way:

\begin{equation}
	\begin{split}
		\mathcal{L}(R,D) &= \theta _{1}\mathcal{L}_{L_1}^{Ld}(R)  \\
		&+ \theta _{2}\mathcal{L}_{sm}^{Ld}(R) + \theta _{3}\sum_{i=1}^{M}\mathcal{L}_{GAN}^{Ld}(R,D_{i}) 
	\end{split}  \label{eq: eq1-sdf-man}
\end{equation}

\noindent
where $\theta _{1}$, $\theta _{2}$, $\theta _{3}$ are the weight values of the different loss terms. $M$ is the number of scale levels used. 
$R$ represents the refiner network and $D$ represents the discriminator network.
$\mathcal{L}_{L_1}^{Ld}(R)$ is a gradient-based $L_1$ distance training loss, which applies a bigger weight to the disparity information at the scene edges to avoid blurring at scene edges. 
$\mathcal{L}_{sm}^{Ld}(R)$ is a gradient-based smoothness term, which is used to propagate more accurate disparity values from scene edges to other areas, assuming that the disparity values of neighboring pixels should be close if their image intensities are similar.
$\mathcal{L}_{GAN}^{Ld}(R,D_{i})$ is a disparity relationship training loss, which assists the refiner in outputting a disparity map whose distribution is closer to the real distribution.
$Ld$ is the labelled data, which is fed in the supervised learning process.

In this paper, \textcolor{black}{we updated the original method~\citep{sdf-man} to improve its performance} in the real robot application with the following two practical strategies: 
\par
(1) \textbf{\textcolor{black}{Maximum Distance}} strategy: The \textcolor{black}{disparity} fusion network does not require to output all the \textcolor{black}{disparity} information in the source stereo images because the mobile robot needs more accurate depths of the nearby surroundings (rather than the remote scene). Thus, a maximum distance threshold $max\_dist$ constrains the output of the \textcolor{black}{disparity} fusion network \textcolor{black}{rather than the maximum disparity threshold $max\_disp$. 
More specifically, in the initial stages, at the end of the refiner network, it uses the function \emph{tanh} to output an intermediate map $w$ and uses the function $initial\_disp = \frac{max\_disp \cdot (w + 1)}{2}$ to convert the intermediate map $w$ into the disparity map $initial\_disp$. The intermediate map's size (width, height, channel) is identical to the disparity map's. In this paper, the difference is that we use a modified function $new\_disp = \frac{2fb}{max\_dist \cdot (w+1)}$ to map the \emph{tanh} output to the new disparity map $new\_disp$ where  $f$ and $b$ are the focal length and baseline of the stereo vision camera. This strategy effectively reduces the disparity fusion error (see the experiments in Section~\ref{exp: depthfusion}). The theoretical proof can be found in Appendix~\ref{Appendix-Formular-derivation-maximum-distance}.} 
\par
(2) \textbf{High Definition} strategy: The initial \textcolor{black}{disparity} fusion network~\citep{sdf-man} outputs a \textcolor{black}{disparity} map with the same resolution as that of the stereo images, which will result in small details being lost in the fused \textcolor{black}{disparity} map. To produce a more detailed result in the fused \textcolor{black}{disparity} map, the new \textcolor{black}{disparity} fusion network is required to output an HD (High Definition) \textcolor{black}{disparity} map first. The ratio between the HD width resolution and the initial width resolution is $\phi_w$ and the ratio between the HD height resolution and the initial height resolution is $\phi_h$. Then the HD \textcolor{black}{disparity} map is downsampled to the initial resolution (same as the input stereo images). 
\textcolor{black}{The refiner and the discriminator from Sdf-man are able to adjust their networks adaptively to any resolution of images, as shown in Sdfman~\citep{sdf-man} (Figures 2,3). 
What is done differently here is: 
1) upscaling the data input first and then inputing the upscaled data into the networks to train autonomously; 
2) downscaling the disparity output from the refiner network to the resolution of the initial stereo images as the final result. The reason why the up-and-down resolution transformation strategy works is that Sdf-man will include more neurons in the refiner and discriminator network structure to capture more small details autonomously when the input resolution becomes higher. Experiments in Section~\ref{exp: depthfusion} demonstrate this is an effective strategy.}

After disparity fusion, a refined disparity map is produced registered to the left view in the stereo configuration. Given that there are still some outliers in the refined disparity map, the refined disparity map is converted into a local point cloud with the outliers removed in the next stage.

\subsubsection{Post-processing} \label{section:post-processing}
The disparity post-processing part consists of three steps: 1) converting the disparity map into a depth map;  2) converting the depth map into a local point cloud using the camera calibration parameters; 3) \textcolor{black}{removing the point cloud outliers}.

In the first step, the refined disparity map is converted into the depth map using Equation~\eqref{eq: eq2-post-processing}.  

\begin{equation}
	depth = \frac{focal \_ length \times baseline} {disparity}
		\label{eq: eq2-post-processing}
\end{equation}

In the second step, the depth map is back-projected into a 3D point cloud using the camera intrinsic parameters~\citep{book-computer-vision}. In Equation~\eqref{eq: eq3-post-processing}, $(u, v)$ is the coordinate of the 2D point on the image plane and $(X, Y, Z)$ is the corresponding 3D point in the camera space. 
$f_x, f_y$ are the focal lengths on the $x, y$ axes and $c_x, c_y$ are the coordinates of the principle point on the $x, y$ axes. $D$ is the depth value.      

\begin{equation}
	D 
	\left[
	\begin{matrix}
	u \\
	v\\
	1
	\end{matrix}
	\right]
	=
	\left[
	\begin{matrix}
	f_x & 0 & c_x \\
	0 & f_y & c_y \\
	0 & 0 & 1
	\end{matrix}
	\right]
	\left[
	\begin{matrix}
	X \\
	Y\\
	Z
	\end{matrix}
	\right]
	\label{eq: eq3-post-processing}
\end{equation}

In the third step, any points that 1) have less than $N_{p}$ neighboring points in a given sphere with the radius $radius$ or 2) are farther away from their $N_n$ neighboring points than a threshold distance ratio $dist\_ratio$ (which is equal to the mean distance to their $N_n$ neighboring points divided by the distance standard deviation to the $N_n$ neighboring points) \textcolor{black}{are treated as outliers and are removed}.

After post-processing stage, the local single-view ($72^{\circ}$) and full-view ($360^{\circ}$) point clouds in the current frame are produced and used in the following pose fusion module.

\subsection{Pose Fusion Module} \label{section:pose-estimation}
\subsubsection{First Stage: Global Pose Graph}  \label{section:global-coarse-graph-construction}
As a single-view point cloud ($72^\circ$) has limited features for point cloud matching, we combine the local point clouds from five views (5 stereo vision cameras) at the same time to form a full-view ($360^{\circ}$) point cloud for point cloud registration. This representation improves tracking robustness against fast or big transformations, by using the full-view ($360^{\circ}$) point clouds for global registration. 
The full-view ($360^\circ$) point cloud $X^m_i$, the single-view  ($72^\circ$) point cloud $X^s_i$ and their corresponding global pose $P_i$ in the world frame will constitute a pose graph node $V_i$. 
The global pose $P_i$ is also the pose of the node $V_i$.
Every pair of nodes $V_i$ and $V_j$ have an edge $E_{ij}$ containing a transformation matrix $T_{ij}$ that aligns their full-view point clouds $X^m_i$ and $X^m_j$.  
The nodes $V_i$ and the edges $E_{ij}$ form a global pose graph $G(V, E)$. $V$ is the set of nodes and $E$ is the set of edges.

Every pair of full-view point clouds in different nodes are registered to get the corresponding edge's transformation matrix by using the \textcolor{black}{feature-based} fast global registration algorithm~\citep{Fast-global-registration}, \textcolor{black}{which essentially implements loop closure (LC).} 
The global pose graph $G(V, E)$ is then optimized to produce a coarse global pose trajectory $\{P_1, ... ,P_n\}$ by minimizing loss:
\begin{equation}
\mathcal{L}(G(V,E)) = \underset{ \{P_1, ... , P_n\} \in SE(3)^n}{\arg\min} \sum_{(i,j)\in E} || T_{ij} - P_{i}P_{j}^{-1} ||_{F}^2
\label{eq: graph-optimization-global}
\end{equation}
$|| \bullet || _{F}$ is the Frobenius norm, $SE(3)$ is the special Euclidean group in 3 dimensions and $n$ is the number of the pose graph nodes.
\textcolor{black}{The loss function represented by Equation~\eqref{eq: graph-optimization-global} derives from the maximum likelihood estimation formula in~\citep{Fast-optimization-PGO} when setting the uncertainty of the translations to be identical to the rotations. We use the optimization method in~\citep{Fast-optimization-PGO} to minimize the loss function based on the implementation available at: \url{https://github.com/gabmoreira/maks}. 
}

\textcolor{black}{
The derivation and proof of Equation~\eqref{eq: graph-optimization-global} can be found in Appendix~\ref{Appendix-Formular-derivation-loss-function}. The same deduction and optimization methods can be applied to Equation~\eqref{eq: graph-optimization-local} below.}

\subsubsection{Second Stage: Refined Pose Graph} \label{section:refined-global-graph-reconstruction}
%Inherited from the global pose graph $G(V,E)$, t
\textcolor{black}{The global pose graph has edges between every pair of nodes possibly, even if there is little or no overlap between their corresponding single-views. This can lead to local distortions. This stage optimizes the global pose graph by using the poses from overlapping views.}
The refined pose graph $\tilde{G}(\tilde{V}, \tilde{E})$ is initialized by the global pose graph $G(V,E)$.
 
 Since the fast global registration algorithm~\citep{Fast-global-registration} is not accurate enough compared with local registration algorithms (e.g. GICP~\citep{GICP}, DUGMA~\citep{DUGMA}), we input each edge's transformation matrix $T_{ij}$ (from $E_{ij}$ in the global pose graph $G(V, E)$) into the local registration algorithm GICP~\citep{GICP} as a global initialization to align the corresponding single-view ($72^\circ$) point clouds\footnote{When using the extrinsic transformation matrices to merge the point clouds from the five stereo vision cameras on the camera ring, the error from the extrinsic parameters will cause the full-view ($360^{\circ}$) point cloud to be not as accurate as the single-view point cloud. } $X^s_i$ and $X^s_j$. 
 The local registration algorithm GICP~\citep{GICP}  outputs a new estimated transformation matrix $T_{ij}^l$ for the corresponding edge.
 
 \textcolor{black}{Then, every pair of point clouds $X^s_i$ and $X^s_j$ are transformed into the same coordinate system using the transformation matrix $T_{ij}^l$. We calculate the number of the corresponding point pairs}
 \textcolor{black}{within a distance threshold (similar to finding corresponding closest point pairs in ICP).}
 \textcolor{black}{The overlap percentage of one point cloud after registration is equal to the number of the corresponding pairs divided by the number of the points in the point cloud. The overlap percentage of the pair of point clouds after registration is equal to the overlap percentage of the point cloud with the fewest points.}
  Based on the registration results above \textcolor{black}{and the coarse global pose trajectory from the first stage}, the edges $\tilde{E}_{ij}$ in the refined pose graph are updated using the following three rules:
  \par
 (1) \textbf{Prune}: If the overlap percentage of the two point clouds after registration is lower than the threshold $OL_{min}$, prune the edge (remove the edge between the two nodes). 
 \par
 (2) \textbf{Update}: If the overlap percentage of the two point clouds after registration is higher than threshold $OL_{max}$ and if the transformation $P_iP_j^{-1}$ (whose 6D pose is denoted as the 6D vector $\vec{v}_P$) between the two nodes ($V_i$, $V_j$) is similar to the newly calculated transformation $T_{ij}^l$ (whose 6D pose is denoted as the 6D vector $\vec{v}_T$), update the edge.
 \par
 (3) \textbf{Keep}:  As for the `else' case, keep but do not update the edge transformation. 
 
 Equation~\eqref{eq: edge-updating} gives the precise logic for the three rules above: 
 
 \begin{equation}
 \tilde{T}_{ij}=
 \begin{cases}
  Null & \beta < {OL}_{min} \; \; \; \; \; \; \; \; \; \; \; \; \; \; \; \; \; \; \; \; \; \; \; \; \; \; \; \; \; \;  \; \; \; \,    [rule1]\\
 T_{ij}^l & \beta > {OL}_{max} \; \& \; |\vec{v}_P - \vec{v}_T| < \vec{v}_{th} \; \; \;  \; \; [rule2]\\
 T_{ij} &  else \; \; \; \; \; \; \; \; \; \; \; \; \; \; \; \; \; \; \; \; \; \; \; \; \; \; \; \; \;  \; \; \; \;  \; \; \; \;  \; \; \; \;  \; \; \, [rule3]\\
 \end{cases}
 \label{eq: edge-updating}
 \end{equation}
 
 In Equation~\eqref{eq: edge-updating}, $\tilde{T}_{ij}$ is the transformation matrix of the edge $\tilde{E}_{ij}$ in the refined pose graph. $\beta$ is the overlap percentage of the two point clouds after registration. $\vec{v}_{th}$ is 6D pose threshold in vector format and $| \bullet |$ means getting the absolute value of each element to form a new vector. 
 $Null$ denotes "deleting this edge". 
 \textcolor{black}{The two rules (Prune and Update) act on the edge set to constrain the loss function - Equation~\eqref{eq: graph-optimization-local}. An accurate constraint could give a more accurate global pose estimation, which is demonstrated by the ablation study in Section~\ref{section:ablation_study_pose_fusion}.}
 After edge refinement, the refined pose graph $\tilde{G}(\tilde{V}, \tilde{E})$ is optimized using Equation~\eqref{eq: graph-optimization-local} to produce a more accurate global pose trajectory $\{\tilde{P}_{1}, ..., \tilde{P}_{n}\}$. The refined accurate global pose $\tilde{P}_{i}$ of each node $\tilde{V}_i$ and their corresponding single-view point cloud (or depth map) will be used in the volumetric fusion process to construct the \textcolor{black}{surface mesh} of the whole garden.
 
 \begin{equation}
 \mathcal{L}(\tilde{G}(\tilde{V}, \tilde{E})) = \underset{ \{\tilde{P}_1, ... , \tilde{P}_n\} \in SE(3)^n}{\arg\min} \sum_{(i,j)\in \tilde{E}} || \tilde{T}_{ij} - \tilde{P}_{i}\tilde{P}_{j}^{-1} ||_{F}^2
 \label{eq: graph-optimization-local}
 \end{equation}
 
 \textcolor{black}{To conclude, we have proposed a two-stage full-view-to-single-view global-coarse-to-local-fine pose trajectory estimation method in this subsection. We name this proposed method for pose trajectory estimation as `Multi-stage Pose Trajectory Estimation with Joint Information (MPTEJI)'. Algorithm~\ref{fig:algorithm-MPTEJI} shows the pseudocode of the proposed algorithm MPTEJI, which gives a formal overview of the whole proposed method.}
 
 \textcolor{black}{}
 \begin{algorithm}{\color{black}}
 	\caption{Multi-stage Pose Trajectory Estimation with Joint Information (MPTEJI)}
 	\label{fig:algorithm-MPTEJI}
 	\color{black}
 	\textbf{Input:} single-view ($72^\circ$) and full-view ($360^\circ$) point clouds
 	\begin{algorithmic}[1]
 		\Procedure{with \textbf{full view}}{} \Comment{\textit{~$1^{st}$ stage}}
 		\State \textbf{global} registration (feature-based) $\to$ Loop Closure
 		\State \textbf{coarse} pose graph $G(V,E)$ ~~~$\gets$ Equation~\eqref{eq: graph-optimization-global}
 		\EndProcedure 		
 		\Procedure{with \textbf{single view}}{} \Comment {\textit{~$2^{nd}$} stage}
 		\State pose graph inheritance  ~~~~~~~~~~~$\gets$ $G(V,E)$
 		\State \textbf{local} registration (ICP-based)  $\gets$ $T_{ij}$
 		\State edge refinement ~~~~~~~~~~~~~~~~~~~~~~$\gets$ Equation~\eqref{eq: edge-updating}
		\State \textbf{refined} pose graph $\tilde{G}(\tilde{V}, \tilde{E})$ ~~~$\gets$ Equation~\eqref{eq: graph-optimization-local}
 		\EndProcedure
 	\end{algorithmic}
  	\textbf{Output:} an accurate global pose trajectory $\{\tilde{P}_{1}, ..., \tilde{P}_{n}\}$   
 \end{algorithm}

\subsection{Volumetric Fusion Module \label{section:volumetric-fusion}}
\textcolor{black}{Fusing the range images (containing depth information) into a voxel-based volumetric scene representation is called volumetric fusion~\citep{Volumetric-fusion}.}
The refined accurate global pose trajectory $\{\tilde{P}_{1}, ..., \tilde{P}_{n}\}$ gives where 
\textcolor{black}{
to integrate the associated RGB-D range images projected from the single-view point clouds into a voxel-grid-based TSDF (Truncated Signed Distance Field) volume. The value of each voxel here represents the signed distance to the closest surface interface in the global space, which is in turn used to obtain the mesh of the reconstructed scene, using the marching cubes~\citep{Marching-cubes-1987} algorithm. }

\textcolor{black}{
More specifically, the single-view point clouds are projected back into the image planes to get the related depth maps first, creating again RGB-D images. 
We use the refined single-view point clouds to compute the depth maps rather than use the original depth maps from the disparity fusion (Section~\ref{section:depth-fusion}) directly because the single-view point clouds after the third step `outlier removal' in the post-processing section (Section~\ref{section:post-processing}) are more accurate.  
Then the pairwise data (the RGB-D images and the corresponding global poses) are integrated into the global TSDF volume using the technique from ~\cite{Kinect-fusion,Scalable-volumetric-fusion}. Finally, we extract the surface mesh using marching cubes~\citep{Marching-cubes-1987,Marching-cube}, based on a publicly available implementation\footnote{~\url{https://github.com/qianyizh/ElasticReconstruction/tree/master/Integrate}}.
}

\textcolor{black}{The volumetric fusion module produces a smooth and watertight 3D mesh of the reconstructed scene in the global coordinate system. The corresponding dense 3D point cloud of the reconstructed scene can be produced by extracting all the vertexes of the 3D mesh above. 
Simply put, the volumetric fusion performs like a weighted average filter in the 3D global space to reduce the noise and remove the outliers from multiple local segments by using the joint global information in the global coordinate system. That is the reason why we use the volumetric fusion to extract the mesh and the corresponding point cloud sequentially, rather than stitching the single-view point clouds together using their corresponding pose directly.}         

\section{Experiments}\label{Experiment}

All the experiments in this section are conducted on a machine with Intel Core i7-12700KF processor (12 cores, 20 threads, 25 MB cache, up to 5 GHz) and Nvidia GeForce GTX 1080 Ti. Section~\ref{exp: dataset} gives the description of the real outdoor garden dataset  we released and used in this paper. Section~\ref{exp: depthfusion} evaluates the performance improvement of the \textcolor{black}{disparity} fusion module quantitatively compared with the initial \textcolor{black}{disparity} inputs~\citep{SGM,Dispnet}, the ground truth of DSF~\citep{DSF} and the initial version of Sdf-man~\citep{sdf-man}. Section~\ref{exp: posefusion} evaluates the global pose trajectory's accuracy from the pose fusion module quantitatively compared with ORB-SLAM3~\citep{orbslam3} and the "reconstruction system" in the latest version (0.15.1) of Open3D~\citep{Open3d}. Section~\ref{exp: volumetricfusion} gives a view of the reconstructed point cloud from the volumetric fusion module qualitatively and quantitatively compared with Open3D~\citep{Open3d}. 

\subsection{Dataset Description} \label{exp: dataset}

\begin{figure*}	
	\begin{center}
		\subfloat[Manually labeled 3D semantic model of the whole garden. Legend: grass (bright green), rail fence (blue), tree (dark green), hedge (brown), board fence (soil color), rose (red), boxwood (dark blue), potted plant (light blue)~\citep{Trimbot-garden-ICCV2017}]{\includegraphics[ width=\linewidth]{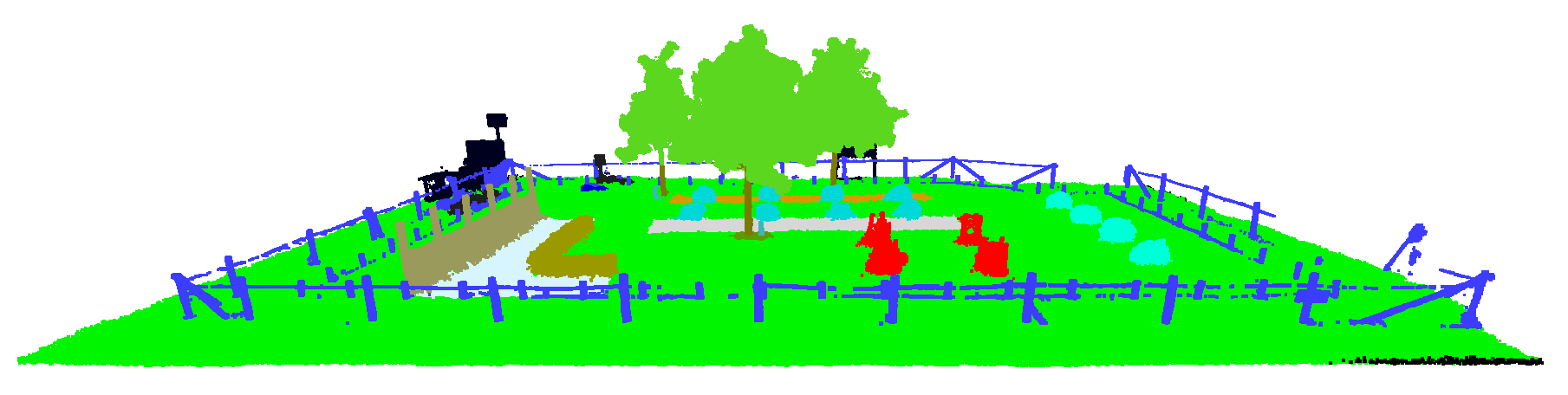}}
	\end{center}
	\begin{multicols}{2}
		\begin{center}
			%\flushleft
			\subfloat[The robot platform used for collecting data]{\includegraphics[ height=4.7cm]{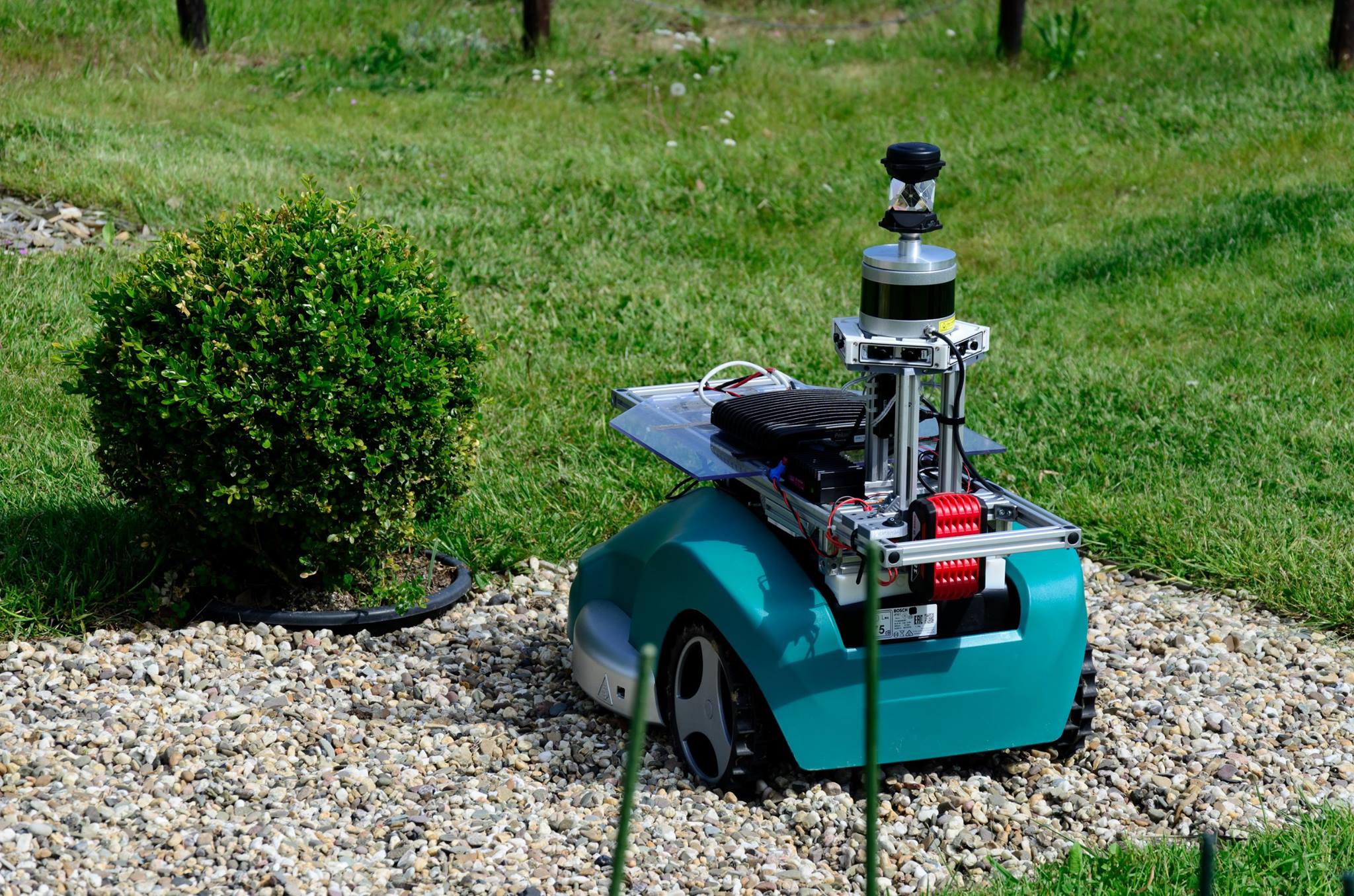}}
		\end{center}
		\begin{center}
			%\flushright
			\subfloat[Route for the training \& testing dataset in the garden]{\includegraphics[ height=4.7cm]{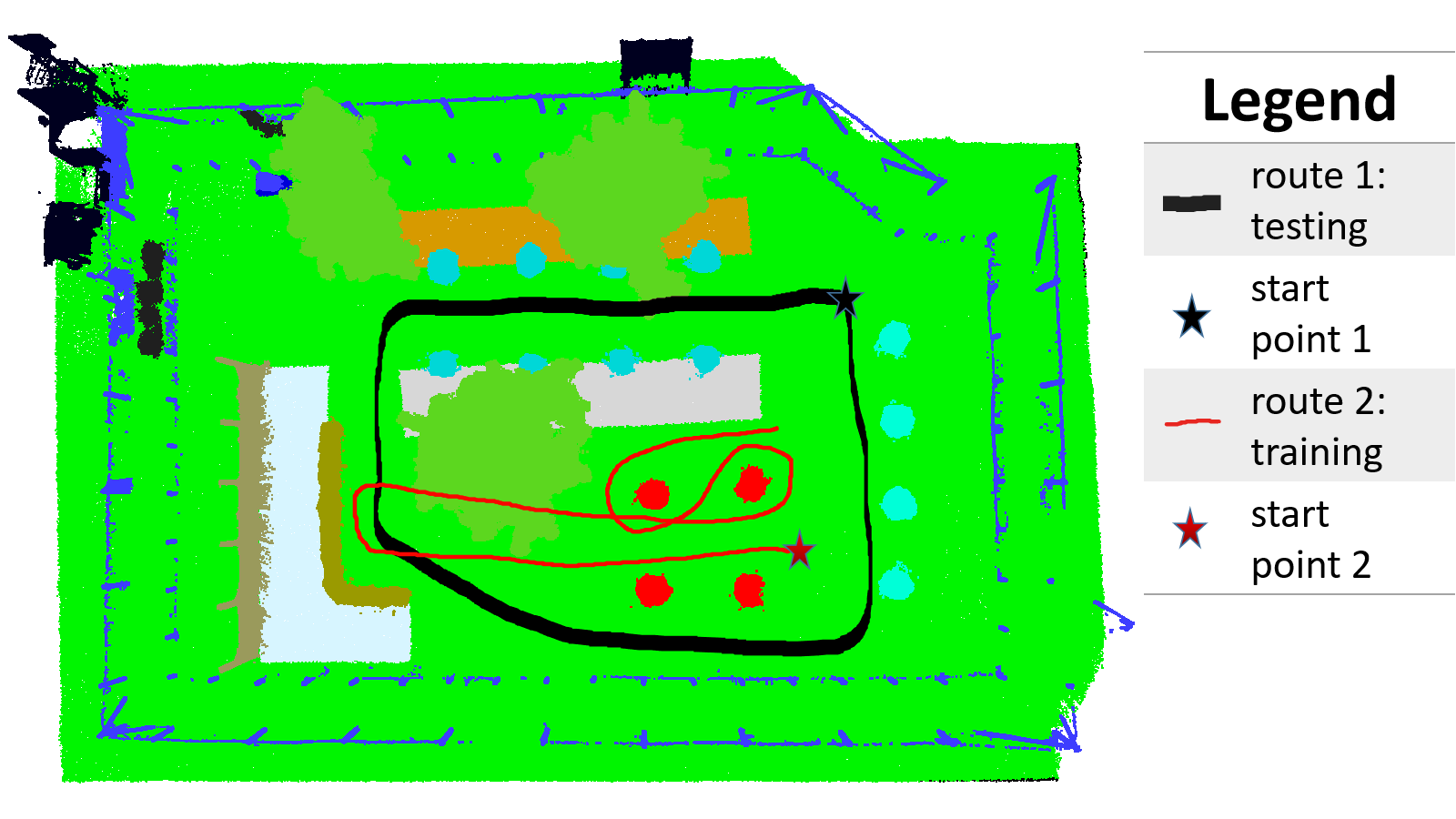}}
		\end{center}
	\end{multicols}
	\caption{Figure (a) shows the 3D model of the real garden; Figure (b) shows the robot platform for collecting data; Figure (c) shows the routes for training and testing dataset. }
	\label{fig:garden-3D-model}
\end{figure*}

Figure~\ref{fig:garden-3D-model} shows the 3D model of the outdoor garden, the robot platform and the route path for collecting the raw data.
\textcolor{black}{All the data in our dataset were recorded within the same half day to avoid interference from vegetation growth.} 
The raw data is from the "test\_around\_garden" bagfile\footnote{ \url{https://www.research.ed.ac.uk/en/datasets/trimbot2020-dataset-for-garden-navigation-and-bush-trimming}.} in the Trimbot Garden 2017 dataset~\citep{Trimbot-garden-ICCV2017, Trimbot-raw-data-2020}. The raw data was divided into two parts in the post-processing step: one for network training and one for testing. Figure~\ref{fig:garden-3D-model} (c) shows the robot navigation path for the training and testing datasets. In Figure~\ref{fig:garden-3D-model} (c), the "route 1" trajectory (black loop curve) around the whole garden is for the SLAM testing and the "route 2" trajectory\footnote{All the scenes in "route 1" can be seen in "route 2".} (red curve) is for the network training (e.g. depth estimation, semantic segmentation, etc.). We use a robot (See Figure~\ref{fig:garden-3D-model} b) equipped with a ring of 5 stereo vision cameras (for live operations),  Velodyne Puck (VLP-16) Lidar sensor (for sparse lidar scans collection), STIM300 IMU sensor and Topcon PS Series Robotic Total Station position tracking system (for ground-truth positions) to collect the raw images, sparse Lidar scans and the global pose of the robot. The raw stereo vision images are calibrated and rectified using the Kalibr package\footnote{https://github.com/ethz-asl/kalibr}. The sparse Lidar scan from the Velodyne Puck (VLP-16) Lidar sensor is projected to the camera plane of each left camera in the 5 stereo settings.  Robot navigation poses were recorded in the coordinate system of Topcon PS Series Robotic Total Station along with STIM300 IMU sensor first and then transformed into each image sensor's global pose. Structure-from-motion~\citep{SFM} is used to refine each image sensor's pose subsequently. 
The 3D model of the whole garden is collected using  \textcolor{black}{Leica ScanStation P15} equipment and is semantically labelled manually.  Figure~\ref{fig:garden-3D-model} (a) shows the semantic 3D model of the whole garden. Using the semantic 3D model and each camera's pose in the garden, the dense depth map and semantic map are acquired by projecting the semantic 3D model into each camera's plane. \textcolor{black}{More details about the data collection process can be found in Appendix~\ref{Appendix_more_description_about_dataset}.}

\begin{table*}[H] 
	\caption{\textcolor{black}{Parameters of the \textit{Trimbot Wageningen SLAM Dataset}}} 
	\label{table: dataset}
	\begin{tabular}{ >{\color{black}} p{9.cm}| >{\color{black}} p{7.5cm}}
		\toprule
		\textbf{Parameter Name} & \textbf{Parameter Value} \\
		\midrule
		The number of panoramic stereo camera rigs  & 1; \\ 
		The number of stereo vision cameras & 5; \\
		The number of image sensors       & 10; \\
		The number of panoramic frames - $360^\circ$     & In the training subset: 68; ~~In the test subset: 67; \\ 
		The number of stereo vision frames - $72^\circ$     & In the training subset: 340; ~In the test subset: 335; \\
		Image resolution & $752 \times 480$ pixels (width $\times$ height); \\
		The mean relative pose between adjacent frames ([translation on $x$ axis, translation on $y$ axis, translation on $z$ axis, roll, pitch, yaw]) & [0.29 m,  0.21 m, 0.00 m, 9.04 ~deg, 0.97 deg, 1.16 deg]; \\
		The standard deviation of the relative pose between adjacent frames ([translation on $x$ axis, translation on $y$ axis, translation on $z$ axis, roll, pitch, yaw])      & [0.18 m, 0.18 m, 0.00 m, 13.32 deg, 0.76 deg, 0.89 deg]; \\
		The maximum translation value on each axis between adjacent frames &   X axis:0.47 m; ~~~~Y axis: 0.67 m; ~~~Z axis: 0.02 m;  \\		
		The maximum rotation value on each axis between adjacent frames &  X axis: 81.64 deg; Y axis: 3.21 deg; Z axis: 4.46 deg;  \\	
		Data support & RGB | intensity, dense depth, sparse lidar, semantics, pose, point cloud, calibration.\\	
		\bottomrule
	\end{tabular}
\end{table*}

\begin{figure*}
	\includegraphics[width=1\linewidth]{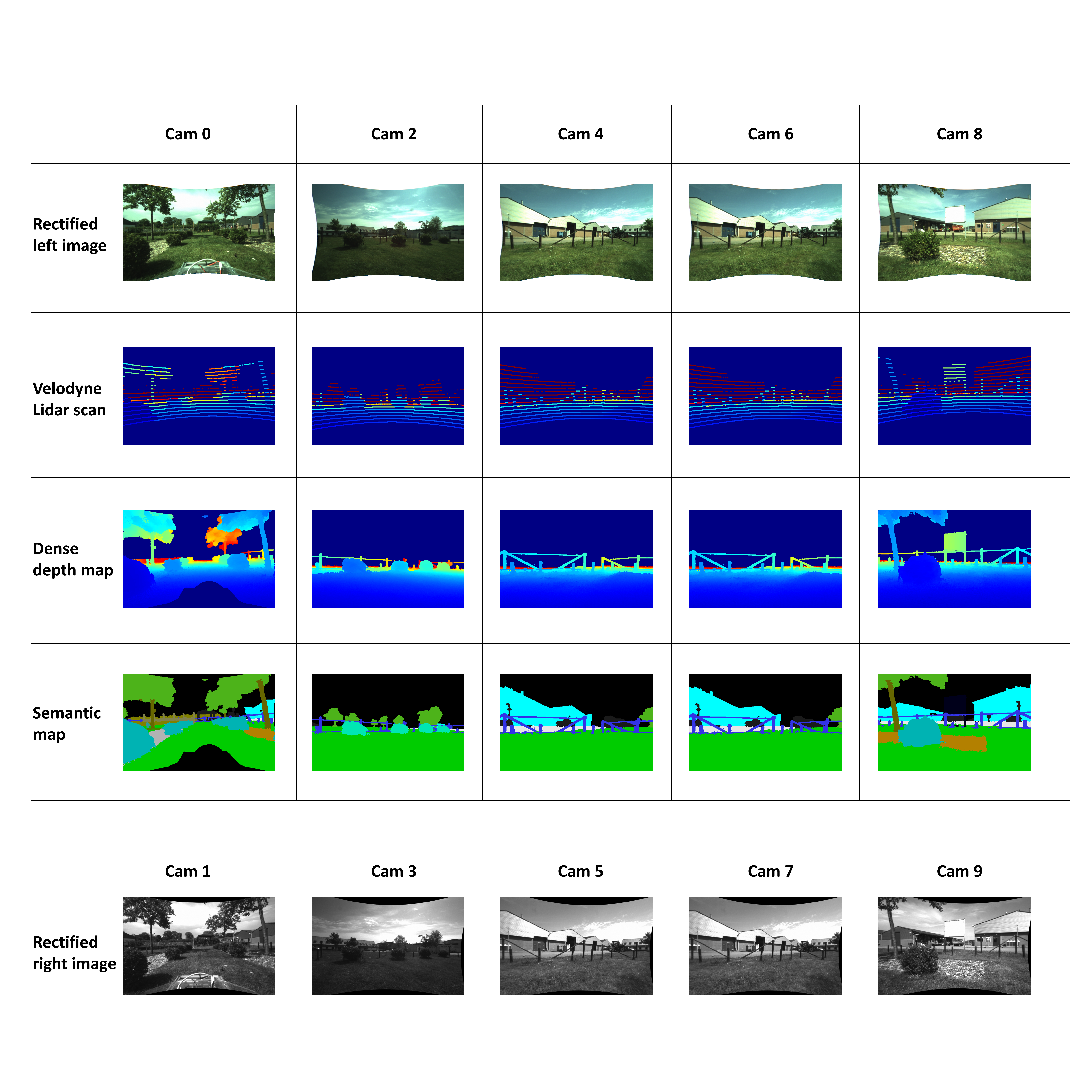}
	\caption{Trimbot Wageningen SLAM Dataset}
	\label{fig:garden-dataset}
\end{figure*}

Figure~\ref{fig:garden-dataset} shows frames from the new dataset ``The Trimbot Wageningen SLAM Dataset'', which is the augmentation of the Trimbot Garden 2017 dataset used in the semantic reconstruction challenge of ICCV 2017 workshop ``3D Reconstruction meets Semantics''~\citep{Trimbot-garden-ICCV2017}. In the new Trimbot Wageningen SLAM dataset, we release all the rectified images from the 10 image sensors (5 stereo vision cameras) ranging from cam\_0 to cam\_9 (See Figure~\ref{fig:garden-with-robot}b for the position of the 10 image sensors). The newly released sparse Lidar scan (from the onboard Lidar sensor - Velodyne Puck (VLP-16)), dense depth map and semantic map are in the coordinate system of each left image sensor (Cam 0, Cam 2, Cam 4, Cam 6, Cam 8).  Each image sensor's global pose in the garden, their intrinsic parameters and distortion models are available in the new dataset. We subsample one out of every 10 frames from the initial raw data bagfile to form the new dataset. 
%In the new dataset, the test subset has 67 frame times (each frame time consists of data from five left camera views and there are 67 * 5 = 335 frames in total) and the training subset has 68 frame times (each frame time consists of data from five left camera views and there are 68 * 5 = 340 frames in total). The mean of the relative 6D pose between adjacent frame times is [0.29 m,  0.21 m, 0.00 m, 9.04 deg, 0.97 deg, 1.16 deg] ([translation on $x$ axis, translation on $y$ axis, translation on $z$ axis, roll, pitch, yaw]) and its standard deviation is [0.18 m, 0.18 m, 0.00 m, 13.32 deg, 0.76 deg, 0.89 deg]. The dataset is difficult for SLAM algorithms because the rotations and translations between consecutive frames are big, which makes the SLAM algorithms easily lose tracking and break down. The maximum translation values between consecutive frame times on the $x$, $y$ and $z$ axes are 0.47 m, 0.67 m and 0.02 m respectively. The maximum rotation values between adjacent frame times on the  $x$, $y$ and $z$ axes are 81.64 deg, 3.21 deg and 4.46 deg. All the images or maps in the dataset have a resolution of $752 \times 480$ ($width \times height$) pixels. 
\textcolor{black}{Table~\ref{table: dataset} lists the key dataset properties and their corresponding values in the Trimbot Wageningen SLAM Dataset.} 
Figure~\ref{fig:garden-dataset} gives an overview of the new dataset.
See the dataset website for more details: \url{https://github.com/Canpu999/Trimbot-Wageningen-SLAM-Dataset}.

\subsection{\textcolor{black}{Disparity} Fusion Module}  \label{exp: depthfusion}
In the \textcolor{black}{disparity} fusion module, we use the SGM~\citep{SGM} (with Matlab implementation\footnote{Matlab Implementation URL:\url{https://www.mathworks.com/help/vision/ref/disparitysgm.html}}) and Dispnet\footnote{The authors of Dispnet~\citep{Dispnet} were our project partners and they trained Dispnet on the project dataset to get their best performance.}~\citep{Dispnet} stereo vision algorithms to get the initial disparity maps. With the initial disparity maps and auxiliary information (left intensity image and left gradient information), we train our supervised disparity fusion network on our outdoor real garden dataset - "Trimbot Wageningen SLAM Dataset". All 340 samples ($\approx 50\%$) in the training set are used to train and all 335 samples ($\approx 50\%$) in the test set are used to test. 
The initial supervised method~\citep{sdf-man} is named "Sdfman-initial" and the updated method using the two practical strategies presented in Section \ref{section:depth-fusion-module} is named "Sdfman-star". 
The parameter $max\_dist$ is set to 5 meters\footnote{\textcolor{black}{As the focal length and baseline are fixed, depth estimation is inversely proportional to its corresponding disparity value - see Equation~\eqref{eq: eq2-post-processing}. When the depth is 5 meters, the corresponding disparity is about 3 pixels. Estimated depth values larger than 5 meters (i.e. disparity value smaller than 3 pixels) have a larger error compared with depths closer than 5 meters.}}. 
The parameters $\phi_w$ and $\phi_h$ (resolution ratio between the HD and initial image width and height) are set to 2. 
Disparity fusion algorithms DSF~\citep{DSF} and "Sdfman-initial"~\citep{sdf-man} are compared to the new method "Sdfman-star". 
Additionally, an ablation study is conducted by adding two internal comparison algorithms ("Sdfman-max-dist" and "Sdfman-HR"). "Sdfman-max-dist" is an internal comparison algorithm that only applies the "Maximum Distance" strategy to "Sdfman-initial". "Sdfman-HR" is an internal comparison algorithm that only applies the "High Definition" strategy to "Sdfman-initial". Table~\ref{table: algorithm_definition_ablation} summarizes the algorithms' names with the corresponding strategies.

\begin{table}[H] 
	\caption{Algorithm definition.} 
	\label{table: algorithm_definition_ablation}
	\centering
	\begin{tabular}{cc}
		\toprule
		\textbf{Algorithm Name} & \textbf{Strategy} \\
		\midrule
		Sdfman-initial  & Default \\ %Is the bold necessary？
		Sdfman-max-dist & Maximum Distance \\
		Sdfman-HR       & High Definition \\
		Sdfman-star     & Maximum Distance + High Definition \\ 
		\bottomrule
	\end{tabular}
\end{table} 

When calculating the error of each algorithm, we omit the pixels whose ground truth depth exceeds the maximum distance threshold $ max\_dist$ = 5~m. Table~\ref{table:stereo-stereo-fusion-result} shows the accuracy of each algorithm. \textcolor{black}{We use meter (m) rather than pixel disparity as the units for the error to give a more intuitive sense of the error magnitudes.} 
With the initial input (Matlab SGM~\citep{SGM}, Dispnet~\citep{Dispnet}), DSF~\citep{DSF} reduces the mean absolute depth error from 0.40~m (Matlab SGM) and 0.24~m (Dispnet) to 0.18~m (DSF), which is larger than that of Sdfman-initial (0.09~m). Compared with Sdfman-initial (0.09~m), Sdfman-max-dist, Sdfman-HR and Sdfman-star are more accurate, which demonstrates that each of the proposed strategies contributes to improving the fusion accuracy. 
Algorithm Sdfman-star performs best with the mean absolute depth error (0.03~m) and achieves this at 34.21 frames per second. 
In the following experiments, we omit the two internal algorithms (Sdfman-max-dist and Sdfman-HR) because they are only used for the ablation study.    

\begin{table*}[H]
	\centering
	\caption{Mean absolute depth error of the \textcolor{black}{disparity} fusion on Trimbot Wageningen SLAM Dataset.} \label{table:stereo-stereo-fusion-result}
	%Please define if necessary. 
	\begin{tabular}{cccccccc}
		\hline & \multicolumn{2}{c}{\bf Inputs } & \multicolumn{2}{c}{\bf Comparison } & \multicolumn{3}{c}{\bf Ablation Study } \\ 
		\hline
		 & Matlab SGM & Dispnet & DSF  & Sdfman-initial & Sdfman-max-dist & Sdfman-HR & Sdfman-star \\
		\hline 
		Error & 0.40 m &  0.24 m & 0.18 m & 0.09 m & 0.07 m &  0.08 m & \textbf{0.03 m} \\
		\hline
	\end{tabular} 
\end{table*} 
\unskip 

Figure~\ref{fig:disp_eval_all_mean} (a) compares the mean absolute error of each frame's depth map in the test dataset from all the algorithms. 
The accuracy of Sdfman-star is better than the other algorithms at all frames, which shows the robustness of Sdfman-star. 
Define parameter badX to be the percentage of pixels whose absolute depth errors in the depth map are bigger than X * 0.025 m (X is a positive number). 
Figure~\ref{fig:disp_eval_all_mean} (b) shows the percentage of badX pixels for different badX thresholds (values of X). 
Compared with the other algorithms, Sdfman-star has fewer pixels whose absolute depth error is bigger than 0.025 m, 0.05 m, 0.075 m and 0.1 m respectively. 
More than 95\% of the pixels (bad4) from Sdfman-star have an absolute depth error less than 0.1 m. 

\begin{figure*}
	\centering

	\subfloat[]{\includegraphics[width = \linewidth]{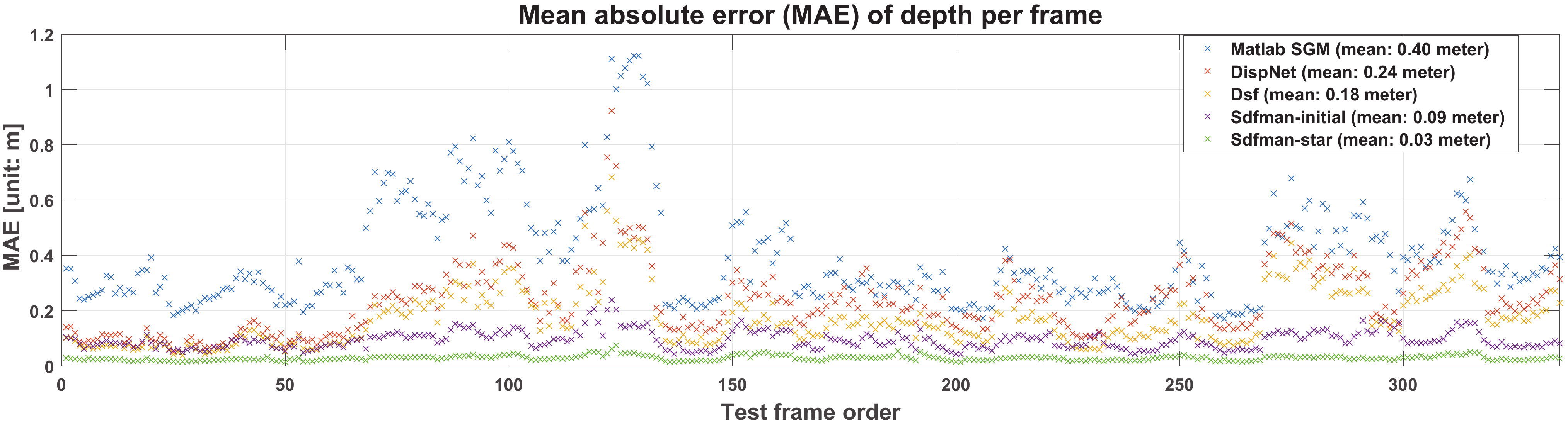}}
	
	\subfloat[]{\includegraphics[width = \linewidth]{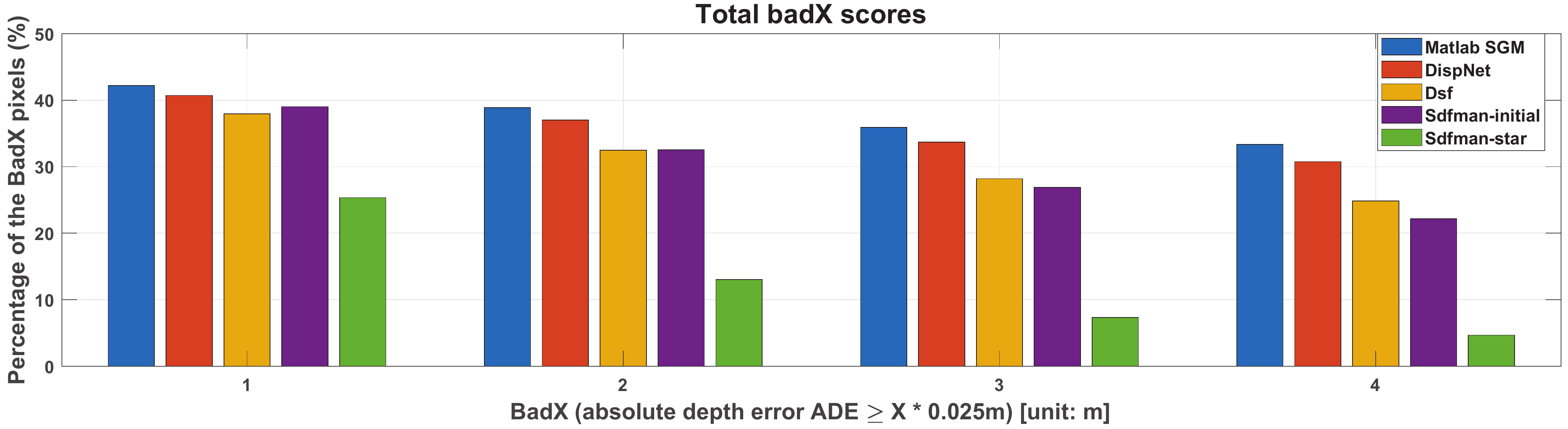}} 
	\hspace{1cm}
	\caption{The mean absolute error (MAE) of the estimated depth map per frame and the total bad X scores}
	\label{fig:disp_eval_all_mean}
\end{figure*}

Figure~\ref{fig:depth-fusion-examples-experiment} shows one qualitative result from one image sensor (Cam 0). Compared with the other algorithms, Sdfman-star is more accurate globally and also preserves small details more vividly (e.g. object edges, the trees' trunks). 

\begin{figure*}[H]
	\centering	
	\setcounter{subfigure}{0}
	\subfloat[Ground truth]{\includegraphics[width = 0.38\linewidth]{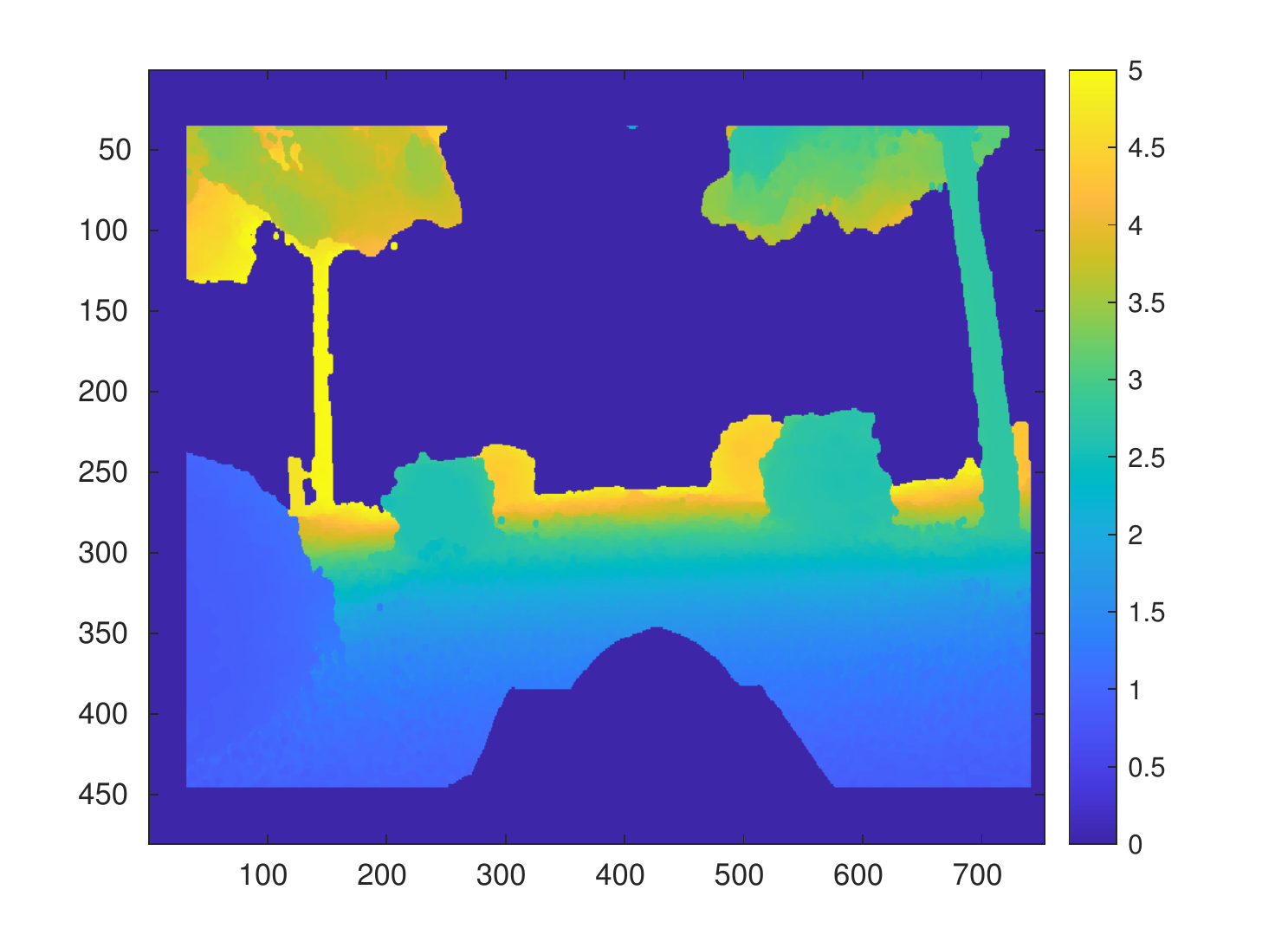}}
	\vspace{0.001cm}
	\hspace{0.1cm}
	\subfloat[RGB image]{\includegraphics[width = 0.38\linewidth]{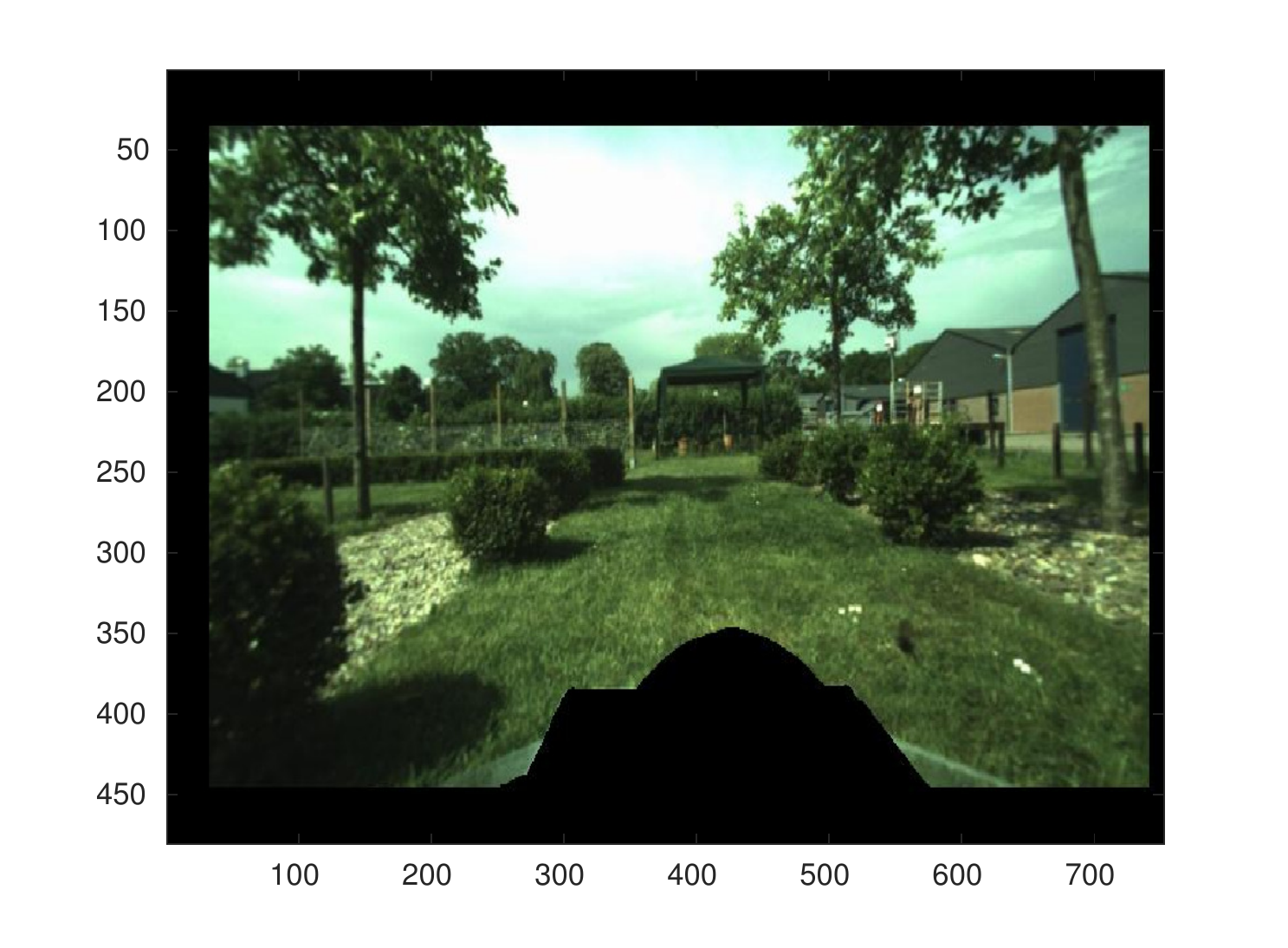}}
	\vspace{0.001cm}
	\subfloat[Matlab SGM]{\includegraphics[width = 0.38\linewidth]{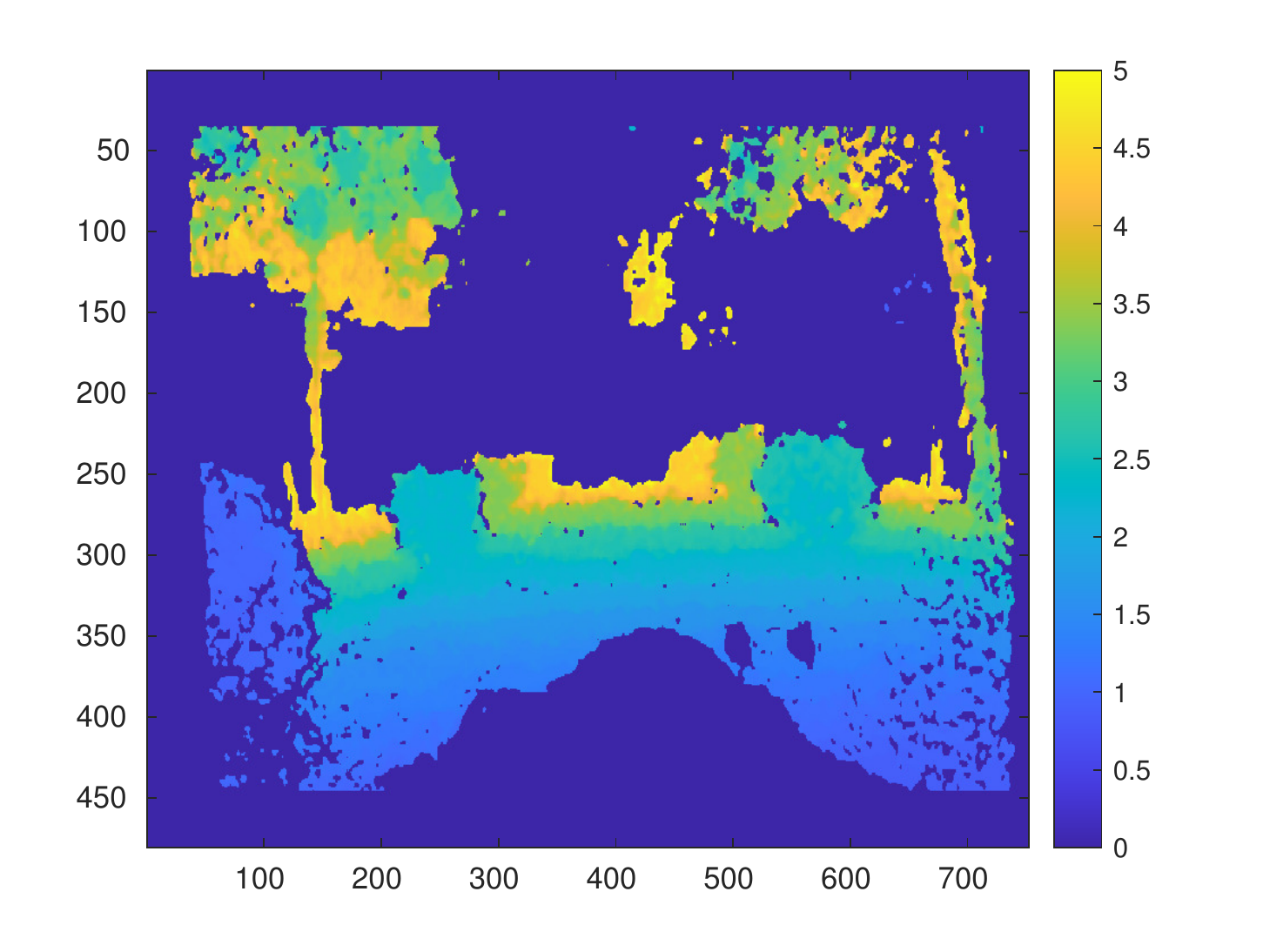}}
	\vspace{0.001cm}
	\hspace{0.1cm} 
	\subfloat[SGM error]{\includegraphics[width = 0.38\linewidth]{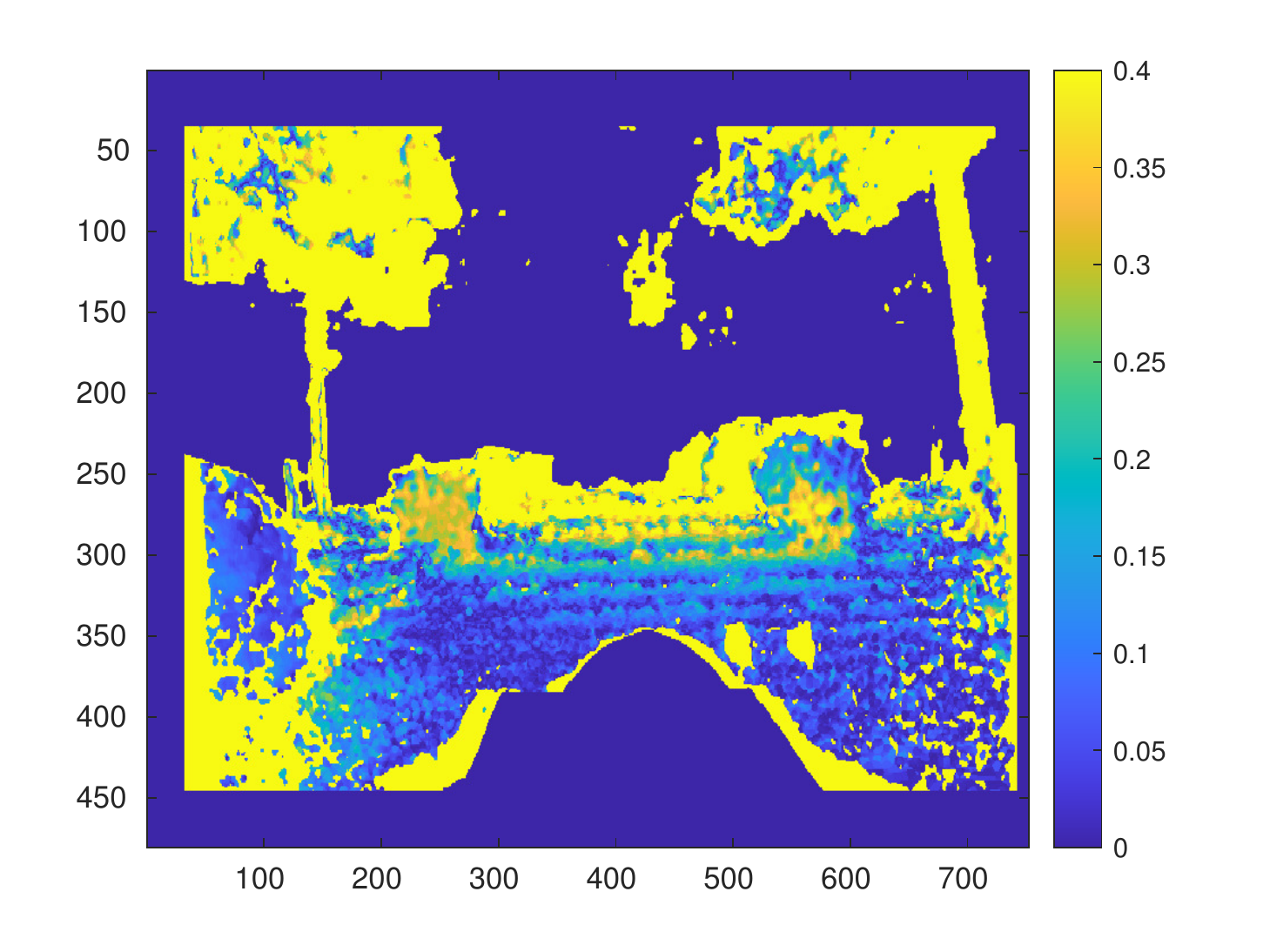}} 
	\vspace{0.001cm}
	\subfloat[Dispnet]{\includegraphics[width = 0.38\linewidth]{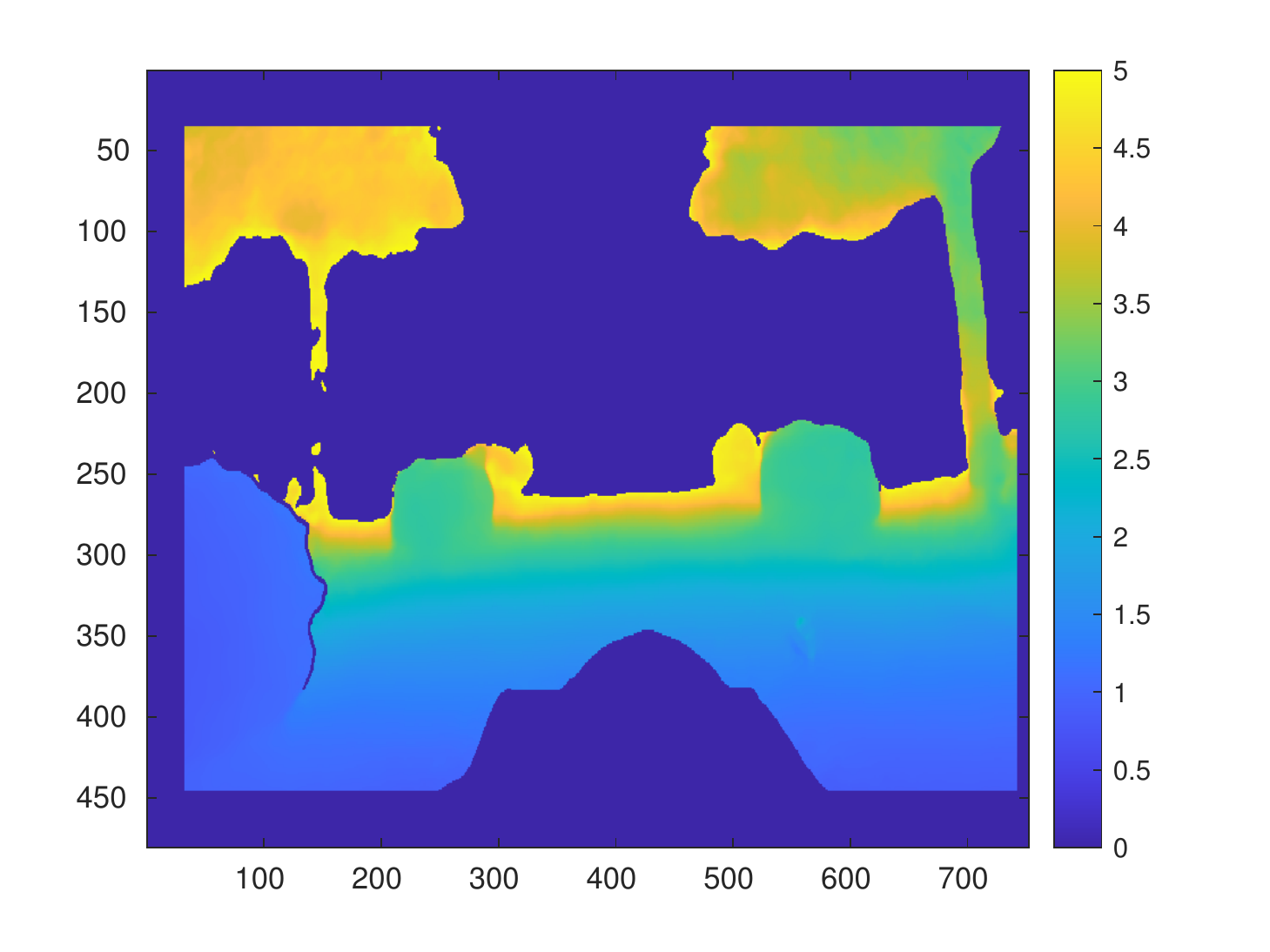}} 
	\vspace{0.001cm}
	\hspace{0.1cm} 
	\subfloat[Dispnet error]{\includegraphics[width = 0.38\linewidth]{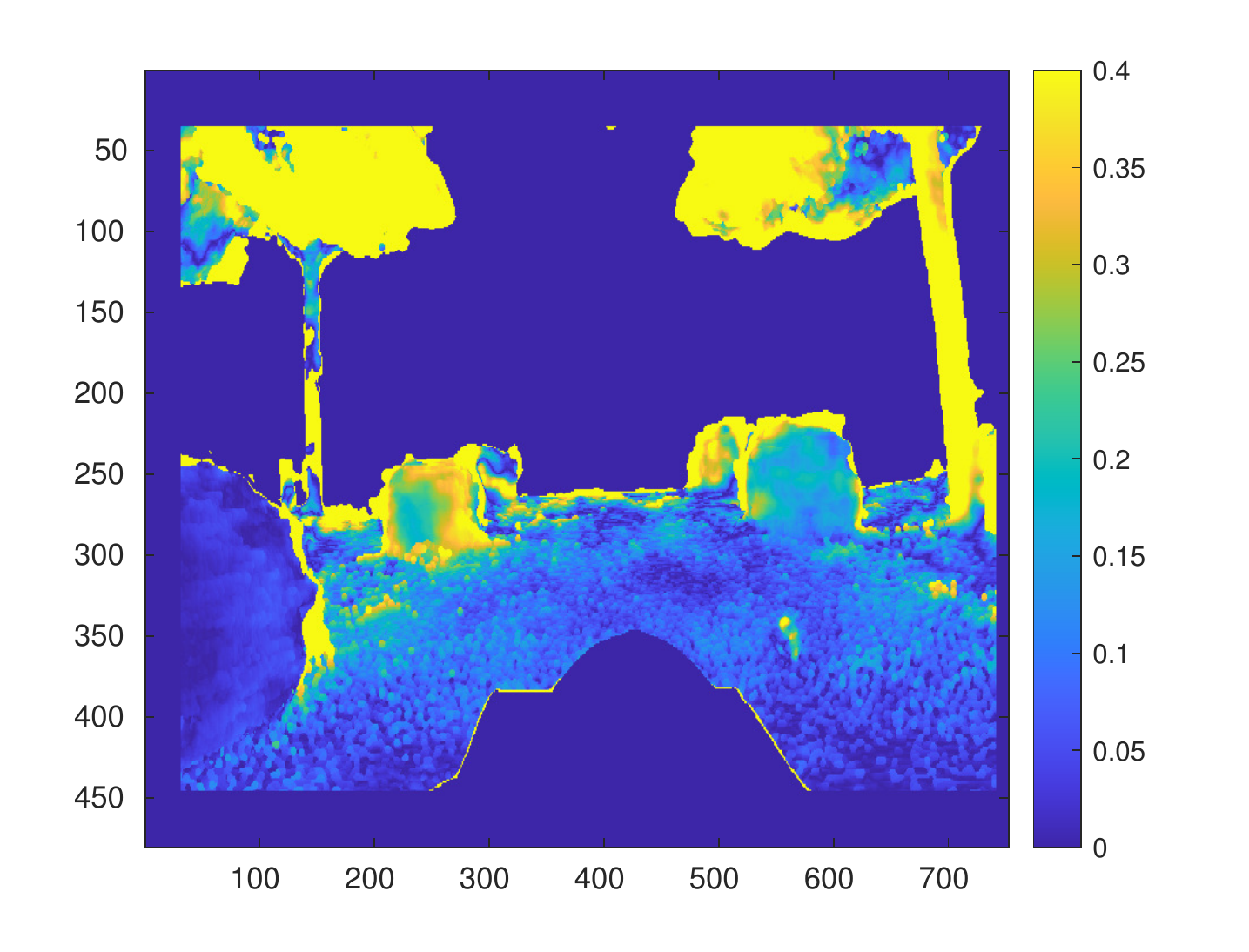}}
	\vspace{0.001cm} 
	\subfloat[Sdfman-initial]{\includegraphics[width = 0.38\linewidth]{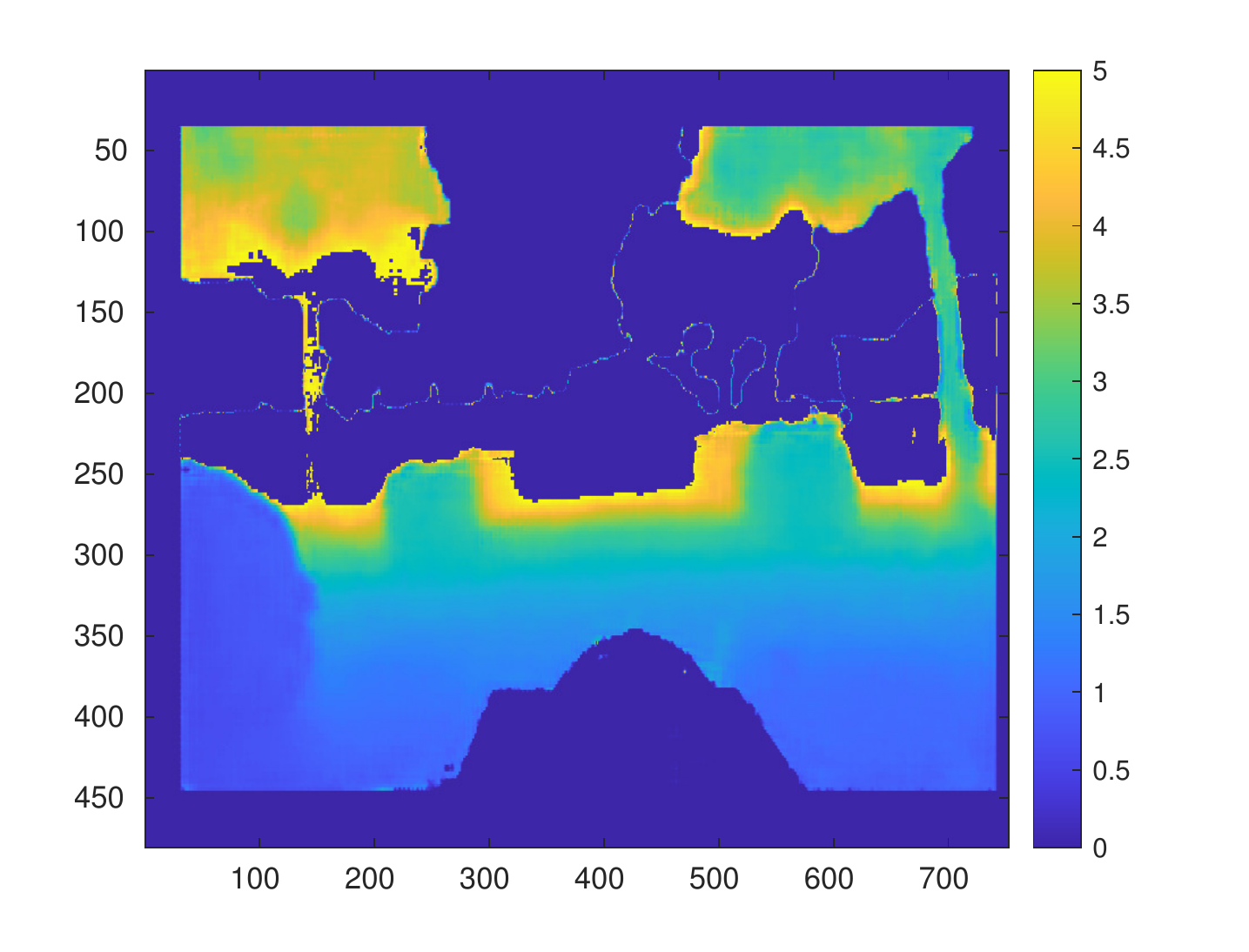}}
	\vspace{0.001cm}
	\hspace{0.1cm} 
	\subfloat[Sdfman-initial error]{\includegraphics[width = 0.38\linewidth]{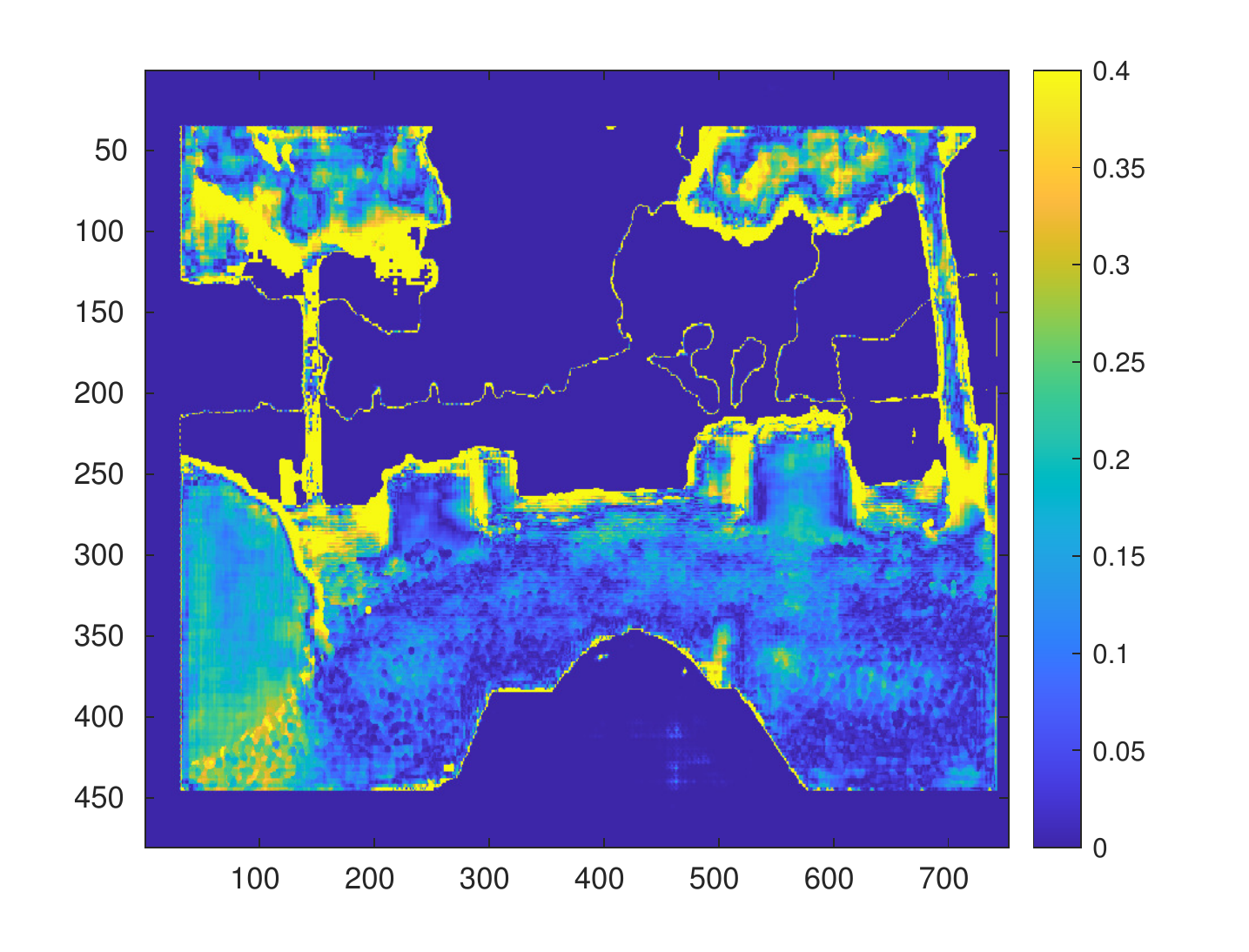}} 
	\vspace{0.001cm}
	\caption{\textit{Cont.}}
\end{figure*}

\begin{figure*}[H]\ContinuedFloat
	\centering
	\subfloat[Sdfman-star]{\includegraphics[width = 0.38\linewidth]{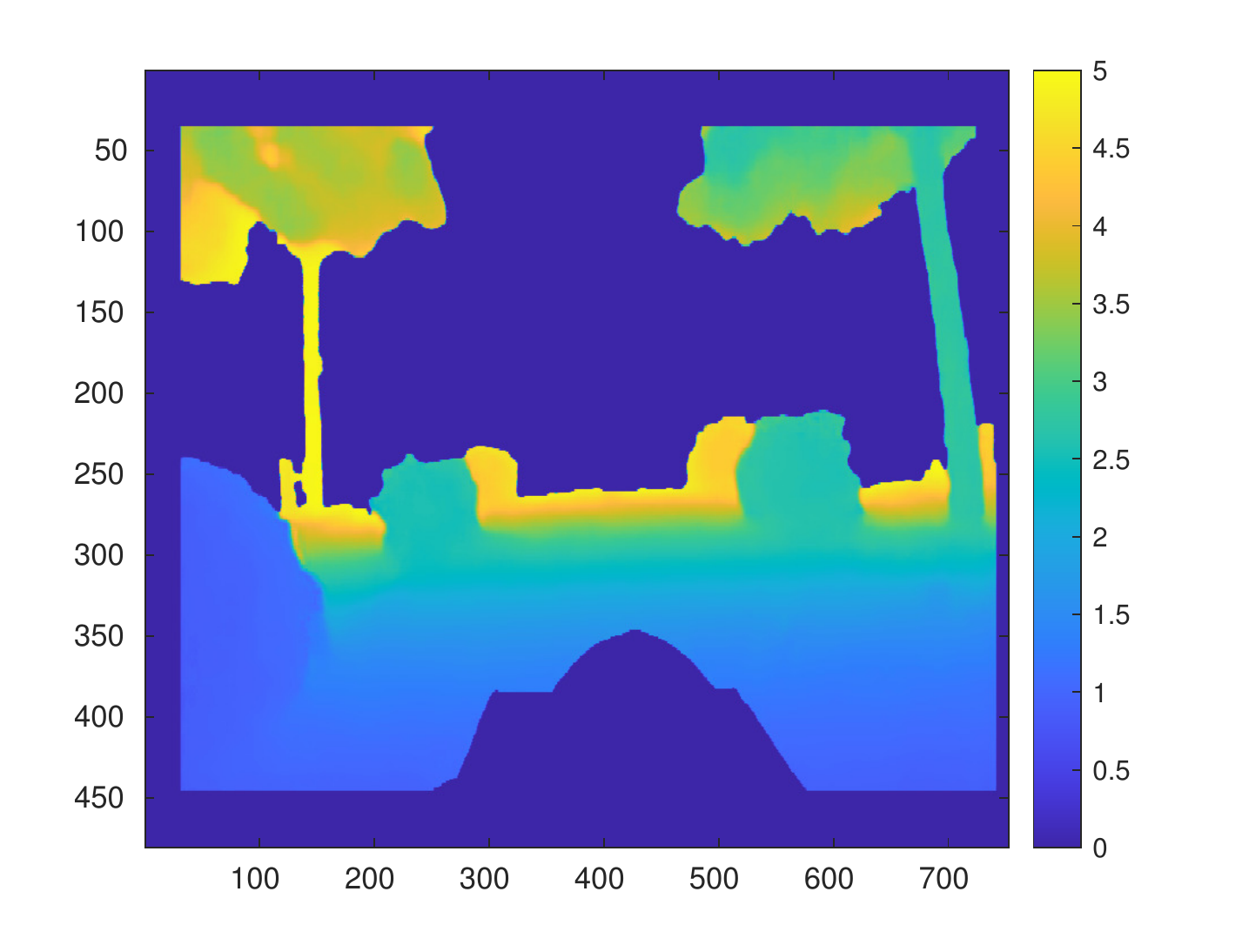}} 
	\vspace{0.001cm}
	\hspace{0.1cm} 
	\subfloat[Sdfman-star error]{\includegraphics[width = 0.38\linewidth]{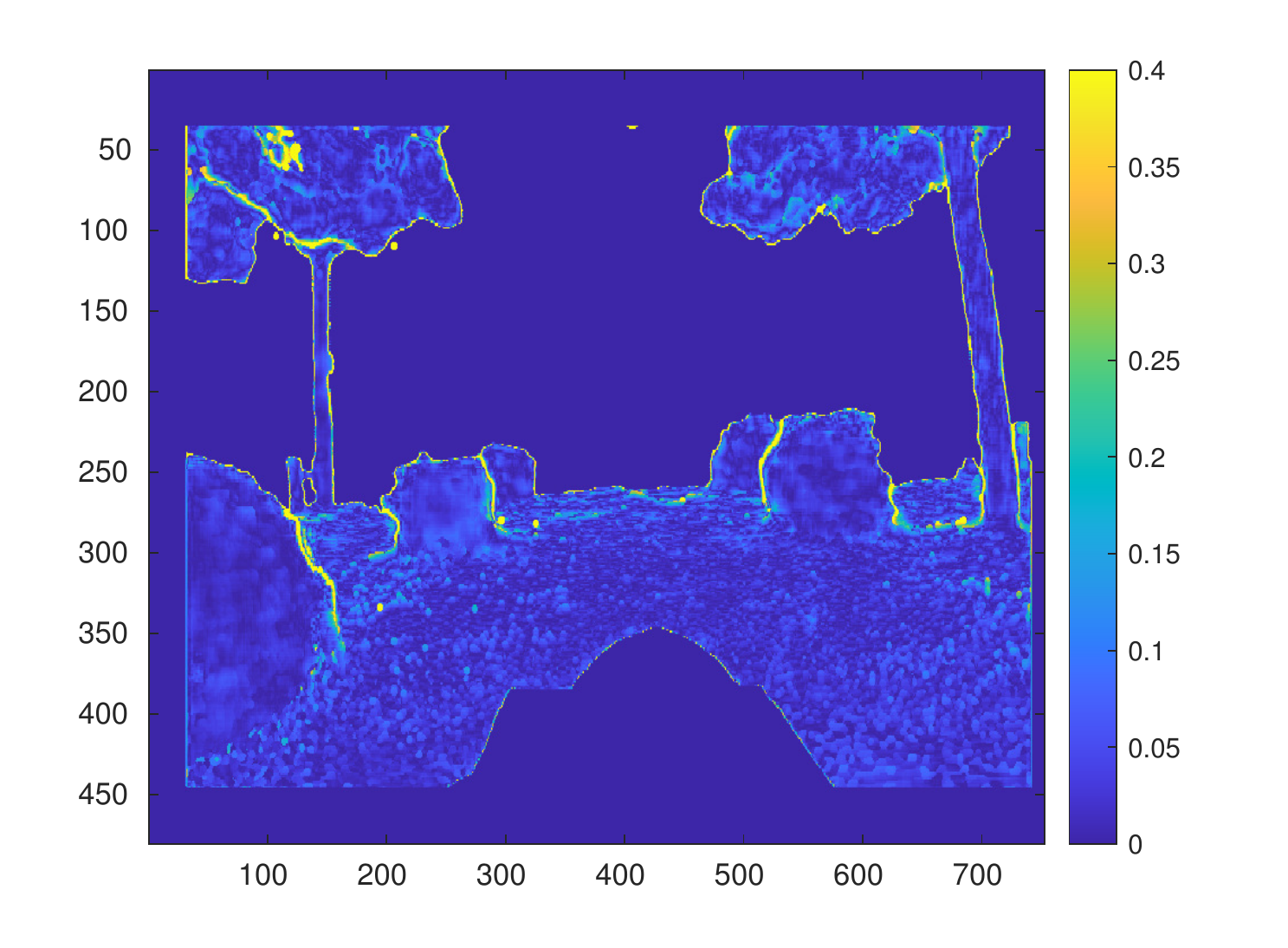}}
	\vspace{0.001cm} 
	\caption{One qualitative result for \textcolor{black}{disparity} fusion. The~lighter pixels in (\textbf{d},\textbf{f},\textbf{h},\textbf{j}) represent bigger depth error. 
 %(\textbf{a})~ground truth; (\textbf{b})~RGB image; (\textbf{c})~Matlab SGM; (\textbf{d})~Matlab SGM error; (\textbf{e})~Dispnet; (\textbf{f})~Dispnet error; (\textbf{g})~Sdfman-initial; (\textbf{h})~Sdfman-initial error (\textbf{i})~Sdfman-star; (\textbf{j})~Sdfman-star error.
 }
	\label{fig:depth-fusion-examples-experiment}
\end{figure*}

\begin{figure*}
	\centering
	\subfloat[Single-view point cloud from Sdfman-star]{\includegraphics[width = 7cm,height=5cm]{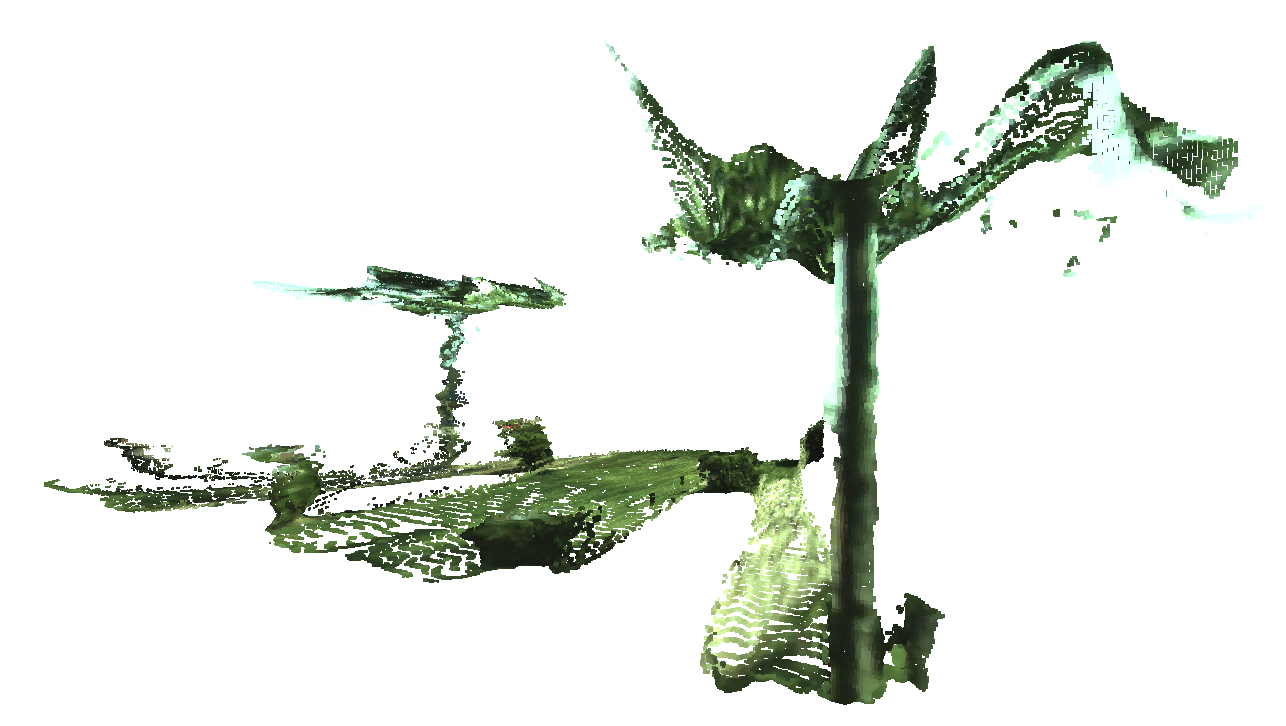}} 
	\hspace{1cm}
	\subfloat[Single-view point cloud from ground truth]{\includegraphics[width = 7cm,height=5cm]{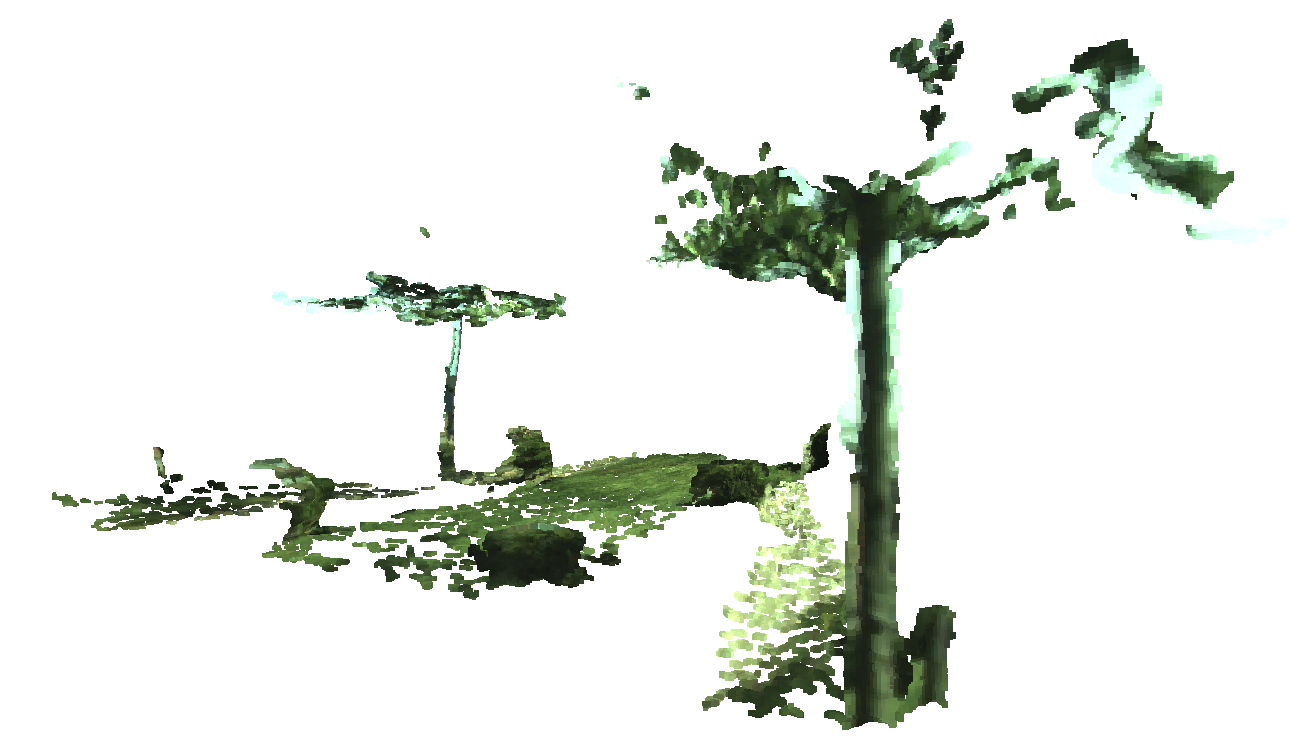}} 
	
	\subfloat[Full-view point cloud from Sdfman-star]{\includegraphics[width = 7cm,height=5cm]{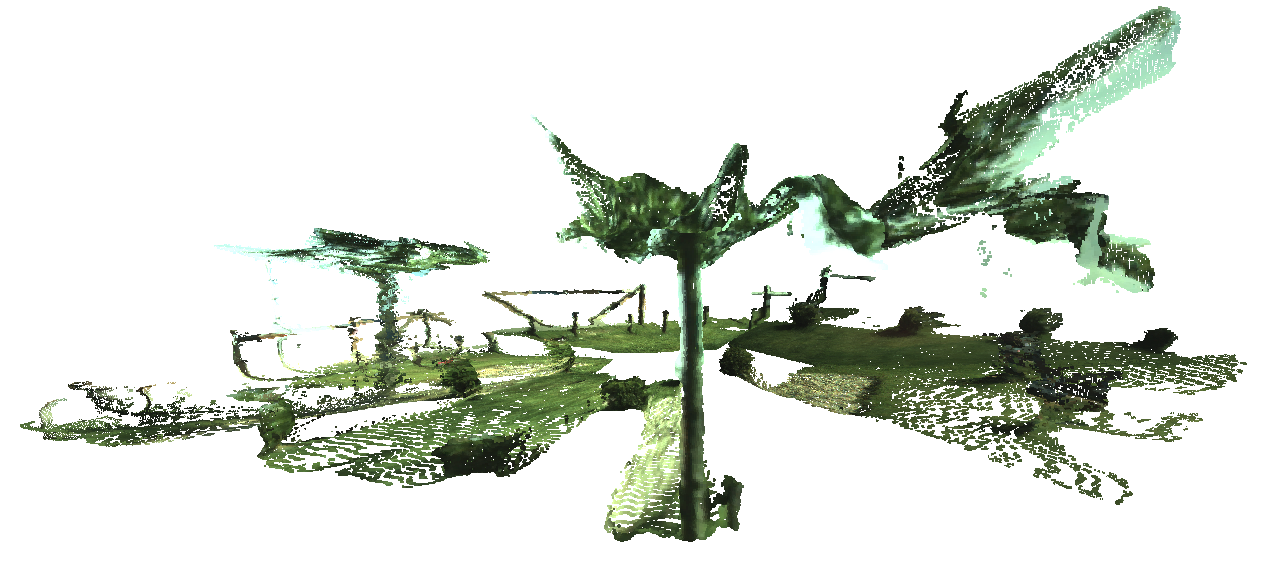}} 
	\hspace{1cm}
	\subfloat[Full-view point cloud from ground truth]{\includegraphics[width = 7cm,height=5cm]{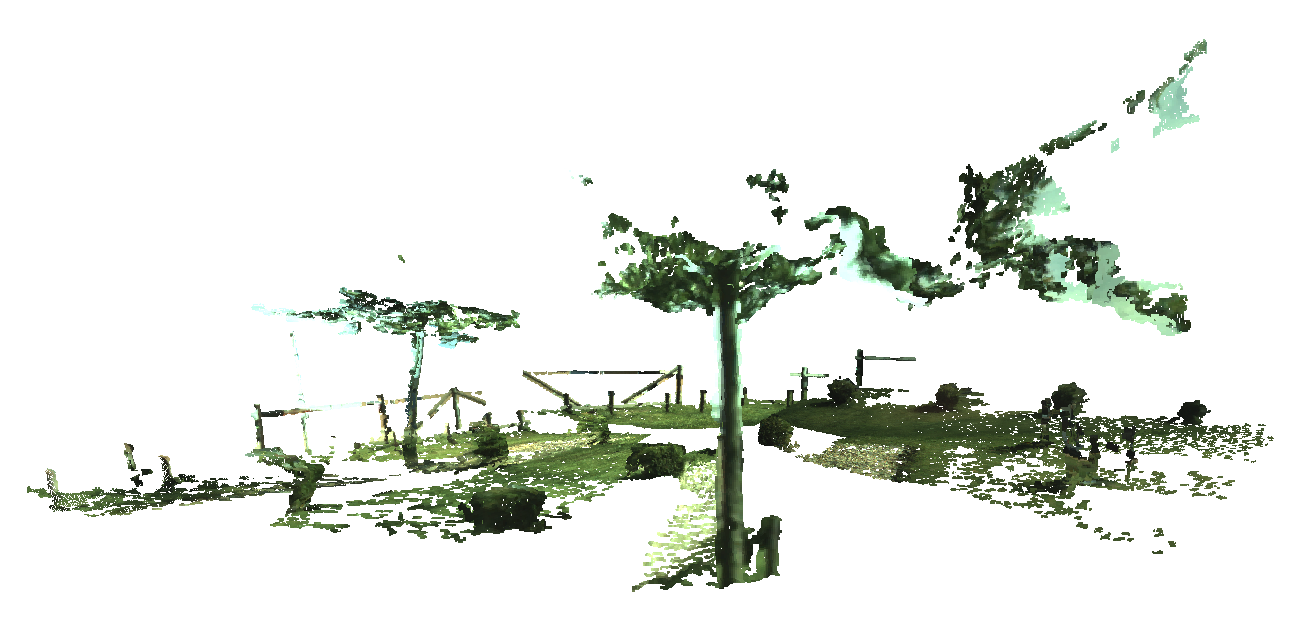}} 
	\caption{ A single-view ($72^{\circ}$) and full-view ($360^{\circ}$) point cloud from Sdfman-star and ground truth}
	\label{fig:point-cloud-from-depth-map-experiment}
\end{figure*}

In the post-processing step, $N_p$ is set as 20 and $radius$ is set as 0.05 m. $N_n$ is set as 20 and $dist\_ratio$ is set as 1.5. Figure~\ref{fig:point-cloud-from-depth-map-experiment} shows one example of the point clouds from the depth maps after \textcolor{black}{outlier removal}. Compared with the ground truth, the remote objects (e.g. trunk) in the point clouds from Sdfman-star are noisy, which can be expected. 

\unskip 

\subsection{Pose Fusion Module}  \label{exp: posefusion}
Section~\ref{section:ablation_study_pose_fusion} presents results from an ablation study to show that the strategies proposed in Section~\ref{section:pose-estimation} are effective. 
Section~\ref{section: comparison_pose_fusion_module} compares  Orbslam3~\citep{orbslam3} and Open3D~\citep{Open3d} with the proposed pose fusion method.

\textcolor{black}{Two methods are used to evaluate the 6D pose estimate accuracy. 
The first one uses the 6D pose vector $[tx, ty, tz, r, p, y]$ ([translation on $X$ axis, translation on $Y$ axis, translation on $Z$ axis, roll, pitch, yaw]). 
The unit for $tx$, $ty$, $tz$ is meters and the unit for $r$, $p$, $y$ is degrees. 
The absolute difference between the ground truth and the estimated 6D vector is a measure of the 6D pose's \textbf{accuracy on each axis}. 
The second method uses the rotation matrix and translation vector. The \textbf{overall accuracy} of the 6D poses is computed using Equation~\eqref{eq:1_exp_pose_fusion} and Equation~\eqref{eq:2_exp_pose_fusion}~\citep{Metrics_3D_Rotation_huynh2009}}

\begin{align}
E_R = ||\boldsymbol{I-R_{gt}R_{est}^{-1}}||_{F} \label{eq:1_exp_pose_fusion}
\end{align}
\begin{align}
E_t = ||\boldsymbol{t_{gt}-t_{est}}||_{F}
\label{eq:2_exp_pose_fusion}
\end{align}
where $|| \bullet || _{F}$ is the Frobenius norm. $\mathbf{R_{gt},t_{gt}}$  are the ground truth and $\mathbf{R_{est},t_{est}}$ are the estimated values, respectively.  
\textcolor{black}{ 
Equation~\eqref{eq:1_exp_pose_fusion} does not have a physical unit although smaller is better and is a measure of better point cloud overlap.
%and readers need professional empirical experience to know the overall registration performance by comparing the numerical error value (scalar) and the registered point clouds' overlapping effect constantly in daily research life.
Equation~\eqref{eq:2_exp_pose_fusion} is the distance between the two coordinate systems' origins (the ground truth and estimated coordinate system) and its unit is meter. 
}
Both ways of the above methods evaluate the 6D pose's accuracy, although their error values and their error estimation methods are different.

\subsubsection{Ablation Study} \label{section:ablation_study_pose_fusion}

\begin{table}[H] 
	\caption{Model definition.} 
	\label{table: model_definition_ablation}
	\centering
	\begin{tabular}{ll}
		\toprule
		\textbf{Model Name} & \textbf{Strategy} \\
		\midrule
		Ours         & the proposed method in Section~\ref{section:pose-estimation} \\
		Ours-global  & disable the refined pose graph \\
					 & in Section~\ref{section:refined-global-graph-reconstruction}  \\ 
		Ours-prune   & disable rule 1: "Prune" \\
		Ours-update  & disable rule 2: "Update" \\
		Ours-gtdepth  & replace the depth from Sdfman-star \\
		& with the depth from GT depth\\ 
		\textcolor{black}{Ours-single-stereo} & \textcolor{black}{            input the stereo images from the} \\
		    & \textcolor{black}{front stereo camera only} \\
		\bottomrule
	\end{tabular}
\end{table}

We define six models to compare each proposed strategy's effectiveness. Table~\ref{table: model_definition_ablation} lists the defined model names and the strategies. 
The model "Ours" is the proposed method in Section~\ref{section:pose-estimation} with the point cloud input from Sdfman-star and is the baseline model, from which the other models are derived by changing only one strategy or factor. 
Model "Ours-global" disables the second-stage pose graph - the refined pose graph in Section~\ref{section:refined-global-graph-reconstruction}. 
Model "Ours-prune" disables rule 1 - "Prune" and thus will not prune any edge, no matter what the edge's reliability is.  
Model "Ours-update" disables rule 2 - "Update" and thus will not update the transformation matrix of each edge. 
Model "Ours-gtdepth" replaces the input point clouds from Sdfman-star with the point clouds from the ground truth depth. 
\textcolor{black}{Model "Ours-single-stereo" only inputs the stereo images from the front stereo camera (which consists of  image sensors Cam0 and Cam1) rather than the panoramic stereo images from the ring of synchronized stereo cameras.}

\begin{figure}
	\includegraphics[width=1\linewidth]{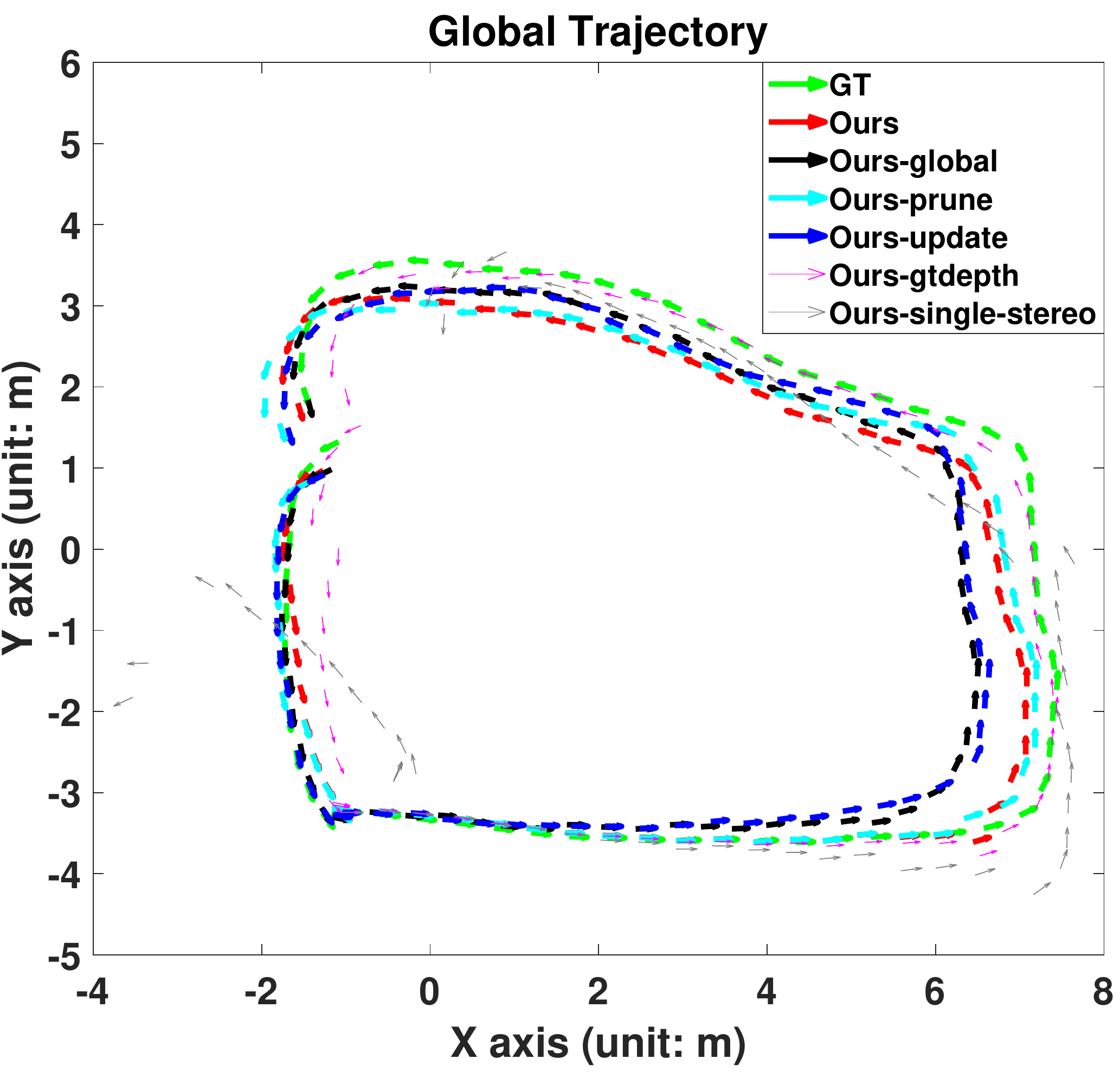}
	\caption{\textcolor{black}{The estimated trajectory using each model}}
	\label{fig:trajectory-ablation-study}
\end{figure}

For all models, the overlapping rate threshold $OL_{min} = 0.33$ and $OL_{max} = 0.35$. The 6D pose vector $\vec{v}_{th}$ is set to [0.4~m, 0.4~m, 0.4~m, $15^\circ$, $15^\circ$, $15^\circ$] ([translation on $X$ axis, translation on $Y$ axis, translation on $Z$ axis, roll, pitch, yaw]). 
Figure~\ref{fig:trajectory-ablation-study} shows the 2D trajectories from all the models when looking downward  from above at the whole garden. 

The trajectory of Ours-gtdepth is closest to the ground truth. The trajectories of Ours and Ours-prune are similar and rank $2^{nd}$ together. The remaining models perform worse. In particular \textcolor{black}{note that Ours-single-stereo did not work correctlyin the latter part of the global pose trajectory, with completely wrong pose estimates.} All the models except Ours-single-stereo have similar performance on the rotation factor, but perform on the translation factor variously.
Table~\ref{table:pose-fusion-ablation-study} shows the overall performance of each model by using the metrics in Equation~\eqref{eq:1_exp_pose_fusion} and Equation~\eqref{eq:2_exp_pose_fusion}. 
\textcolor{black}{
	\begin{table*}[h!]
		\centering
		\caption{Ablation study for pose fusion} \label{table:pose-fusion-ablation-study}
		\begin{tabular}{|p{1.7cm}|| p{1.7cm}| p{1.7cm}| p{1.7cm}| p{1.7cm}| p{2cm}|  p{2.5cm}|}
			\hline 
			\textit{Metric}  & Ours &  Ours-global & Ours-prune  & Ours-update & Ours-gtdepth & \textcolor{black}{Ours-single-stereo}  \\
			\hline 
			$\overline{E_R}$ & 0.11 & 0.08 & 0.12 & 0.08 & 0.08 & \textcolor{black}{0.98} \\ 
			\hline 
			$\sigma_{E_R}$ & 0.07 & 0.04 & 0.07 & 0.04 & 0.05 & \textcolor{black}{0.99} \\  
			\hline 
			$\overline{E_t}$  (m) &  0.33 & 0.48 & 0.34 & 0.52 & 0.27 & \textcolor{black}{2.47}\\ 
			\hline 
			$\sigma_{{E_t}}$  (m) & 0.18 & 0.32 & 0.20 & 0.29 & 0.16 & \textcolor{black}{1.99} \\ 
			\hline	
			$Time$  (s) & 233.24 & 210.86 & 233.14 & 233.15 & 230.41 & \textcolor{black}{179.28}\\ 
			\hline		
		\end{tabular} 	
	\end{table*}   
}
The running time of all the models are close (about 230 s) except Ours-global (210.86 s) and Ours-single-stereo (179.28 s). 
\textcolor{black}{From the quantitative aspect, it is obvious that Ours-single-stereo fails compared with the other models. Thus, using the panoramic stereo images from the ring of synchronized stereo vision cameras in the proposed framework is vital to overcome the challenges of the fast or large transformations between adjacent frames when a real robot navigates in a real outdoor environment. The reason is that the $360^\circ$ field of view makes the overlap between successive views high, which ensures the success of the global point cloud matching in the first stage of the pose fusion - global coarse pose graph optimization - to avoid the possibility of the whole 3D reconstruction framework collapsing. This strongly supports our major contribution (3)  because we are the first to combine the two-stage full-view-to-single-view global-coarse-to-local-fine pose graph optimization with a ring of synchronized stereo vision cameras simultaneously to handle the robot's fast movement in the real world.} 
The performance the other models (except Ours-single-stereo) on the rotation factor are similar (0.08 - 0.12) but the performance on the translation factor fluctuates  (0.27 m - 0.52 m).  
The translation accuracy of the model "Ours-global" and "Ours-update" is much lower than that of the model "Ours", which demonstrates the two-stage from-coarse-to-fine pose graph optimization and rule 2 - "Update" are effective. The accuracy of the model "Ours-prune" is slightly worse than that of the model "Ours" on both rotation and translation factors, which shows that rule 1 - "Prune" is effective. \textcolor{black}{Rule 1 "Prune" and rule 2 "Update" improve the performance because they make the transformation of the edge set $\tilde{E}$ more accurate and reliable which improves the constraint encoded in the loss function (see Equation~\eqref{eq: graph-optimization-local}). 
A more accurate constraint leads to to a more accurate pose estimate.} 
The accuracy of model "Ours-gtdepth" is better than that of the model "Ours", which demonstrates that better recovery of the input point clouds leads to better pose fusion accuracy. 
Thus, one topic for future work is to continue improving the accuracy of the input point clouds.

\subsubsection{Comparison with Existing Methods} \label{section: comparison_pose_fusion_module}
\begin{figure*}	
	\begin{center}
		\subfloat[Global Pose Trajectories from Different Algorithms]{\includegraphics[ width=0.75\linewidth]{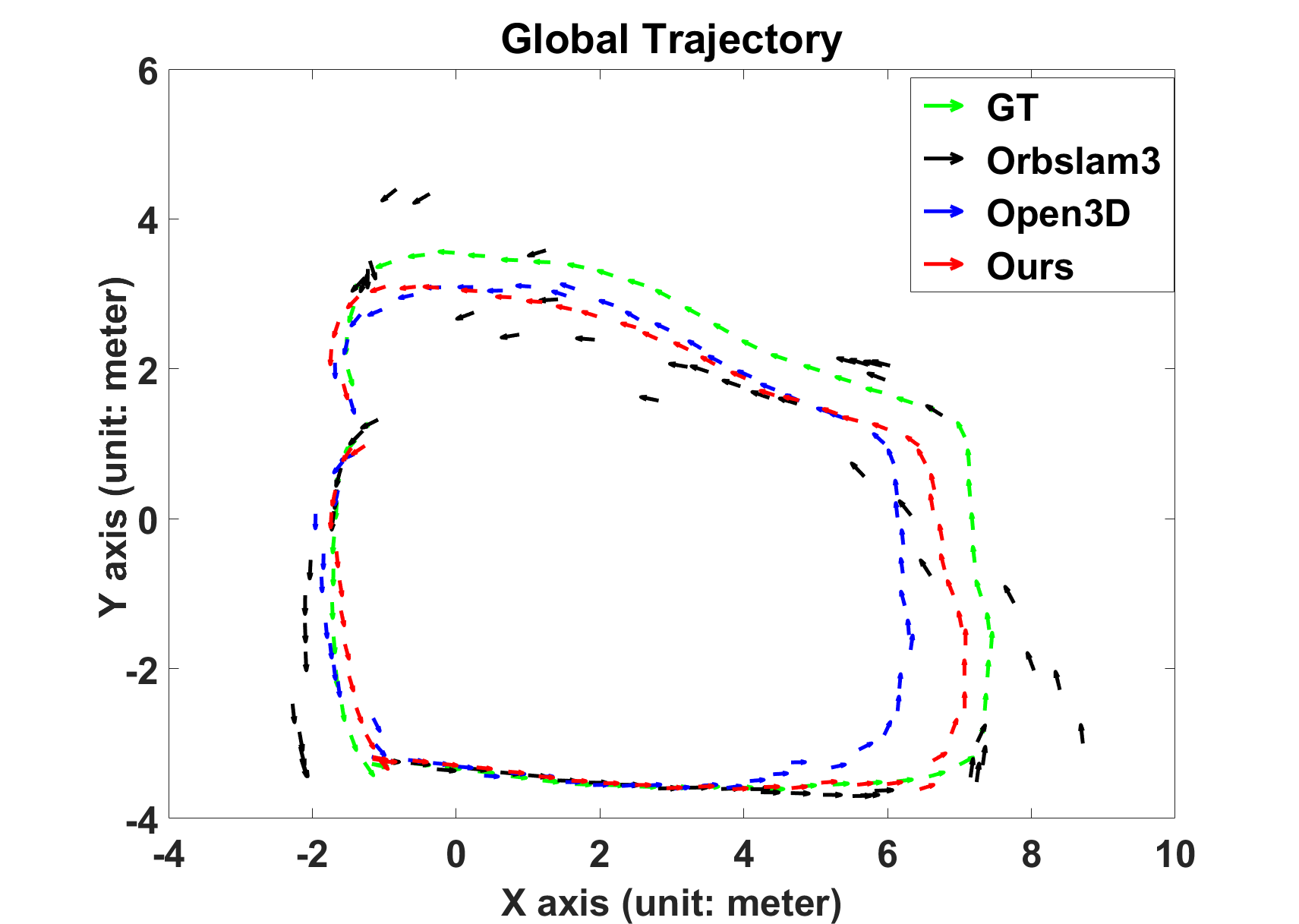}}
	\end{center}
	
	\begin{multicols}{2}
		
		\begin{center}
			%\flushleft
			\subfloat[Overall Rotation error]{\includegraphics[ width=1.12\linewidth]{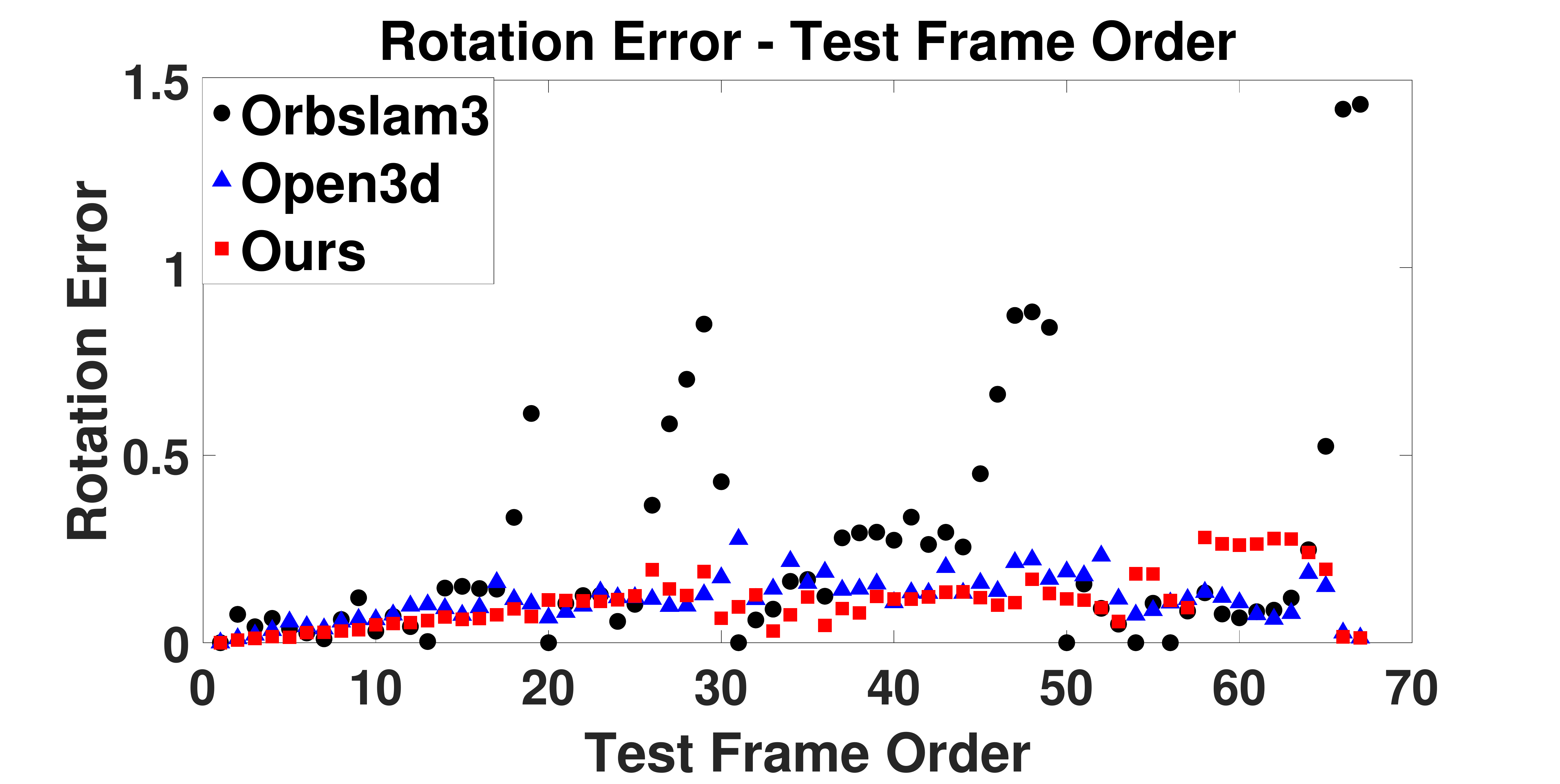}}
		\end{center}
		
		\begin{center}
			%\flushright
			\subfloat[Overall Translation error]{\includegraphics[ width=1.12\linewidth]{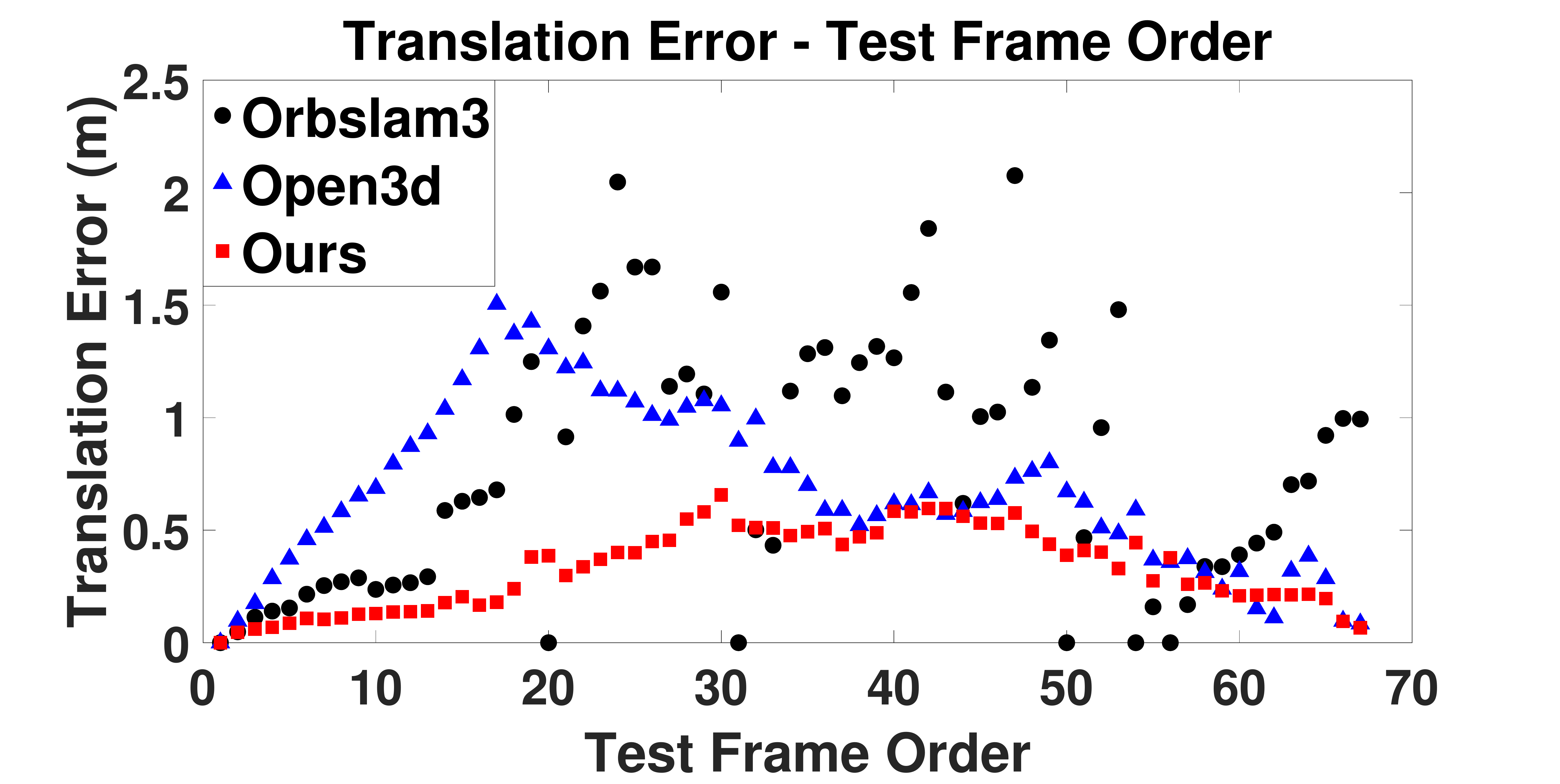}}
		\end{center}

	\end{multicols}
	
	\caption{Figure (a) shows the global pose trajectory of the sensor in the real garden; Figure (b), (c) show the overall rotation and translation error at each frame in the test dataset.}
	\label{fig:trajectory-comparison}
\end{figure*}

We compare the model "Ours" with existing state-of-the-art algorithms, represented by \textcolor{black}{the RGBD SLAM algorithm in Orbslam3~\citep{orbslam3}} and the reconstruction system in Open3d~\citep{Open3d} as available online\footnote{Orbslam3: \url{https://github.com/UZ-SLAMLab/ORB_SLAM3} and Open3D: \url{https://github.com/isl-org/Open3D}}. 
\textcolor{black}{The depth maps that all the algorithms receive as input are the output from the Sdfman-star fusion algorithm.}
The parameter setting in model "Ours" is the same as that in Section~\ref{section:ablation_study_pose_fusion}. 
\textcolor{black}{For Orbslam3, we set the number of features per image "ORBextractor.nFeatures" as 10000. 
The number of levels in the scale pyramid "ORBextractor.nLevels" is 15. 
The fast threshold "ORBextractor.iniThFAST" is 5 and "ORBextractor.minThFAST" is 3. 
The number of camera frames per second is 1. 
The rest of the parameters are the same as those in their released code.} \textcolor{black}{As the Orbslam3 framework does not support panoramic data, we input the RGB images and the corresponding depth maps from the image sensor `Cam0' into the RGBD SLAM algorithm in the Orbslam3 framework.}  
\textcolor{black}{To make Orbslam3 work better on the difficult dataset "Trimbot Wageningen SLAM Dataset", we additionally provide the ground truth pose to Orbslam3 when the adjacent frames have a large rotation and Orbslam3 lost tracking (at all the corners of the trajectory). 
More specifically, at Frames 20, 31, 50, 54, 56, we provide the corresponding ground truth pose to Orbslam3.
See Figure~\ref{fig:trajectory-comparison}~(b) and Figure~\ref{fig:trajectory-comparison}~(c) where both the rotation and translation error of Orbslam3 are equal to 0.} 
Open3D failed to work if we only input the single-view point clouds. 
To make Open3D perform better, we modified its initial code to make it use our full-view and single-view point clouds.
\textcolor{black}{We also provide the comparison results under the same conditions in Appendix~\ref{Appendix-fair-comparison} Fair Comparison With More Open-source Frameworks}. Readers can test their own code on the "Trimbot Wageningen SLAM dataset". 

Figure~\ref{fig:trajectory-comparison} (a) shows the estimated global trajectory from GT, Orbslam3, Open3D, and Ours. Figure~\ref{fig:trajectory-comparison} (b) and (c) show the rotation and translation error at each frame time in the test dataset using Equation~\eqref{eq:1_exp_pose_fusion} and Equation~\eqref{eq:2_exp_pose_fusion}. From Figure~\ref{fig:trajectory-comparison} we could see Open3D and Ours perform much more accurately and robustly than Orbslam3. Open3D and Ours have similar performance on the rotation and Ours performs more accurately than Open3D on the translation.

If we use the absolute difference between the ground truth and the estimated 6D vector to describe the 6D pose accuracy, Table~\ref{table:pose-fusion-accuracy-on-each-axis} compares the performance of the algorithms on each axis. 
Our approach's mean bias and the related standard deviation of the translation on the $x$, $y$, and $z$ axis are generally smaller than those of Orbslam3 and Open3D. 
The rotation performance of Ours and Open3D on $x$, $y$, and $z$ axis is more accurate and robust than that of Orbslam3. 
Our rotation performance on $x$, $y$, and $z$ axis is similar to that of Open3D.  

\begin{table}[h!]
	\centering
	\caption{Performance comparison of the algorithms on each axis} \label{table:pose-fusion-accuracy-on-each-axis}
	\begin{tabular}{|p{1.5cm}|| p{1.5cm}| p{1.5cm}| p{1.5cm}|}
		\hline 
		\textit{Metric}  &  \textit{Orbslam3} & \textit{Open3D}  & \textit{Ours}\\
		\hline 
		$\overline{tx}$  (m)  &   0.42 & 0.55 & \bf 0.18 \\
		\hline 
		$\sigma_{tx}$  (m) & 0.36 & 0.44 &  \bf  0.14 \\ 
		\hline 
		$\overline{ty}$  (m) & 0.44 & 0.21 &  \bf  0.20\\ 
		\hline 
		$\sigma_{ty}$  (m) &   0.50 &  \bf  0.15 &  0.18 \\
		\hline 
		$\overline{tz}$ (m) & 0.26 & 0.18 &  \bf  0.11 \\
		\hline 
		$\sigma_{tz}$  (m) & 0.35 & 0.14 &  \bf  0.08 \\
		\hline 
		$\overline{r}$  (deg) &   8.40 &  \bf  2.77 & 3.00 \\
		\hline 
		$\sigma_{r}$  (deg) & 12.28 &  \bf  1.95 & 3.34 \\
		\hline 
		$\overline{p}$  (deg) & 1.96 & 2.40 &  \bf  1.03 \\ 
		\hline 
		$\sigma_{p}$  (deg)  &   3.14 & 1.82 &  \bf 1.03 \\
		\hline 
		$\overline{y}$  (deg) & 3.50 & 2.12 &  \bf 1.91 \\
		\hline 
		$\sigma_{y}$  (deg) & 4.98 & 1.58 &  \bf  1.52 \\ 	
		\hline	
	\end{tabular} 	
\end{table}  

Table~\ref{table:pose-fusion-external-comparison} shows the overall performance of each algorithm using Equation~\eqref{eq:1_exp_pose_fusion} and Equation~\eqref{eq:2_exp_pose_fusion}. 
Ours performs best, although it increases the running time slightly. 
Given the bad performance of Orbslam3 on the real outdoor garden dataset, we will omit Orbslam3 in the following text and compare Ours with Open3D in Section~\ref{exp: volumetricfusion} "Volumetric Fusion Module" only.

\begin{table}[h!]
	\centering
	\caption{Overall performance comparison with external algorithms} \label{table:pose-fusion-external-comparison}
	\begin{tabular}{|p{1.5cm}|| p{1.5cm}| p{1.5cm}| p{1.5cm}| }
		\hline 
		\textit{Metric}    & Orbslam3 & Open3D  & Ours \\
		\hline 
		$\overline{E_R}$  & 0.25 &  0.12 & \bf 0.11 \\ 
		\hline 
		$\sigma_{E_R}$  & 0.31 & \bf 0.06 &  0.07 \\ 
		\hline 
		$\overline{E_t}$  (m)  &  0.78 & 0.68 & \bf 0.33 \\ 
		\hline 
		$\sigma_{{E_t}}$  (m)  &  0.57 & 0.37 & \bf 0.18 \\ 
		\hline	
		$Time$  (s)  & \bf 67.82 & 229.76 & 233.24 \\ 
		\hline		
	\end{tabular} 	
\end{table}

\subsection{Volumetric Fusion Module}  \label{exp: volumetricfusion}
In this part, we set the maximum depth for integrating as 5 meters. The size of TSDF (Truncated Signed Distance Field) cube is 10 meters. The length of each voxel is 0.01 m (1 cm). The truncation value for the signed distance function (SDF) is set to 0.06. 

\begin{figure*}
	\begin{center}
		\subfloat[Reconstructed 3D Garden]{\includegraphics[ width=\linewidth]{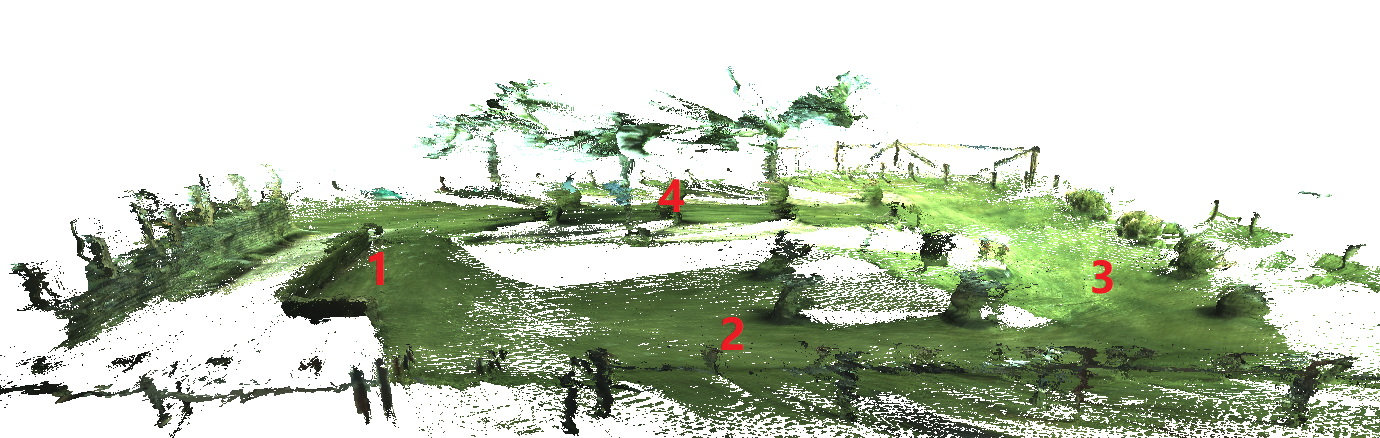}}
	\end{center}
	
	\begin{multicols}{2}
		
		\begin{center}
			%\flushleft
			\subfloat[Close-up shot 1]{\includegraphics[ width=0.96\linewidth]{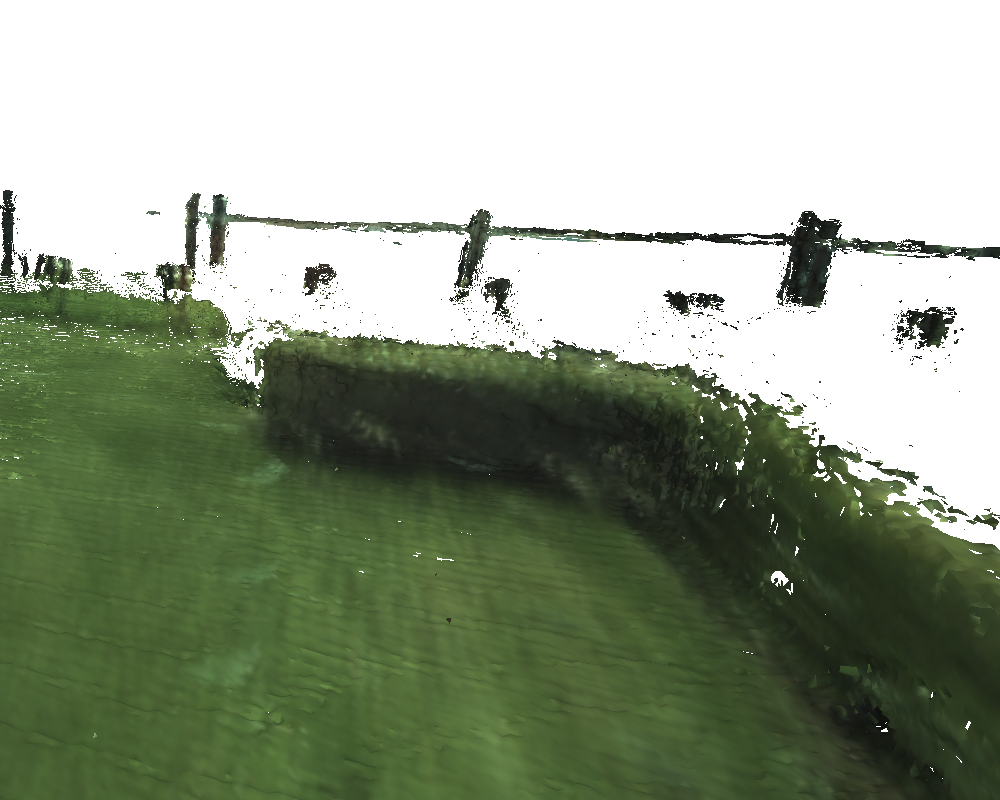}}
		\end{center}
		
		\begin{center}
			%\flushright
			\subfloat[Close-up shot 2]{\includegraphics[ width=0.96\linewidth]{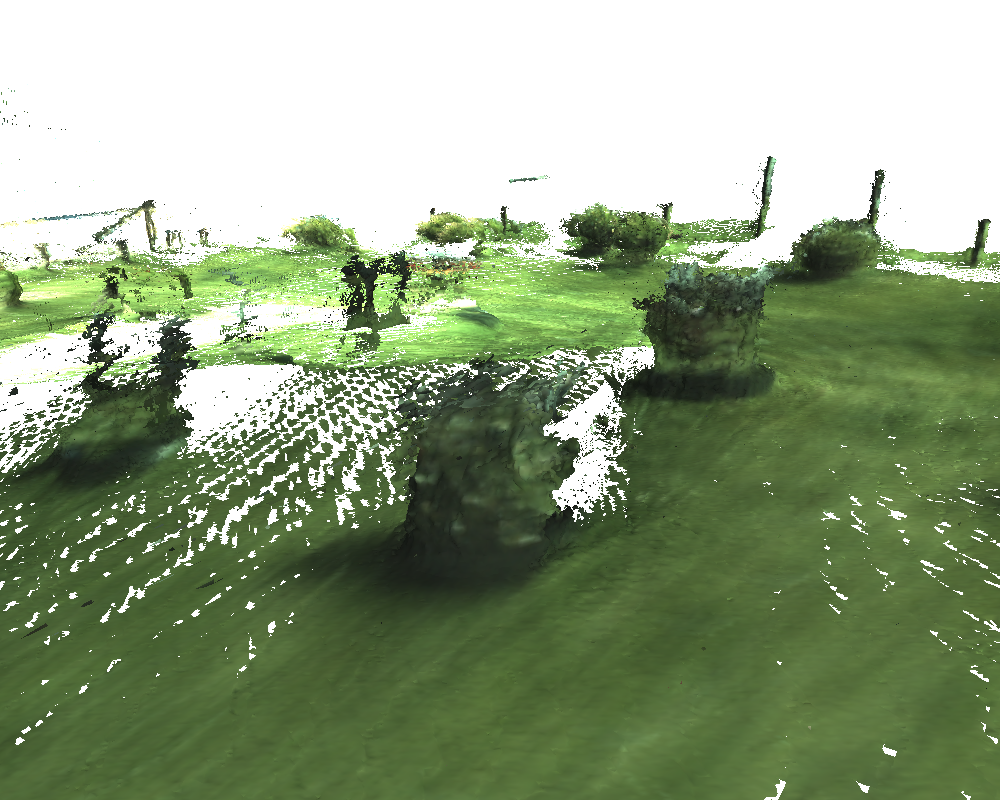}}
		\end{center}
	\end{multicols}

	\begin{multicols}{2}
	\begin{center}
		%\flushleft
		\subfloat[Close-up shot 3]{\includegraphics[ width=0.96\linewidth]{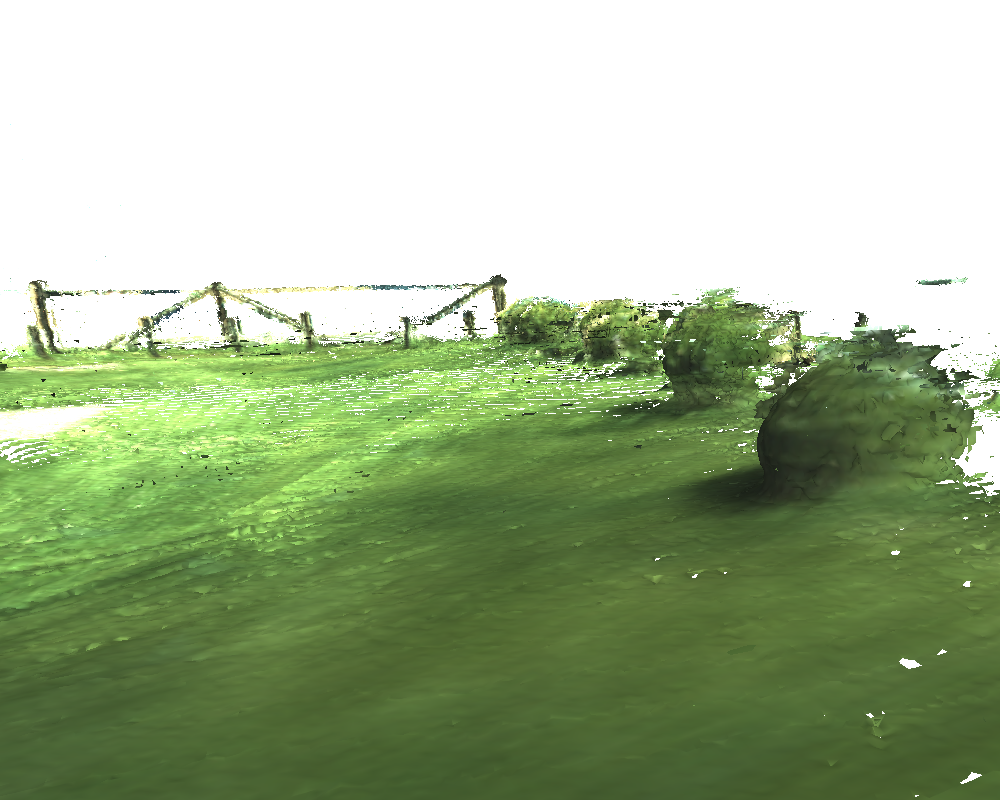}}
	\end{center}
	\begin{center}
		%\flushright
		\subfloat[Close-up shot 4]{\includegraphics[ width=0.96\linewidth]{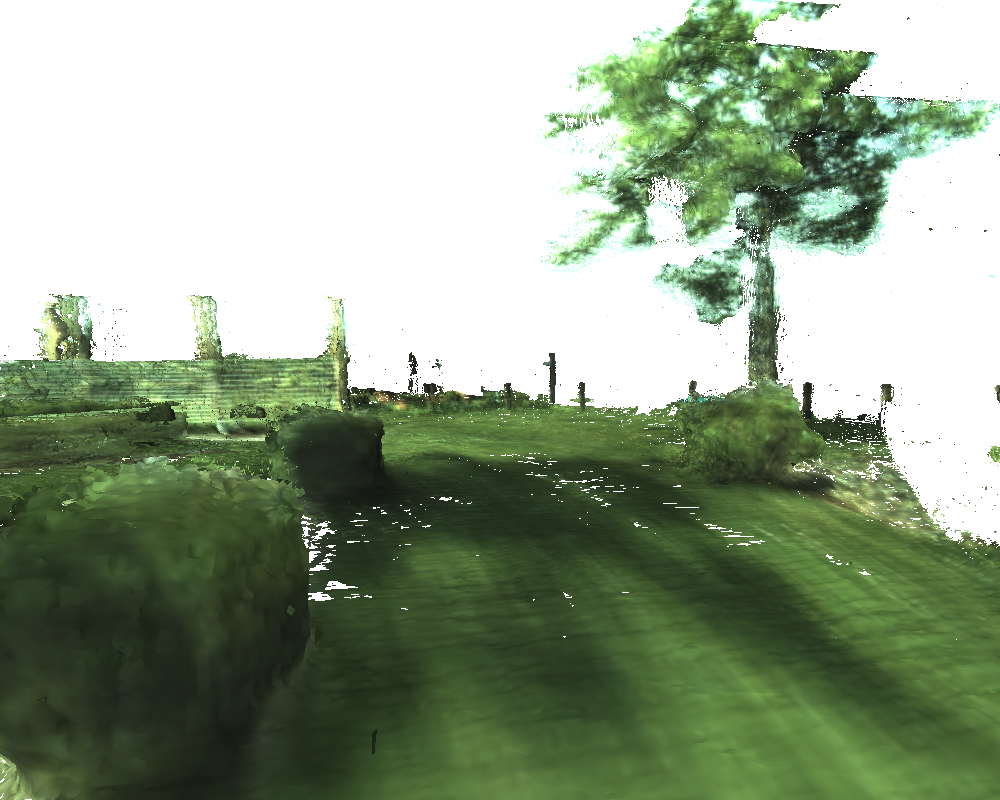}}
	\end{center}
	\end{multicols}

	\caption{Figure (a) shows the reconstructed 3D model of the real garden; Figure (b) - (e) shows close-up shots for sites 1 - 4. }
	\label{fig:reconstructed-garden}
\end{figure*}

Figure~\ref{fig:reconstructed-garden} shows the mesh of the reconstructed garden and its details at different sites. 
Figure~\ref{fig:reconstructed-garden} (a) shows the overview of the reconstructed whole garden. 
Figure~\ref{fig:garden-3D-model} (a) shows the ground truth. 
Figure~\ref{fig:reconstructed-garden} (b) (c) (d) (e) show close-up views at sites 1, 2, 3, 4 in Figure~\ref{fig:reconstructed-garden} (a). 
The white blank areas in all the figures are regions that have not been scanned during driving. 
These regions did not have target plants for the trimming robot and thus were not scanned. 
From the details, the reconstructed scene is good enough for the remote visualization and coarse robot task planning. 
A video that shows the reconstructed garden is at: \url{https://youtu.be/zGxcj0_NXCA}. 

In the following, the reconstructed gardens from all the algorithms are compared to the ground truth 3D model of the whole garden in the same world coordinate system. 
The evaluation method consists of estimating the mean and standard deviation
of the minimum distance between each point of the reconstructed garden and
its closest point in the ground truth garden model. Thus, this metric measures how close the
reconstructed garden is, on average, with respect to the ground truth garden model.

Table~\ref{table:reconstruction-accuracy} shows the mean $\overline{dist}$ and standard deviation $\sigma_{dist}$ of the minimum distance between the corresponding closest points. The mean and standard deviation of the minimum distance's absolute bias on $x$, $y$, and $z$ axis are ($\overline{dx}$, $\sigma_{dx}$), ($\overline{dy}$, $\sigma_{dy}$) and ($\overline{dz}$, $\sigma_{dz}$) respectively. The maximum of the minimum distance's absolute mean bias on all the three axes is $\overline{dz}$ (0.15 m) and the mean of the minimum distance $\overline{dist}$ is 0.18 m, which are good enough for the user's remote visualization and robot global task planning\footnote{Our trimming robot did coarse global task planning on the reconstructed global model first. When the robot arrives at the specific location for trimming, the robot arm will move the depth cameras on the robot arm to scan the target locally and build the accurate local 3D model with the precise pose update from the robot arm's joints.} on the reconstructed global model.

\begin{table}[h!]
	\centering
	\caption{Reconstruction Accuracy } \label{table:reconstruction-accuracy}
	\begin{tabular}{|p{2.5cm}|| p{2cm}| p{2cm}| }
		\hline 
		\textit{Metric}    & Open3D  & Ours \\
		\hline 
		$\overline{dx}$ (m)  & 0.06 & \bf 0.05 \\ 
		\hline 
		$\sigma_{dx}$  (m)  & \bf 0.09 &  \bf 0.09 \\ 
		\hline 
		$\overline{dy}$  (m)  &  \bf 0.05 &  \bf 0.05 \\ 
		\hline 
		$\sigma_{dy}$ (m)  & 0.09 & \bf 0.07 \\ 
		\hline 
		$\overline{dz}$ (m)  & 0.20 & \bf 0.15 \\ 
		\hline 
		$\sigma_{dz}$ (m)  & 0.18 & \bf 0.12 \\ 
		\hline 
		$\overline{dist}$ (m)  & 0.24 & \bf 0.18 \\ 
		\hline 
		$\sigma_{dist}$ (m)  & 0.19 & \bf 0.14 \\ 
		\hline	
		Time  /s  & 8.48 & \bf 8.23 \\ 
		\hline	
	\end{tabular} 	
\end{table}

\begin{figure*}
	\includegraphics[width=\linewidth]{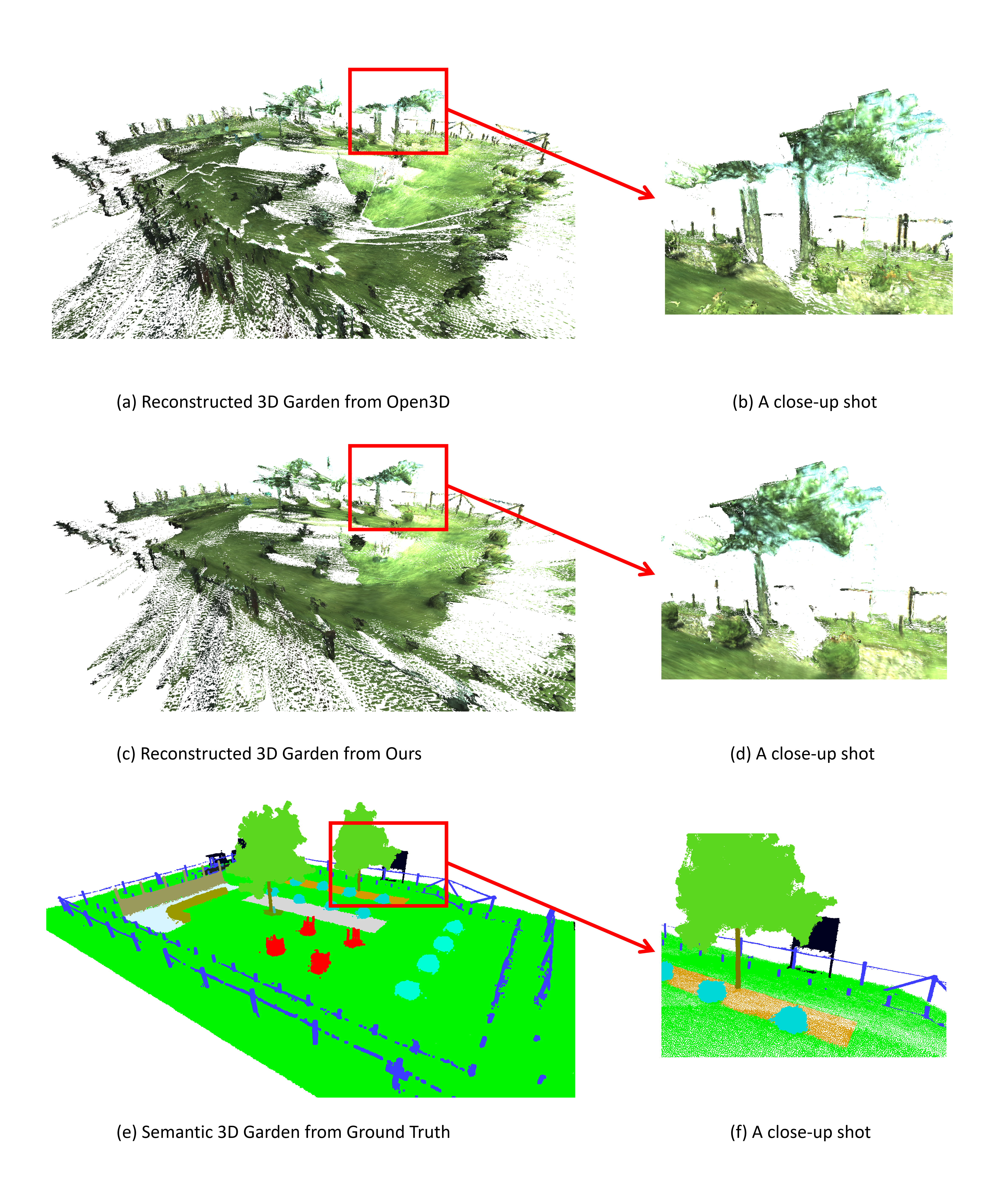}
	\caption{Comparison of the reconstructed garden from Open3d, Ours and GT.}
	\label{fig:reconstructed-garden-comparision}
\end{figure*}

Figure~\ref{fig:reconstructed-garden-comparision} (a) (c) (e) show the reconstructed gardens from Open3D, Ours, and the ground truth garden model. Figure~\ref{fig:reconstructed-garden-comparision} (b) (d) (f) show the details on the same site in the real garden. Compared with ours, we could see Open3D fails to align the point clouds of the same tree, and makes it seem that there were two trees on that site. 

\textcolor{black}{In addition to the experiments above, there are two other experiments. In the first experiment, the proposed framework is successfully tested with scene appearance and sunlight change. 
%The related results validate our proposed framework's robustness and accuracy again. 
More details can be found in Appendix~\ref{Appendix-robustness-test}. 
The second experiment compares our proposal with a popular commercial software application `ContextCapture' on the Trimbot Wageningen SLAM Dataset. The proposed approach again has better performance. 
For more details, see
Appendix~\ref{Appendix-contextcapture}.
}

\section{Conclusion and Discussion}\label{Conclusion}
This paper
\textcolor{black}{presented an improved approach for recovering accurate outdoor 3D scene reconstruction, based on \textcolor{black}{disparity} fusion, pose fusion and volumetric fusion, and demonstrated its performance by reconstructing a real outdoor garden containing a variety of different natural and man-made structures.}
 Avoiding the need for expensive and sparse Lidar scans, the proposed approach inputs the disparity maps from two different stereo vision algorithms into a \textcolor{black}{disparity} fusion network to produce accurate disparity maps, which is a cheap, accurate and robust solution to get higher quality depth data. 
 The depth data is converted into point clouds, \textcolor{black}{whose outliers are removed},  and then input into the pose fusion module. 
 The pose fusion module uses a two-stage from-global-coarse-to-local-fine pose graph optimization to estimate a more accurate global pose trajectory. 
 More specifically, in the first stage, we use fast global point cloud registration~\citep{Fast-global-registration} and full-view ($360^\circ$) point clouds to build a coarse global pose graph, which is robust to fast motion and big transformations between two consecutive frames. In the second stage, a local point cloud registration algorithm GICP~\citep{GICP} extended with three domain rules optimizes the refined pose graph, which produces a more accurate global pose trajectory. 
 With the accurate global pose trajectory and the fused depth maps, the mesh of the whole garden can be reconstructed by volumetric fusion, as demonstrated on a real outdoor dataset. 

The key to a good 3D reconstruction of the real garden is the accurate depth map and global pose trajectory. 
In future work, more advanced \textcolor{black}{disparity} fusion networks will be explored to continue to improve the \textcolor{black}{disparity} accuracy. 
The accuracy of the 6D pose that registers the point clouds affects the accuracy of the edges in the pose graph, which in turn influence the optimized global pose trajectory. 
More advanced and faster global and local point cloud registration algorithms that are robust against strong noise and large occlusions will be explored to get more accurate initial 6D pose estimation. 
\textcolor{black}{The  research proposed in this paper is adapted for use by a robot in a garden environment, but it can be generalized to different outdoor application scenarios. 
However, this requires the related ground truth for the network training and performance evaluation.
Expanding into related domains of robot applications is our priority for the near future.}   

% To print the credit authorship contribution details
\printcredits

\section*{Declaration of Competing Interest}
The authors declare that they have no known competing financial interests or personal relationships that could have appeared to influence the work reported in this paper.

\section*{Acknowledgements}
Before 2020, the research was funded by the TrimBot2020 project (Grant Agreement No. 688007,
URL: \url{http://trimbot2020.webhosting.rug.nl/}) from the European Union Horizon 2020 programme.
After 2020, the research funding is from Shenzhen Amigaga Technology Co. Ltd. by the Gagabot2022 project (Grant Agreement No. P987001), from the Human Resources and Social Security Administration of Shenzhen Municipality by Overseas High-Caliber Personnel project (Grant NO. 202102222X, Grant NO. 202107124X) and from Human Resources Bureau of Shenzhen Baoan District by High-Level Talents in Shenzhen Baoan project (Grant No. 20210400X, Grant No. 20210402X). We thank Gabriel Moreira from~\cite{Fast-optimization-PGO,Primal-dual-PGO} for the help with pose graph optimization. We thank all the partners from TrimBot2020 consortium for their help when we did this work, and for their contributions to test garden design, robot and sensor design and construction, data collection, and ground truthing, as well as many other contributions to the TrimBot2020 project.

\newpage

\appendix
\refstepcounter{section}
\section*{\textcolor{black}{Appendix A. List of Abbreviations}} \label{Abbreviations}
\noindent 
\textcolor{black}{
	\begin{tabular}{@{}ll}
        FOV & Field Of View\\
		GAN & Generative Adversarial Network \\
		HD & High Definition\\	
		ICP & Iterative Closest Point \\
		LC & Loop Closure \\
  	MPTEJI & Multi-stage Pose Trajectory Estimation \\
               & with Joint Information \\
		ORB & Oriented FAST and Rotated BRIEF \\
		PS-SLAM & Panoramic Stereo SLAM \\	
		SDF & Signed Distance Function\\
  SFM & Structure From Motion \\
		SIFT & Scale Invariant Feature Transform \\
  SLAM & 	Simultaneous Localization And Mapping \\	
		TOF & Time of Flight \\
		TSDF & Truncated Signed Distance Field\\
\end{tabular}}

\refstepcounter{section}
\section*{\textcolor{black}{Appendix B. List of Symbols}} \label{symbols}
\noindent
\begin{tabular}{@{}ll}
	$max\_dist$ & The maximum distance threshold\\ 
	$\phi_w$    & The width resolution ratio between the HD\\ 
	            & and the initial image\\            
	$\phi_h$    & The height resolution ratio between the HD\\
	            & and the initial image\\
	$N_{p}$     & The threshold number of the neighboring\\
	            & points in a sphere \\
	$radius$    & The radius of the sphere\\
	$N_n$       & The number of the points in the neighborhood\\ 
	$dist\_ratio$ & The distance ratio to remove the points\\
	$OL_{min}$ &  The minimum overlapping rate threshold \\
	$OL_{max}$ &  The maximum overlapping rate threshold \\
	$\vec{v}_{th}$ & 6D pose threshold in vector format \\
\end{tabular}

\color{black}

\refstepcounter{section}
\section*{\textcolor{black}{Appendix C. Formula derivation}} \label{Appendix-Formular-derivation}
\noindent 
\renewcommand\theequation{\Alph{section}.\arabic{equation}} 
\setcounter{equation}{0}
\subsection{\textcolor{black}{Loss Function}} \label{Appendix-Formular-derivation-loss-function}
\textcolor{black}{
   In this subsection, we will prove that Equation~\eqref{eq: graph-optimization-global} in this paper is equal to equation 4 in the paper~\citep{Fast-optimization-PGO} under the assumption that the uncertainty of the rotation is the same as the translation's. 
   Although `Equation 1' in the paper~\citep{Fast-optimization-PGO} is similar to our Equation~\eqref{eq: graph-optimization-global}, the authors~\citep{Fast-optimization-PGO} did not prove that the Frobenius-norm-based transformation difference loss function (Equation~\eqref{eq: graph-optimization-global}) is a special case of the maximum likelihood loss function in the pose graph optimization, which is the motivation for this section. }

\textcolor{black}{
   In Equation~\eqref{eq: graph-optimization-global}, $G(V, E)$ is a connected pose graph with $|V| = n$ poses (or vertices). 
   The rigid transformation $T_{ij}$ (here, computed using a point cloud registration algorithm) from the $i^{th}$ pose (denoted by $P_i$) to  the $j^{th}$ pose (denoted by $P_j$) could be written as $\{\tilde{\bold{R}}_{ij}, \tilde{t}_{ij}\}$ for the edge $(i,j) \in E$. 
   $E$ is the edge set. 
   $\tilde{\bold{R}}_{ij}$ and $\tilde{t}_{ij}$ are the corresponding relative rotation and translation estimates. 
   The $i^{th}$ pose  $P_i$ could be written as   $\{\bold{R}_{i}, t_{i}\}_{i=1,...,n}$. 
   In the following, we will use block-matrix notation to represent Equation~\eqref{eq: graph-optimization-global}.  
   \newline
   As the transformation is rigid, thus: 
   \newline
   $\tilde{\bold{R}}_{ij}\tilde{\bold{R}}_{ij}^T=\bold{I}$, $\bold{R}_{i}\bold{R}_{i}^T=\bold{I}$, $\bold{R}_{j}\bold{R}_{j}^T=\bold{I}$.
   \newline 
   Let us write:
   \newline
	$T_{ij} =
	   \left[\begin{array}{c:c}
	   	\tilde{\bold{R}}_{ij} & \tilde{t}_{ij}\\
	   	\hdashline
	   	\bold{0} & 1\\ 
	   \end{array}\right]$, 
	$P_{i} =
		\left[\begin{array}{c:c}
		\bold{R}_{i} & t_{i}\\
		\hdashline
		\bold{0} & 1\\ 
	\end{array}\right]$,
	\newline
	$P_{j} =
		\left[\begin{array}{c:c}
		\bold{R}_{j} & t_{j}\\
		\hdashline
		\bold{0} & 1\\ 
	\end{array}\right]$,
	$P_{j}^{-1} =
	\left[\begin{array}{c:c}
	\bold{R}_{j}^{T} & -\bold{R}_{j}^{T}t_{j}\\
	\hdashline
	\bold{0} & 1\\ 
	\end{array}\right]$;
   \newline
   Thus, Equation~\eqref{eq: graph-optimization-global} can be transformed into:
   \begin{equation}
	   \begin{split}
	   		&{\arg\min} \sum_{(i,j)\in E} \left\Vert T_{ij} - P_{i}P_{j}^{-1} \right\Vert_{F}^2 \\
	   	  = &{\arg\min} \sum_{(i,j)\in E} \left\Vert 
	  	    \left[\begin{array}{c:c}
			   	  \tilde{\bold{R}}_{ij} - \bold{R}_i\bold{R}_j^T & \tilde{t}_{ij} - t_i + \bold{R}_i\bold{R}_j^Tt_j \\
			   	  \hdashline
			   	  \bold{0} & 0\\ 
	   	    \end{array}\right] 
	   	    \right\Vert_{F}^2 \\
 	   	  = &{\arg\min} \sum_{(i,j)\in E} (\left\Vert \tilde{\bold{R}}_{ij} - \bold{R}_i\bold{R}_j^T   \right\Vert_{F}^2
 	   	     + \left\Vert \tilde{t}_{ij} - t_i + \bold{R}_i\bold{R}_j^Tt_j  \right\Vert_{F}^2)
	   \end{split}
	   \label{eq: graph-optimization-global-ap1}
	\end{equation} 		
	According to the definition, Frobenius norm of a matrix $\bold{A}$ is defined as the square root of the sum of the absolute squares of its elements in the matrix, which is equal to the square root of the matrix trace of $\bold{A} \bold{A}^T$. Additionally, $tr(A) = tr(A^T)$.
	\newline
	Expanding the term $\left\Vert \tilde{\bold{R}}_{ij} - \bold{R}_i\bold{R}_j^T   \right\Vert_{F}^2$ in Equation~\eqref{eq: graph-optimization-global-ap1}:  
	\begin{equation}
		\begin{split}
			& \left\Vert \tilde{\bold{R}}_{ij} - \bold{R}_i\bold{R}_j^T   \right\Vert_{F}^2 \\
		  = & tr\{ (\tilde{\bold{R}}_{ij} - \bold{R}_i\bold{R}_j^T)(\tilde{\bold{R}}_{ij} - \bold{R}_i\bold{R}_j^T)^T \} \\
		  = & tr(\tilde{\bold{R}}_{ij}\tilde{\bold{R}}_{ij}^T + \bold{R}_i\bold{R}_j^T\bold{R}_j\bold{R}_i^T
	          - \tilde{\bold{R}}_{ij}\bold{R}_j\bold{R}_i^T - \bold{R}_i\bold{R}_j^T\tilde{\bold{R}}_{ij}^T) \\
	      = & tr(\bold{I} + \bold{I}) -tr(\tilde{\bold{R}}_{ij}\bold{R}_j\bold{R}_i^T) 
	          - tr\{(\tilde{\bold{R}}_{ij}\bold{R}_j\bold{R}_i^T)^T\} \\
	      = & 6 - 2tr(\tilde{\bold{R}}_{ij}\bold{R}_j\bold{R}_i^T) \\
		 \end{split}
		 \label{eq: graph-optimization-global-ap1-term}
	\end{equation}
	\newline
	Substitute the term $\left\Vert \tilde{\bold{R}}_{ij} - \bold{R}_i\bold{R}_j^T   \right\Vert_{F}^2$ in Equation~\eqref{eq: graph-optimization-global-ap1} with Equation~\eqref{eq: graph-optimization-global-ap1-term} and neglect the constant term:
   \begin{equation}
	\begin{split}
		&{\arg\min} \sum_{(i,j)\in E} \left\Vert T_{ij} - P_{i}P_{j}^{-1} \right\Vert_{F}^2 \\
		= & {\arg\textcolor{green}{\min}} \sum_{(i,j)\in E} \{ \left\Vert \tilde{t}_{ij} - t_i + \bold{R}_i\bold{R}_j^Tt_j  \right\Vert_{F}^2
		    - 2tr(\tilde{\bold{R}}_{ij}\bold{R}_j\bold{R}_i^T) \}
		\end{split}
		\label{eq: graph-optimization-global-ap2}
	\end{equation}	
	In Equation~\eqref{eq: graph-optimization-global-ap2} we minimize the loss function to get the estimated rotation
and translation by maximizing its negative. Thus, divide the right part of the equal sign by the negative constant $-2\sigma_R^2$, we get:
   \begin{equation}
		\begin{split}
		&{\arg\min} \sum_{(i,j)\in E} \left\Vert T_{ij} - P_{i}P_{j}^{-1} \right\Vert_{F}^2 \\
		= & {\arg\textcolor{green}{\max}} \{ -\frac{1}{2\textcolor{blue}{\sigma_R^2}}\sum_{(i,j)\in E} \left\Vert \tilde{t}_{ij} - t_i + \bold{R}_i\bold{R}_j^Tt_j  \right\Vert_{F}^2 \\
		  & + \frac{1}{\sigma_R^2} \sum_{(i,j)\in E} tr(\tilde{\bold{R}}_{ij}\bold{R}_j\bold{R}_i^T) \}
		\end{split}
		\label{eq: graph-optimization-global-ap3}
	\end{equation}	
	\newline
	Compare our term Equation~\eqref{eq: graph-optimization-global-ap3} with the log-likelihood term equation 4 in the paper~\citep{Fast-optimization-PGO}, which is shown in the following Equation~\eqref{eq: graph-optimization-global-ap-ref}:
   \begin{equation}
	\begin{split}
		& log L(\theta | y) \\
		= &  -\frac{1}{2\textcolor{blue}{\sigma_t^2}}\sum_{(i,j)\in E} \left\Vert \tilde{t}_{ij} - t_i + \bold{R}_i\bold{R}_j^Tt_j  \right\Vert_{F}^2 \\
		& + \frac{1}{\sigma_R^2} \sum_{(i,j)\in E} tr(\tilde{\bold{R}}_{ij}\bold{R}_j\bold{R}_i^T)
		\end{split}
		\label{eq: graph-optimization-global-ap-ref}
	\end{equation}
	\newline
	When the noise level for the rotation and translation in the paper~\citep{Fast-optimization-PGO} is assumed to be equal (i.e. $\sigma_t = \sigma_R$), Equation~\eqref{eq: graph-optimization-global-ap3} (in our paper) and Equation~\eqref{eq: graph-optimization-global-ap-ref} (which is same with equation 4 in~\cite{Fast-optimization-PGO}) are completely the same. Thus, we could use the optimization method\footnote{The URL of the released code: \url{https://github.com/gabmoreira/maks}} in ~\cite{Fast-optimization-PGO} to optimize our error function. Finding the optimum rotation parameters first and then solving for the translation parameters turns it into a least-squares problem. 
 To deduce the Equation~\eqref{eq: graph-optimization-global-ap-ref}, refer to Page 9 - 10 in (URL: \url{https://drive.google.com/file/d/1ML7mkLSIALm3x5DtID7ozHD3YL7S1iNC/view?usp=sharing}) or the most related papers~\citep{carlone2015initialization,carlone2015lagrangian,Fast-optimization-PGO,Primal-dual-PGO}. 
	\newline
	To conclude, Equation~\eqref{eq: graph-optimization-global} can be turned into a maximum likelihood estimation problem under the assumption of the proper noise level for rotation and translation. From another aspect, there is a more intuitive way to express the physical meaning of Equation ~\eqref{eq: graph-optimization-global}. That is: estimate the pose of each node accurately, which in turn makes the existing relative pose measurements between different nodes closer to the post-calculated relative pose between different nodes based on their estimated global pose.  
}

\subsection{\textcolor{black}{Maximum Distance}} \label{Appendix-Formular-derivation-maximum-distance}

\textcolor{black}{In the initial work Sdf-man~\citep{sdf-man}, at the end of the refiner network (see Figure 2 on page 7 in~\cite{sdf-man}) the method uses the function `tanh' to output an intermediate map $w$ and each value in $w$ is in $(-1,1)$. 
\begin{equation}
	initial\_disp = \frac{max\_disp \cdot (w + 1)}{2}
	\label{eq:initial-convert1-maximum-distance-appendix}
\end{equation}	
Then it uses Equation~\eqref{eq:initial-convert1-maximum-distance-appendix} to convert the intermediate map $w$ to the disparity map $initial\_disp$. $max\_disp$ is the maximum disparity threshold. 
Converting the disparity map $initial\_disp$ into a depth map using Equation~\eqref{eq: eq2-post-processing} gives Equation~\eqref{eq:initial-convert2-maximum-distance-appendix}.
\begin{equation}
	initial\_depth = \frac{2fb}{max\_disp \cdot (w+1)}
	\label{eq:initial-convert2-maximum-distance-appendix}
\end{equation}
 $f$ and $b$ are the focal length value and baseline value of the stereo vision camera. 
 As $w$ ranges from -1 to 1, the $initial\_depth$ values will range from $\frac{fb}{max\_disp}$ to $+\infty$.}

\textcolor{black}{
In this paper, the difference is that we use a new  Equation~\eqref{eq:initial-convert3-maximum-distance-appendix} to map the intermediate map $w$ to the new disparity map $new\_disp$ rather than Equation~\eqref{eq:initial-convert1-maximum-distance-appendix}.
\begin{equation}
	new\_disp = \frac{2fb}{max\_dist \cdot (w+1)}
	\label{eq:initial-convert3-maximum-distance-appendix}
\end{equation}
$max\_dist$ is set as the maximum distance threshold of interest. 
Converting the new disparity map $new\_disp$ into the depth map format using Equation~\eqref{eq: eq2-post-processing} gives Equation~\eqref{eq:initial-convert4-maximum-distance-appendix}, whose value domain is $(0, max\_dist)$.
\begin{equation}
	new\_depth = \frac{max\_dist \cdot (w + 1)}{2}
	\label{eq:initial-convert4-maximum-distance-appendix}
\end{equation}
}

\textcolor{black}{
Comparing the value domain of $new\_depth$ in Equation~\eqref{eq:initial-convert4-maximum-distance-appendix} and $initial\_depth$ Equation~\eqref{eq:initial-convert2-maximum-distance-appendix}, the domain $(\frac{fb}{max\_disp}, +\infty)$ of $initial\_depth$ is larger than the domain $(0, max\_dist)$ of $new\_depth$ considerably, though their definition domains are the same. Thus, as for noise with the same granularity in the input, $new\_depth$ from the proposed strategy will output a more robust and accurate result. By setting the maximum distance threshold of interest and the new mapping function Equation~\eqref{eq:initial-convert3-maximum-distance-appendix}, we effectively narrow the value domain of the depth output to increase depth accuracy. This alternative approach has also been confirmed by the experiment results in Section~\ref{exp: depthfusion}}.

\refstepcounter{section}
\section*{Appendix D. More Details about Data Collection}  \label{Appendix_more_description_about_dataset}
% \appendix
\renewcommand\thefigure{\Alph{section}.\arabic{figure}} 
\renewcommand\thetable{\Alph{section}.\arabic{table}}

% \section{\textcolor{black}{Ablation Study}}
\setcounter{figure}{0}    
\setcounter{table}{0}

\begin{figure}
	\includegraphics[width=\linewidth]{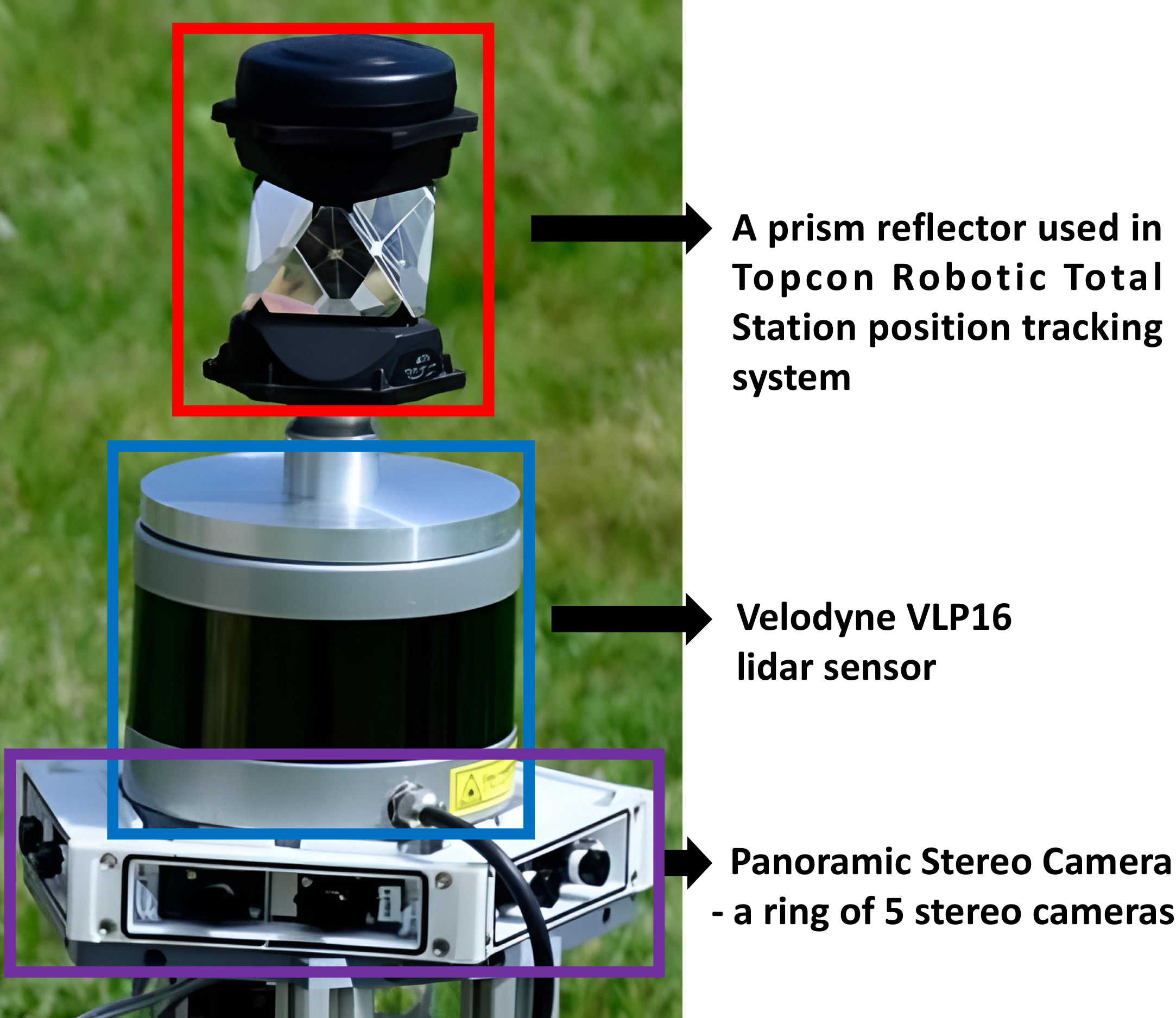}
	\caption{\textcolor{black}{Some sensors for collecting the data}}
	\label{fig:sensor-set-appendix}
\end{figure}

\textcolor{black}{As Figure~\ref{fig:garden-3D-model} (b) shows, the trimming robot Trimbot\footnote{Trimbot's hardware was mainly developed by our Trimbot2020 consortium member Robert Bosch GmbH based on the Bosch Indigo lawn mower.} navigates around the outdoor garden\footnote{The garden was constructed by our Trimbot2020 consortium member Wageningen Research in Netherlands.} to collect the raw data. The top of the robot resembles a `tower' (see Figure~\ref{fig:sensor-set-appendix}), which consists of a prism retroreflector, a Velodyne VLP16, and the panoramic stereo camera (a ring of 5 stereo cameras). }

\begin{figure}
	\includegraphics[width=\linewidth]{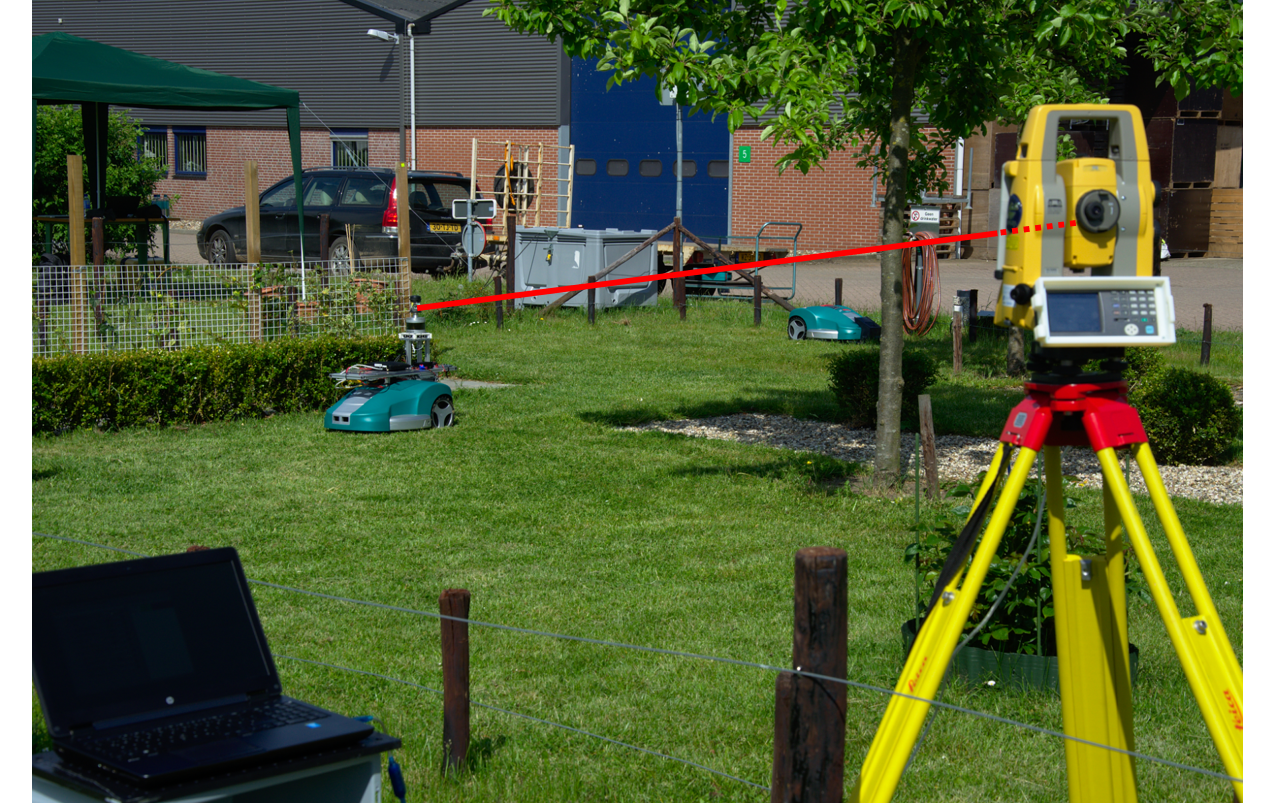}
	\caption{\textcolor{black}{Topcon PS Series Robotic Total Station estimated the robot's 3D position by emitting laser beams and receiving the  beams reflected by he prism reflector on the top of the trimming robot Trimbot.}}
	\label{fig:topcon-appendix}
\end{figure}

\textcolor{black}{The prism reflector on the robot is used to reflect the laser beam from Topcon PS Series Robotic Total Station (See Figure~\ref{fig:topcon-appendix}), to estimate the robot's position in 3D space. 
According to the datasheet of the Topcon PS Series Robotic Total Station\footnote{Topcon PS Series Robotic Total Station: \url{https://drive.google.com/file/d/1Z54jDM0fkqhNgYq1SBsZRFX-XG14alIC/view?usp=sharing}}, the distance measurement accuracy with the prism could be down to $1.5 mm + 2 ppm$. 
To estimate the robot's orientation [roll, pitch, yaw], a STIM300 IMU sensor inside the trimming robot is used to record the acceleration and rotation rate measurements. 
According to the datasheet of STIM300 IMU sensor\footnote{STIM300 IMU: \url{https://drive.google.com/file/d/1PhFtSSABCs0msnu2Gwza0EhAg4mpUZar/view?usp=sharing}}, the gyroscope input range is $\pm400 deg/sec$ and its angular random walk is $0.15 deg/\sqrt{hr}$. The accelerometer range is $\pm10 g$ and its velocity random walk is $0.06 m/s/\sqrt{hr}$. The in-run bias stability of the gyroscope and accelerometer is $0.5 deg/hr$ and, $0.05 mg$ respectively. Based on the precise measurements from the STIM300 IMU sensor, the orientation of the robot is estimated in an offline post-processing step by strap-down integration. The 3D position $[tx, ty, tz]$ from Topcon PS Series Robotic Total Station and the orientation $(roll, pitch, yaw)$ are appended to constitute the 6-DoF pose of the robot. Then, by calibration, the 6-DoF pose of the robot is transformed to get the pose of each rigidly placed image sensor in the panoramic stereo camera. 
Finally, structure-from-motion~\citep{SFM} is used to refine each image sensor's pose to form the ground truth pose of each image sensor, particularly to fix poses where the line of sight between the Topcon and prism was interrupted by obstacles.}

\textcolor{black}{The Velodyne VLP16 lidar sensor is mounted on top of the panoramic stereo camera to record a reference point cloud from the Trimbot robot's  perspective. 
The lidar sensor has a $360^\circ$ horizontal field of view with an angle resolution spanning from $0.1^\circ$ to $0.4^\circ$, which corresponds to the rotation rate from 5 Hz to 20 Hz. In the Trimbot Wageningen SLAM Dataset, the angular resolution is set to $0.2^\circ$ and the rotation rate is set to $10 Hz$. The Velodyne VLP16 lidar sensor has 16 horizontal rays, which are  distributed within a vertical field of view of $\pm15^\circ$. According to its datasheet\footnote{Velodyne VLP16: \url{https://drive.google.com/file/d/1ZYrYHf7wqI5PjuuKpxO7SPzDfmb33b1Y/view?usp=sharing}}, its scanning range could be up to 100 m with an accuracy of $\pm3 cm$. Then the lidar point cloud is projected to the image planes of the 10 image sensors in the panoramic stereo camera to form sparse depth maps.}

\begin{figure}
	\centering

	\subfloat[Top View]{\includegraphics[ width=1\linewidth]{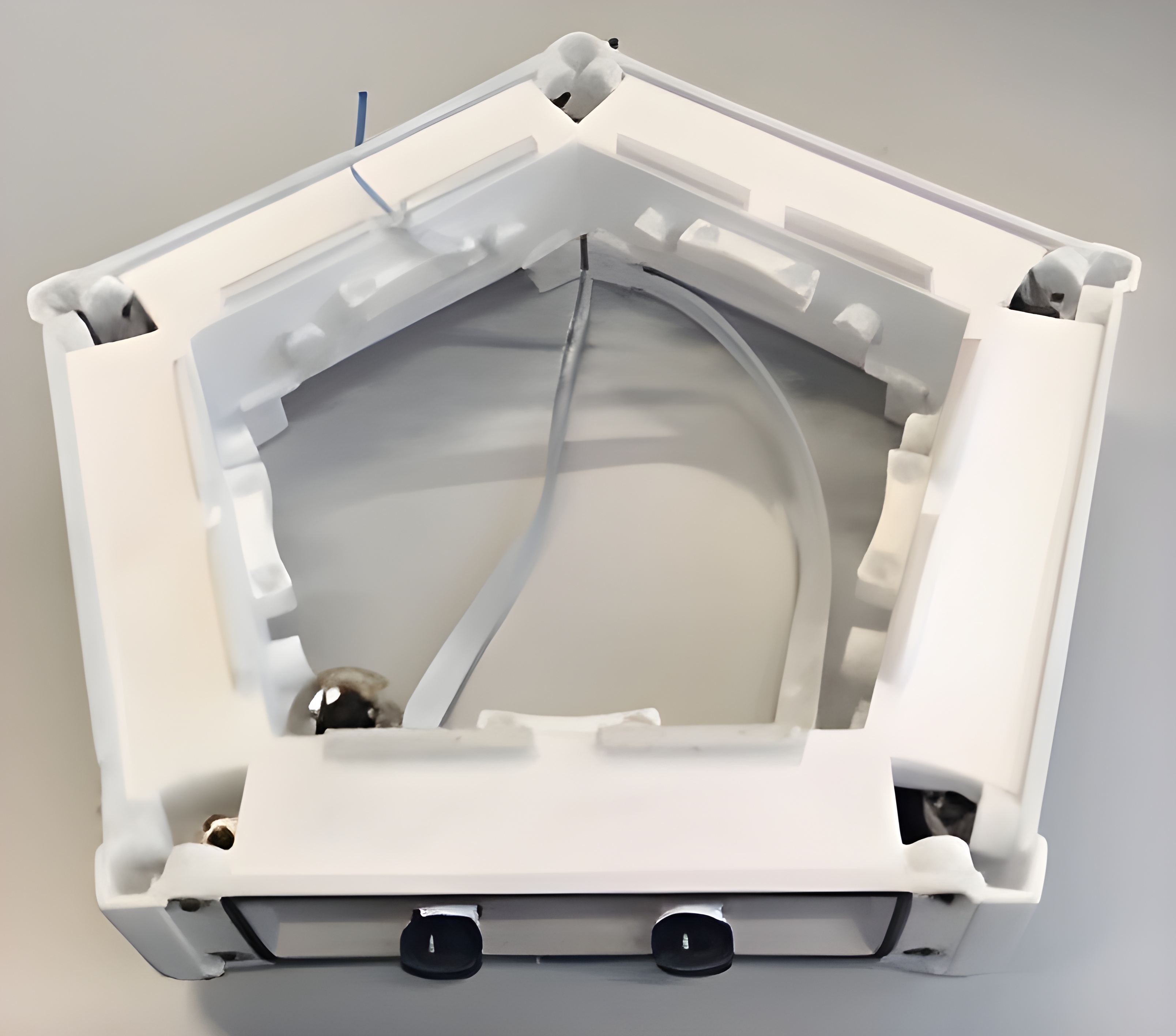}} \\
	\subfloat[Side View]{\includegraphics[ width=1\linewidth]{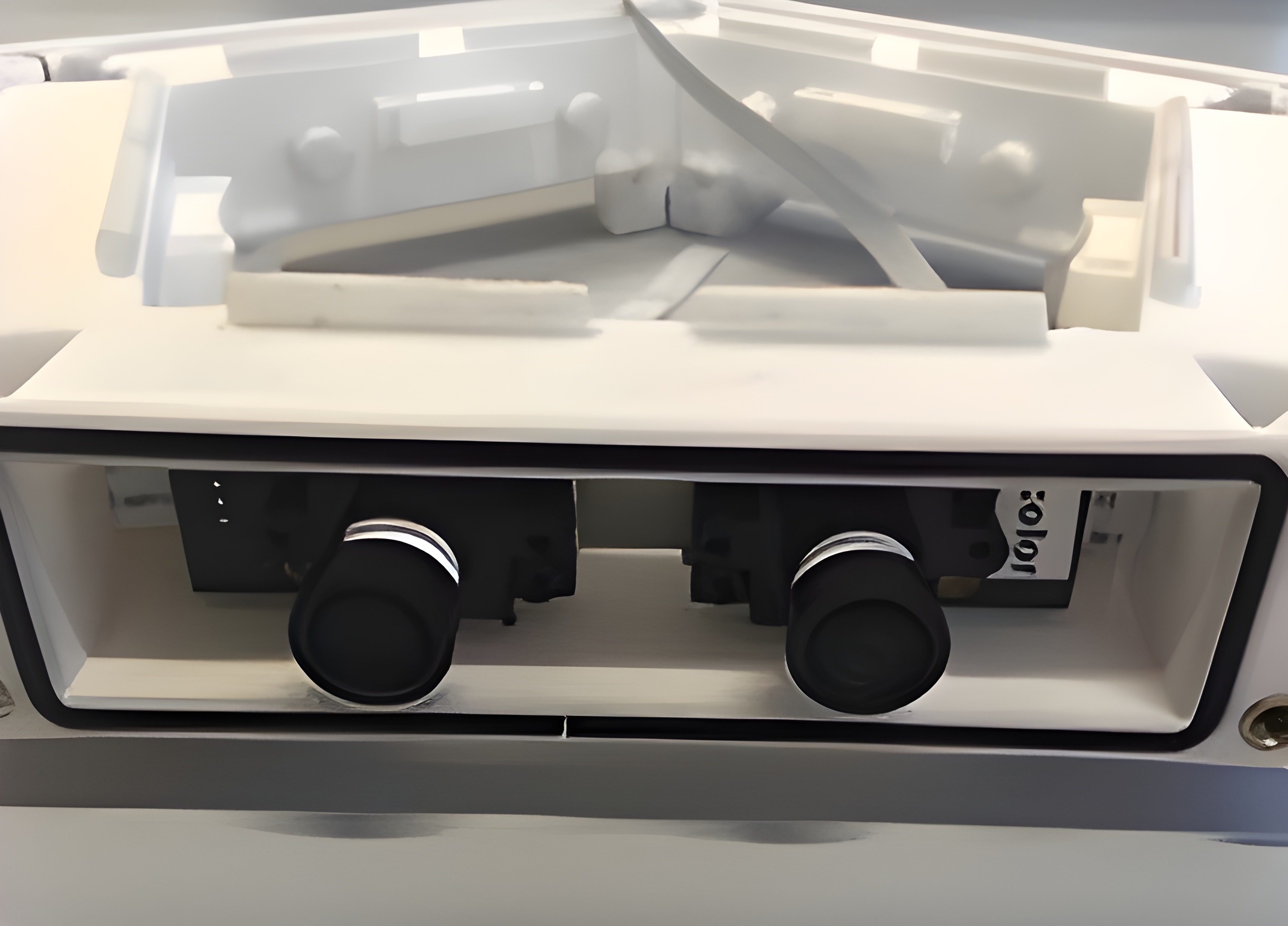}}

	\caption{\textcolor{black}{Figure (a) (b) shows the panoramic stereo camera from the top view and side view respectively. The images are from ETH Zürich (Trimbot2020 consortium member).}}
	\label{fig:panoramic-stereo-cameras-appendix}
\end{figure}

\textcolor{black}{The panoramic stereo camera\footnote{The panoramic stereo camera hardware was developed by our Trimbot2020 consortium member ETH Zürich based on their previous work~\citep{FPGA_Stereo_dominik}.} is built with ten MT9V024
CMOS image sensors from ONSemiconductors\footnote{MT9V024 datasheet URL:\url{https://github.com/Canpu999/Trimbot-Wageningen-SLAM-Dataset/blob/main/Image-sensor-MT9V024-datasheet-ON_Semiconductor.pdf}}. The housing of the panoramic stereo camera has a pentagon shape and is manufactured by 3D printing. Figure~\ref{fig:panoramic-stereo-cameras-appendix} shows the panoramic stereo camera from the top and side views. The image sensors have $752 \times 480$ ($Horizontal \times Vertical$) pixels resolution with global shutter. The maximum frame rate of a single image sensor could be up to 60 FPS at full resolution. The five synchronized stereo vision cameras (10 image sensors) could only stream the synchronized panoramic stereo images (10 synchronized images) at 12 HZ because it is limited by the bandwidth of the panoramic stereo camera's data bus. The image sensor's operating temperature ranges from $-40 ^{\circ}C$ to $+100 ^{\circ}C$ ambient. The active imager size is $4.51 mm (H) \times 2.88 mm (V)$, whose diagonal size is 5.35 mm. The pixel size is 6.0 $\mu$m $\times$ 6.0 $\mu$m. We calibrated the image sensor using the \textit{Kalibr} package\footnote{\url{https://github.com/ethz-asl/kalibr}} by setting the camera model as pinhole and the distortion model as radial-tangential. 
\textcolor{black}{The calibration accuracy of image sensors in terms of a mean reprojection error was 0.00 pixels on the X and Y axis, with standard deviation ranging from 0.1 pixels to 0.2 pixels.} 
Given that the image sensors' temperature and long operating time influences the robustness of the commercial hardware's imaging quality, which will possibly influence the calibration accuracy, more engineering robustness tests would be future work.}

\begin{figure*}
	
	\begin{multicols}{2}
		
		\begin{center}
			%\flushleft
			\subfloat[The point cloud from view 1 with colored height]{\includegraphics[ width=0.92\linewidth]{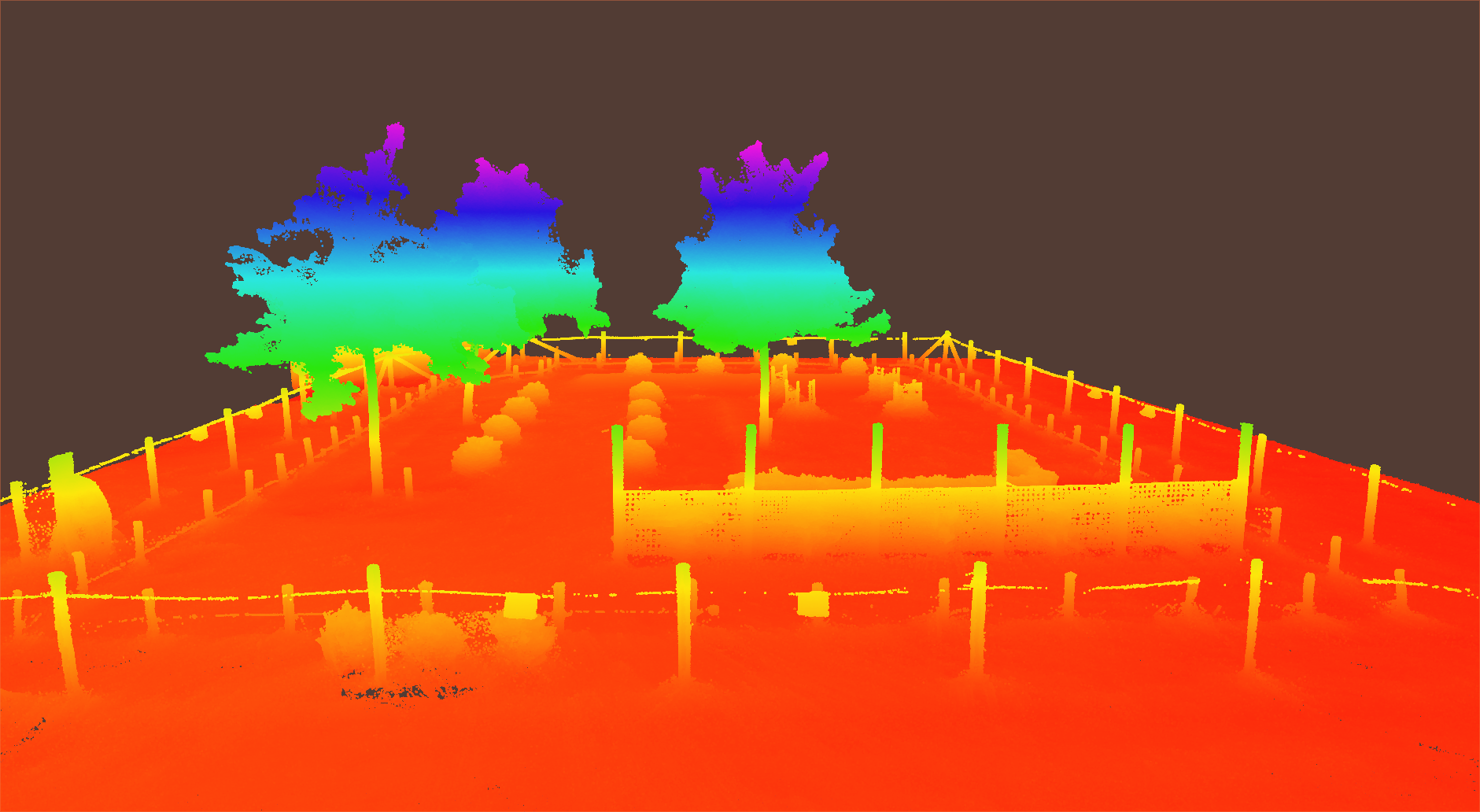}}
		\end{center}
		
		\begin{center}
			%\flushright
			\subfloat[The point cloud from view 1 with registered RGB data]{\includegraphics[ width=0.92\linewidth]{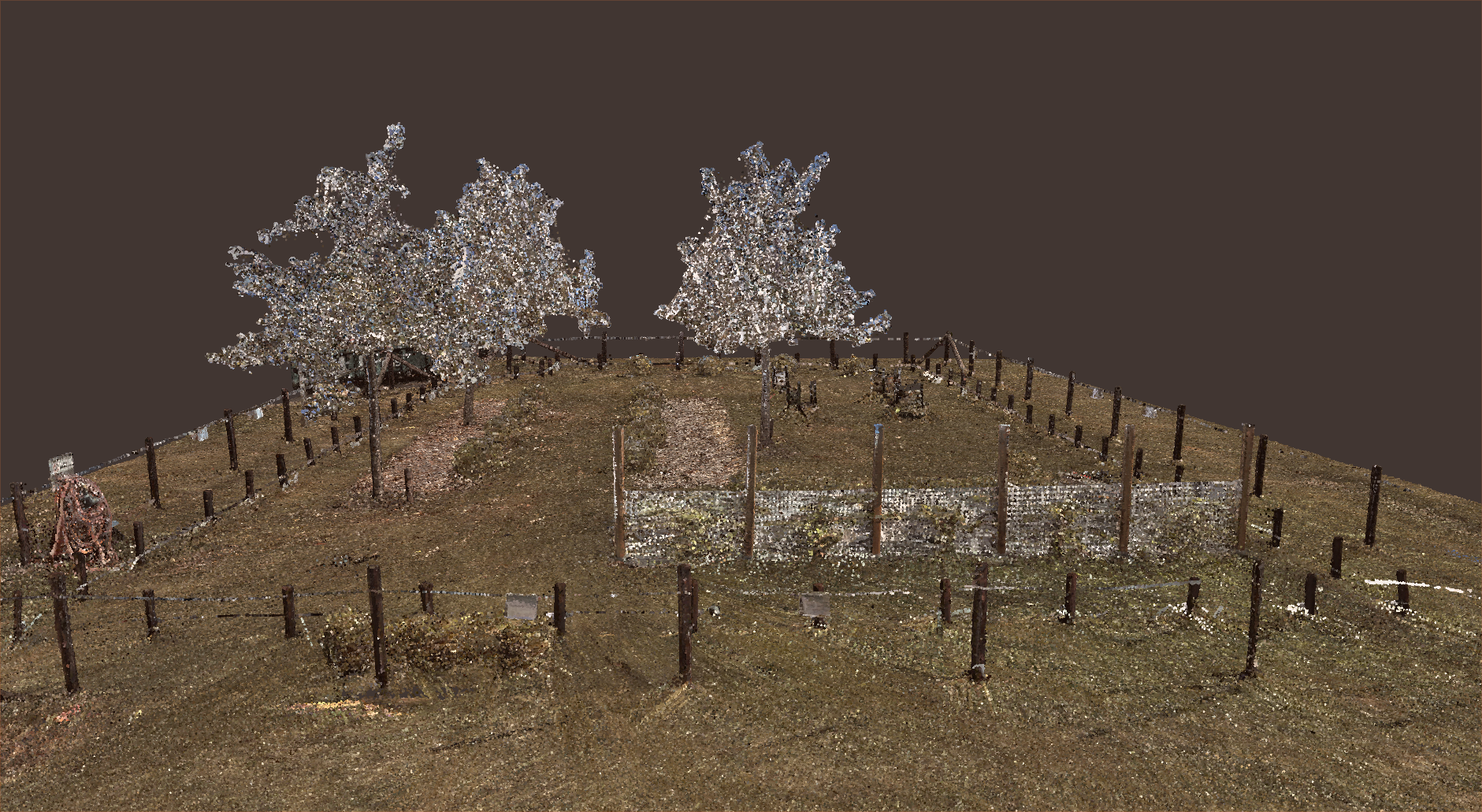}}
		\end{center}
	\end{multicols}
	
	\begin{multicols}{2}
		\begin{center}
			%\flushleft
			\subfloat[The point cloud from view 2 with colored height]{\includegraphics[ width=0.92\linewidth]{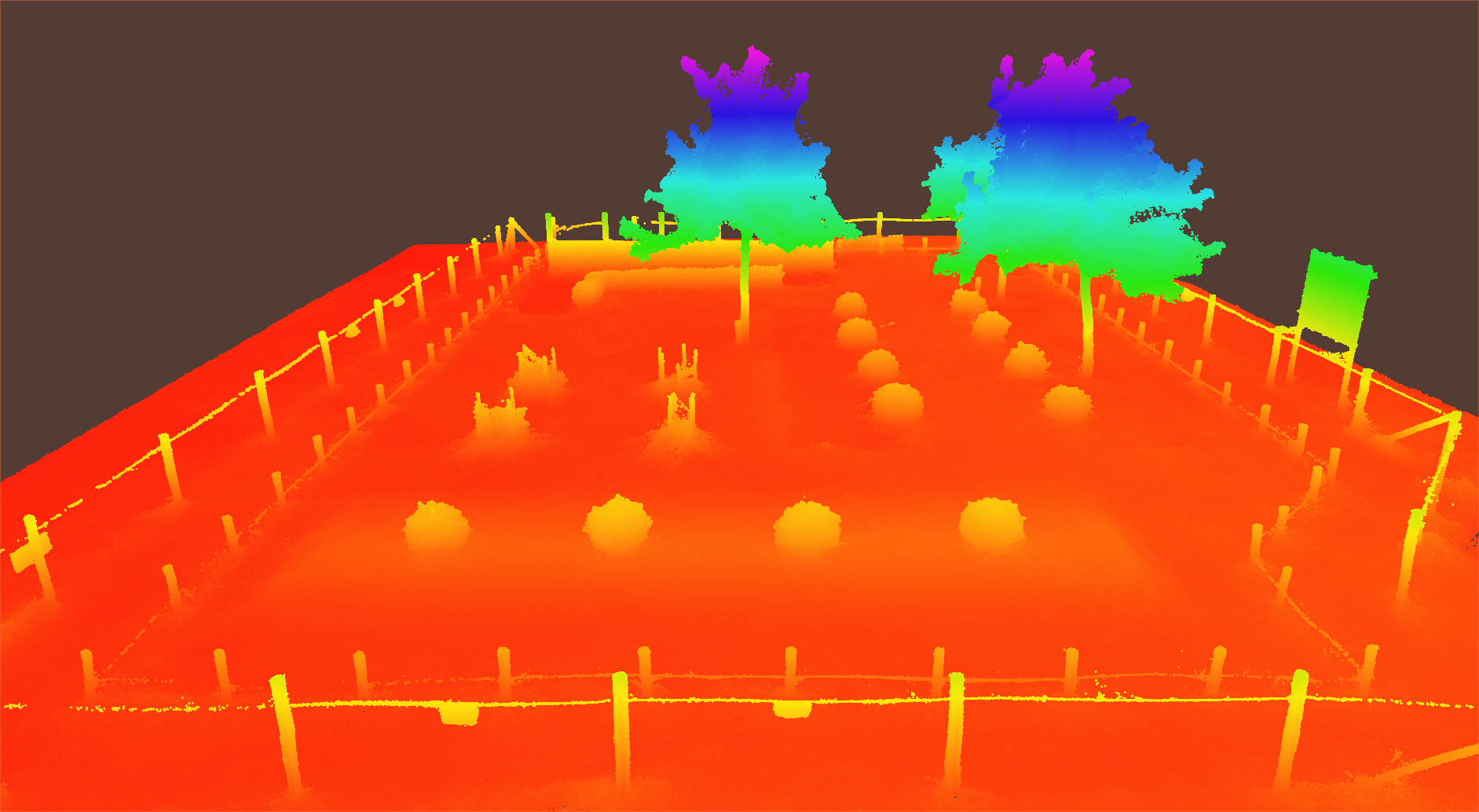}}
		\end{center}
		\begin{center}
			%\flushright
			\subfloat[The point cloud from view 2 with registered RGB data]{\includegraphics[ width=0.92\linewidth]{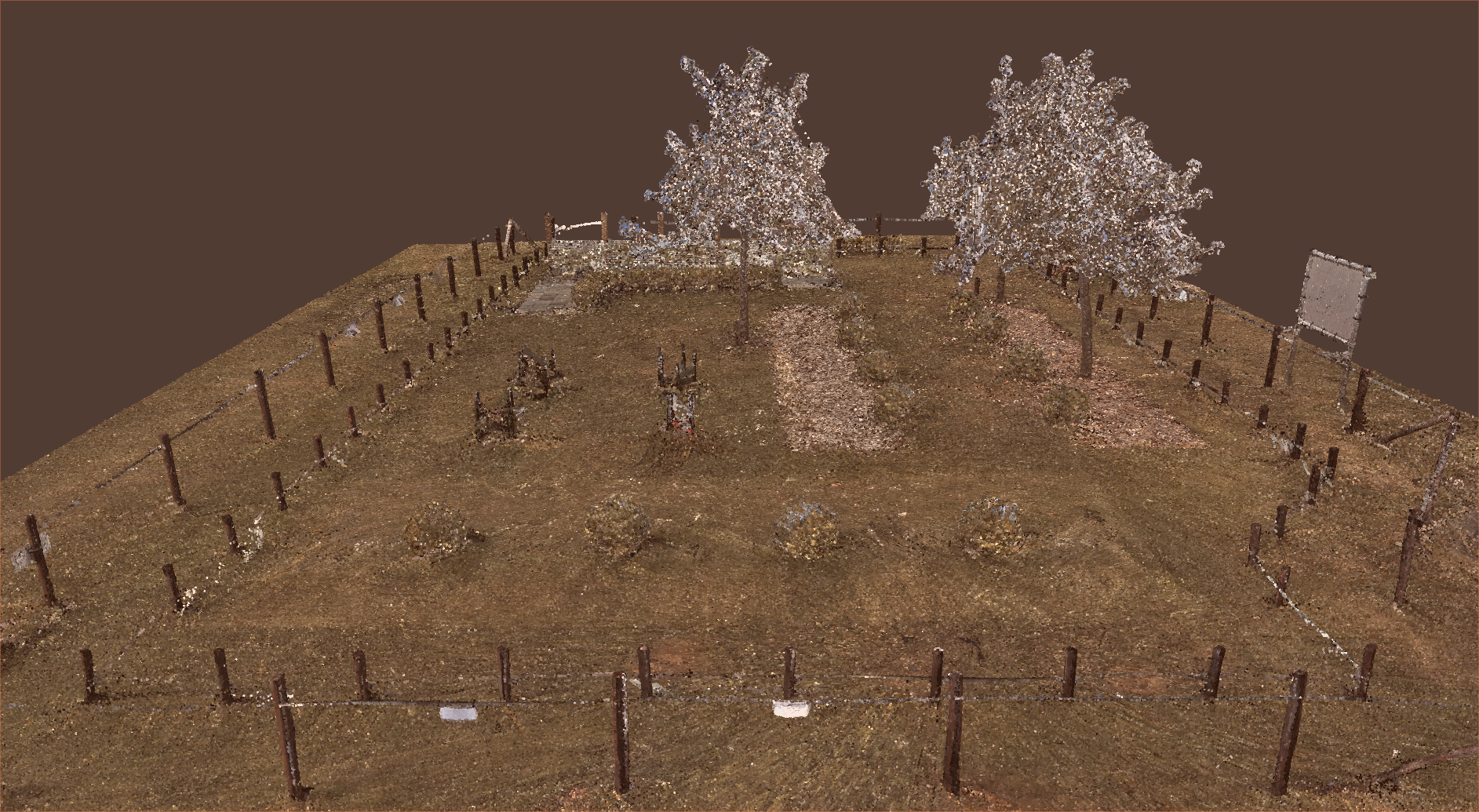}}
		\end{center}
	\end{multicols}

	\begin{multicols}{2}
		\begin{center}
			%\flushleft
			\subfloat[The point cloud from view 3 with colored height]{\includegraphics[ width=0.92\linewidth]{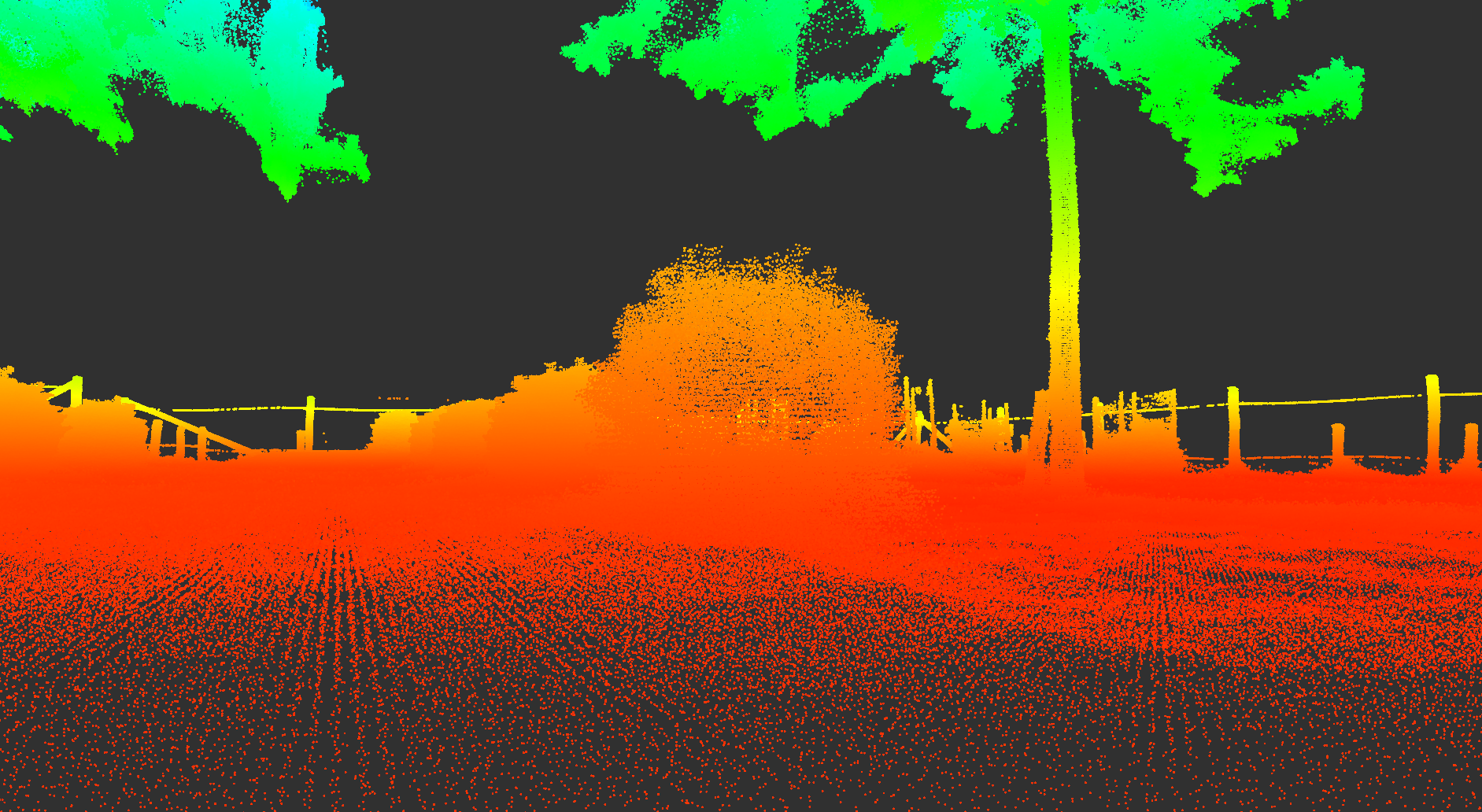}}
		\end{center}
		\begin{center}
			%\flushright
			\subfloat[The point cloud from view 3 with registered RGB data]{\includegraphics[ width=0.92\linewidth]{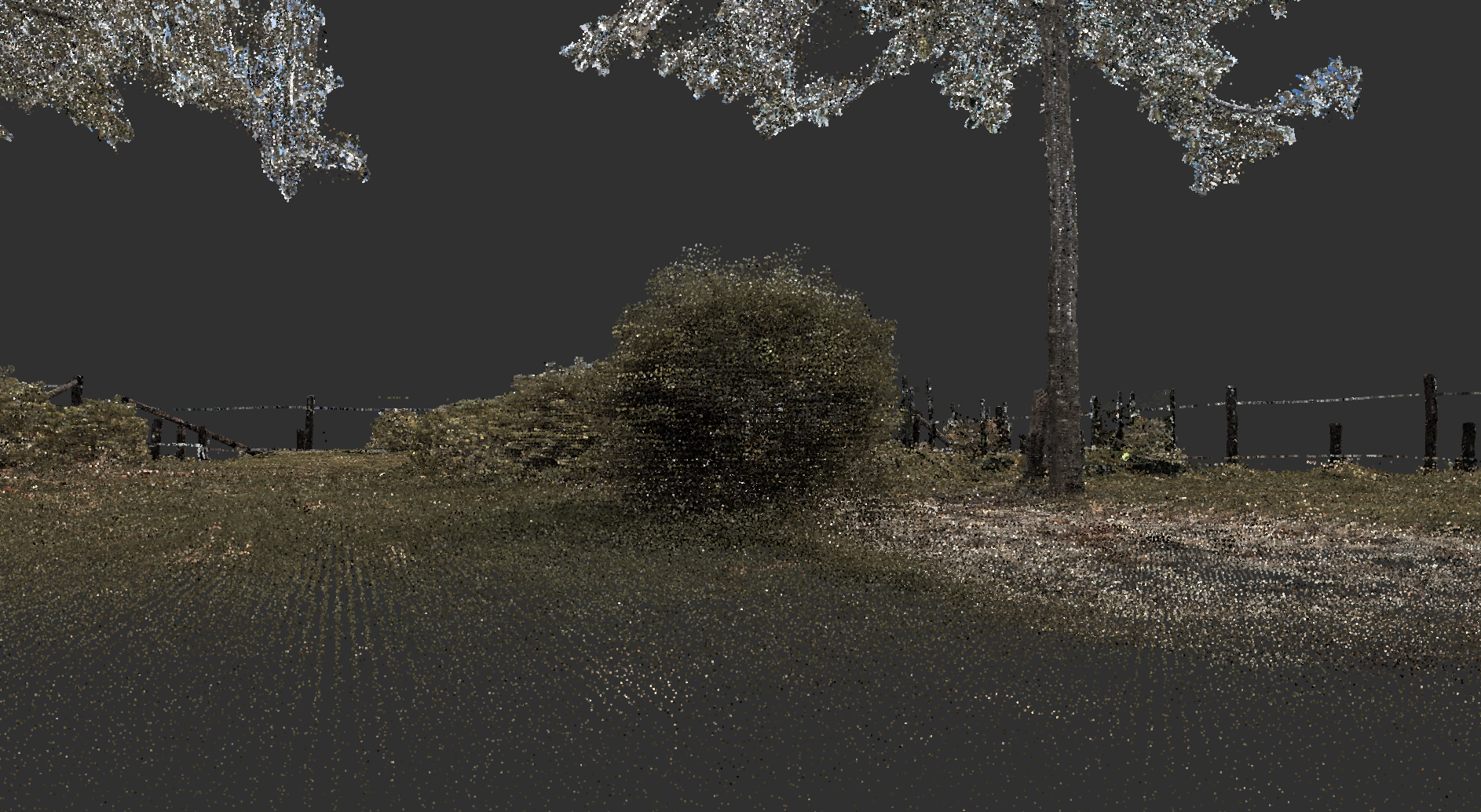}}
		\end{center}
	\end{multicols}

	\begin{multicols}{2}
		\begin{center}
			%\flushleft
			\subfloat[The point cloud from view 4 with colored height]{\includegraphics[ width=0.92\linewidth]{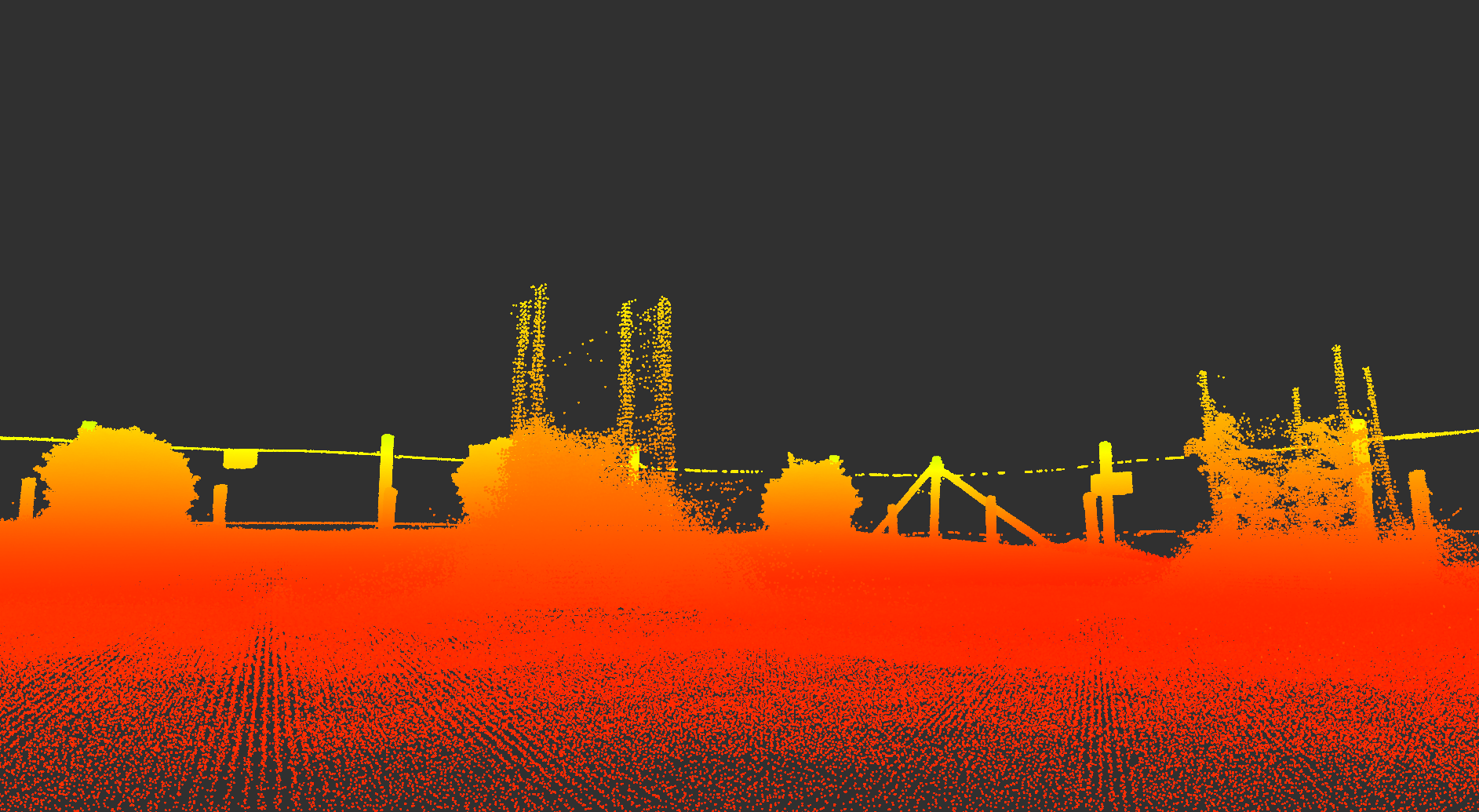}}
		\end{center}
		\begin{center}
			%\flushright
			\subfloat[The point cloud from view 4 with registered RGB data]{\includegraphics[ width=0.92\linewidth]{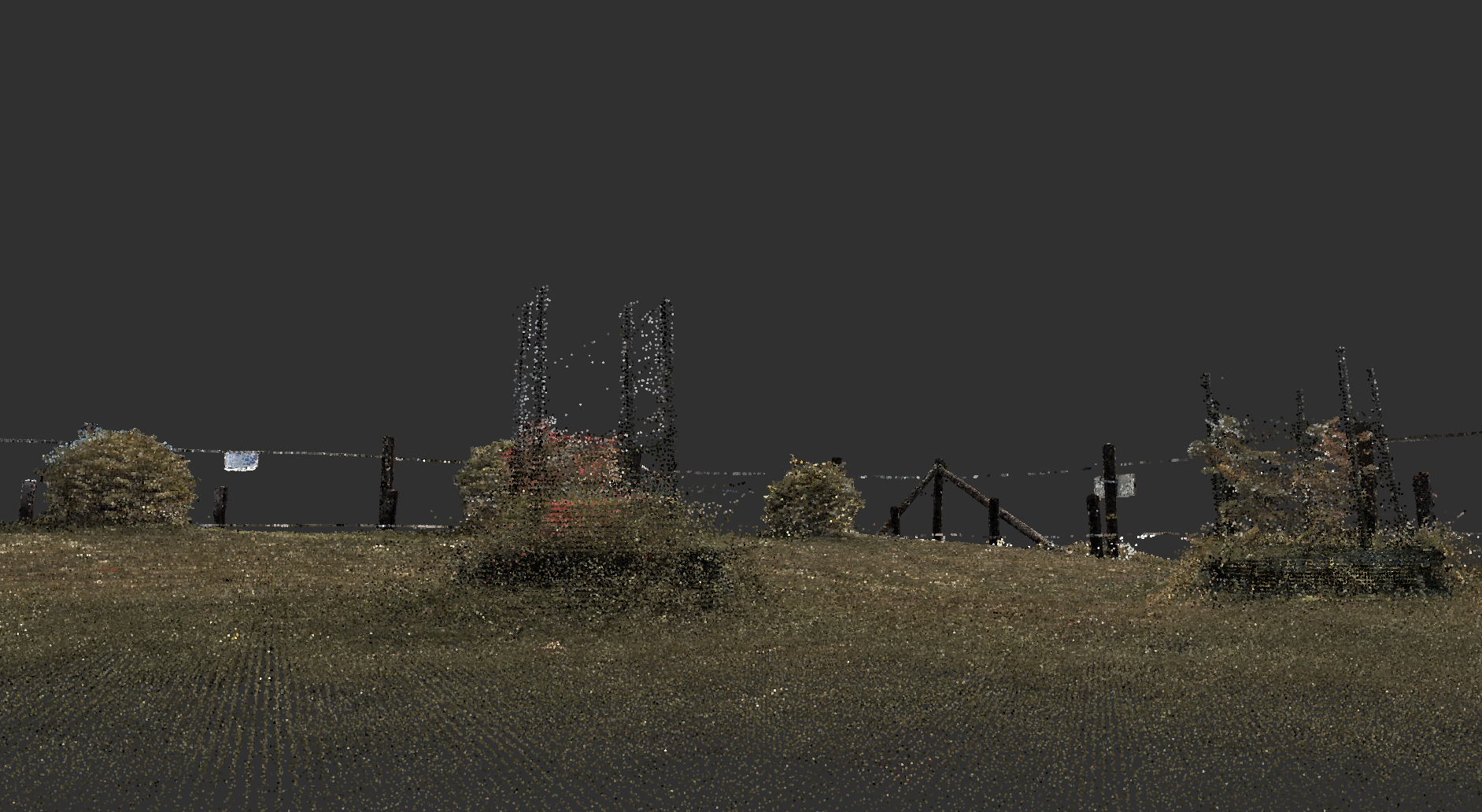}}
		\end{center}
	\end{multicols}

	\caption{\textcolor{black}{Figure (a) (c) (e) (g) show the point clouds with colored height. Figure (b) (d) (f) (h) show the point clouds with registered RGB data. All point clouds are from the Leica ScanStation P15. The images are from our partner Robert Bosch GmbH in the Trimbot2020 project.}}
	\label{fig:point-cloud-from-Leica-ScanStation-P15-appendix}
\end{figure*}

\textcolor{black}{The 3D point cloud of the whole garden was collected by a Leica ScanStation P15\footnote{Leica ScanStation P15 Datasheet URL: \url{https://github.com/Canpu999/Trimbot-Wageningen-SLAM-Dataset/blob/main/Leica_ScanStation_P15_datasheet.pdf}}. The Leica ScanStation P15 measures the distance by a laser scanner and incorporates the
corresponding RGB data from a color camera into the laser scanner's coordinate system. The laser scanner's working distance range varies from $[0.4m, 40m]$. The 3D position accuracy could be low to 3 mm at 40 meters and the linearity error is smaller than 1 mm. The angular accuracy is 8'' in the horizontal direction and 8'' in the vertical direction. Figure~\ref{fig:point-cloud-from-Leica-ScanStation-P15-appendix} shows the garden's point clouds from different views with RGB data and colored height (different colors represent different heights). As the distance measurement and the RGB measurement are conducted sequentially, both measurements are not exactly synchronized.
As a consequence, there may be wrong RGB values on moving objects' point clouds. 
%Take a moving leaf as an example. 
A moving leaf's point cloud (taken at one timestamp) will show the sky's color (taken at another timestamp) if the leaf moved at one timestamp (when the laser scanner was acquiring data) to another place at another timestamp (when the RGB sensor was acquiring data). 
Reflectance properties of glossy leaves can also result in color changes.
Thus, in Figure~\ref{fig:point-cloud-from-Leica-ScanStation-P15-appendix}, the tree's point clouds show a mixed color (green and sky-white). 
That is the reason why we did not include the point cloud with RGB data into the Trimbot Wageningen SLAM Dataset. 
%It's a disadvantage of this work, but we could not fix it because we could not control the wind outdoors or redesign this device.
}

\refstepcounter{section}
\section*{Appendix E. Additional Qualitative Results} \label{Appendix-more-qualitative-results}
% \appendix
\renewcommand\thefigure{\Alph{section}.\arabic{figure}} 
\renewcommand\thetable{\Alph{section}.\arabic{table}}

% \section{\textcolor{black}{Ablation Study}}
\setcounter{figure}{0}    
\setcounter{table}{0}

\subsection{Robustness to Environment Change} \label{Appendix-robustness-test}
\begin{figure*}
	\includegraphics[width=1\linewidth]{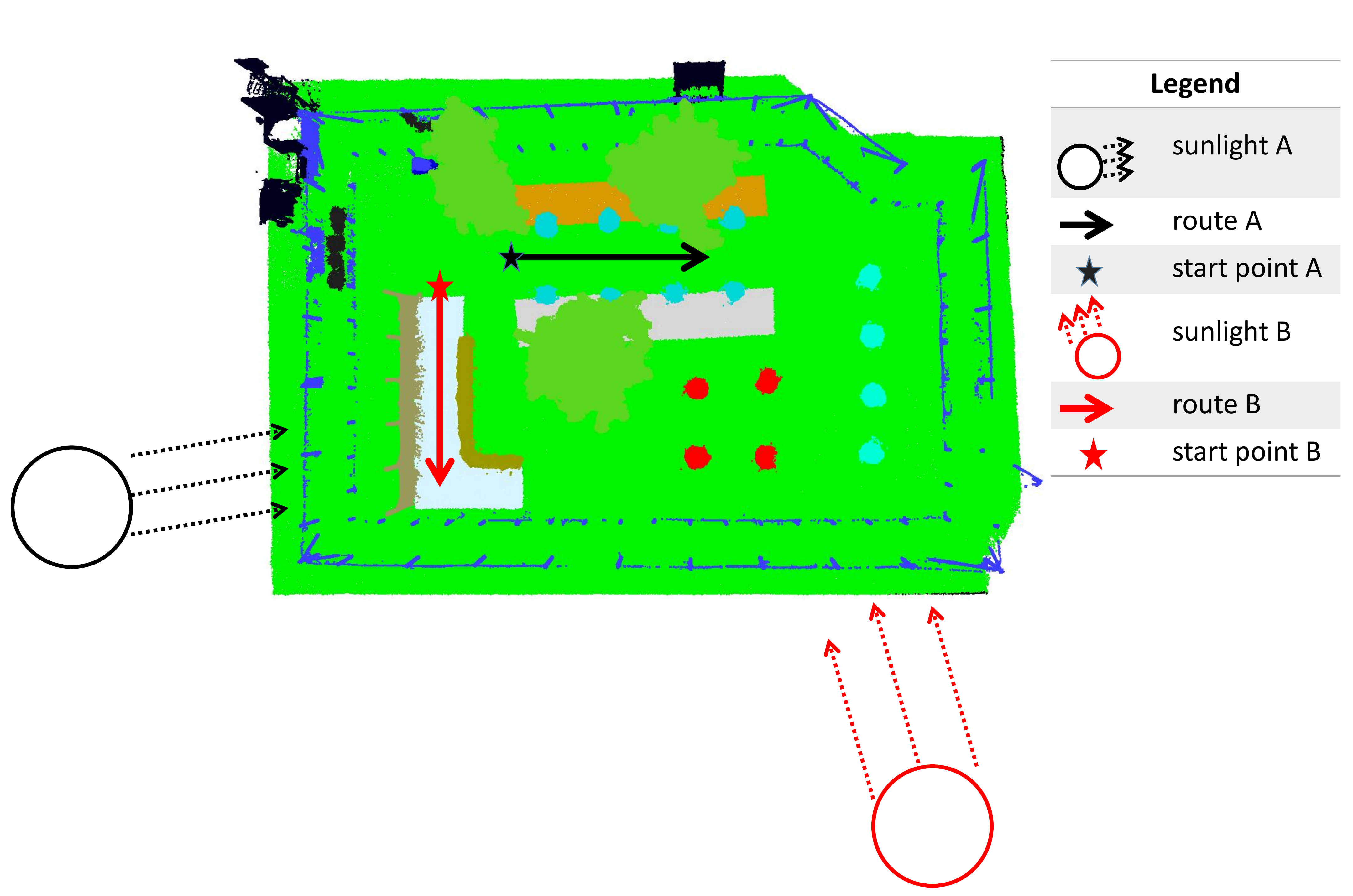}
	\caption{\textcolor{black}{The experiment setting in the robustness test}}
	\label{fig:sunshine-appendix}
\end{figure*}

\textcolor{black}{As sunlight and the scene appearance  outdoors will influence an image's appearance directly, thus we vary sunlight from different time periods and scene appearance from different places in the following experiment to demonstrate the robustness of the proposed framework. 
Figure~\ref{fig:sunshine-appendix} shows the experimental setting for the robustness test. 
In Scene 1, the robot drove straight along route A under sunlight A. 
In Scene 2, the robot drove straightly along route B (whose travelled distance is same with that of route A) under the same sunlight A. 
In Scene 3, the robot drove straightly along the route A, but under the different sunlight B. 
Since factors other than scene appearance and lighting should not influence the image appearance directly, we classify the other factors as Uncontrolled Random Factors. 
Table~\ref{table:robustness-appendix} gives the precise value of each factor, and the values of the uncontrolled random factors are from a weather website \footnote{\url{https://tcktcktck.org/netherlands/gelderland/wageningen}}.
}

\begin{figure*}
	
	\begin{multicols}{2}
		
		\begin{center}
			%\flushleft
			\subfloat[Scene 1]{\includegraphics[ width=1\linewidth]{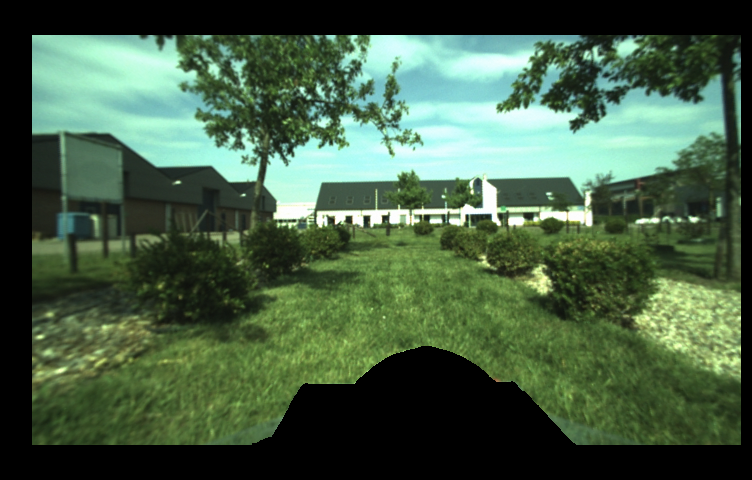}}
		\end{center}
		
		\begin{center}
			%\flushright
			\subfloat[Reconstructed Model 1]{\includegraphics[ width=0.82\linewidth]{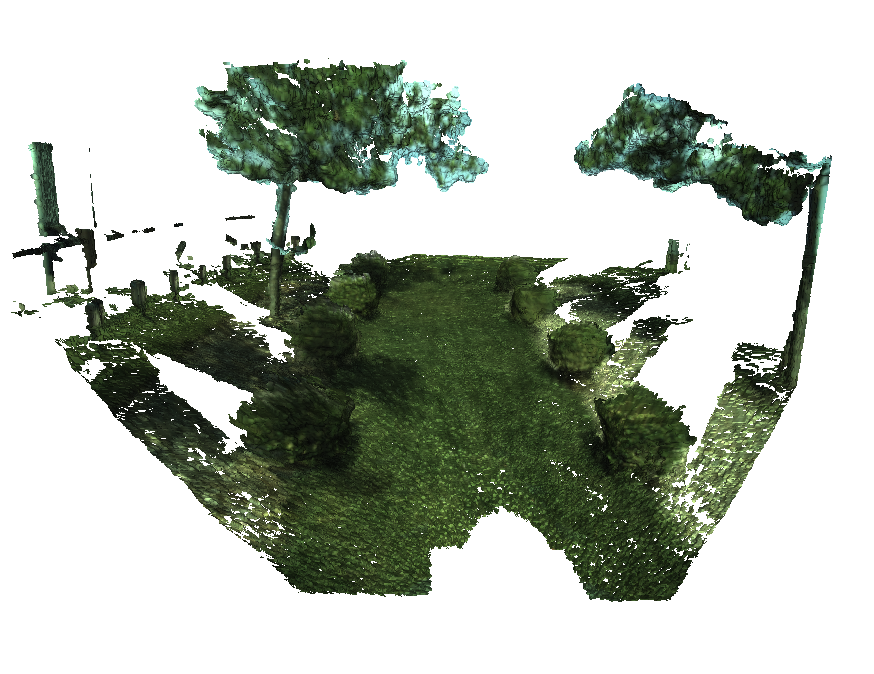}}
		\end{center}
	\end{multicols}
	
	\begin{multicols}{2}
		\begin{center}
			%\flushleft
			\subfloat[Scene 2]{\includegraphics[ width=1\linewidth]{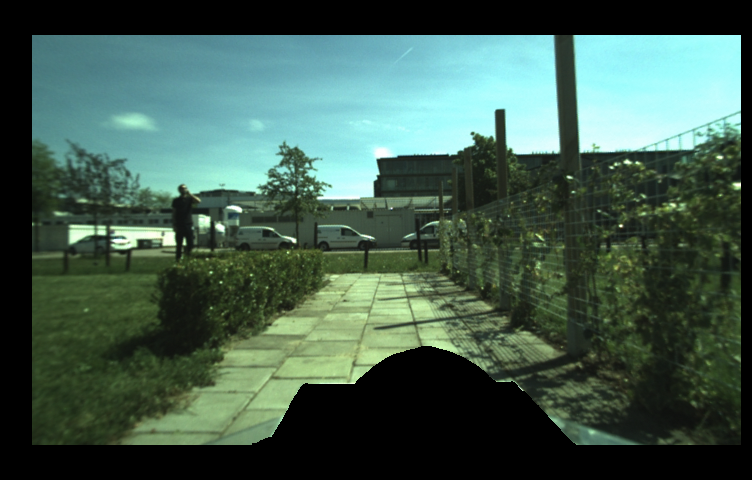}}
		\end{center}
		\begin{center}
			%\flushright
			\subfloat[Reconstructed Model 2]{\includegraphics[ width=0.8\linewidth]{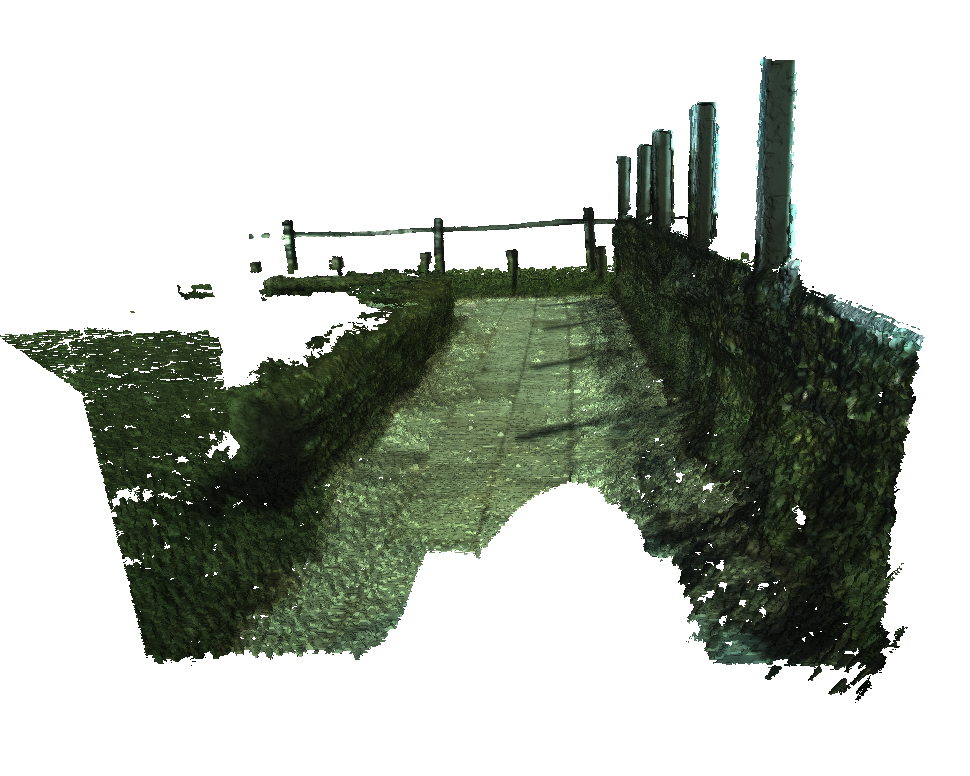}}
		\end{center}
	\end{multicols}

	\begin{multicols}{2}
	\begin{center}
		%\flushleft
		\subfloat[Scene 3]{\includegraphics[ width=1\linewidth]{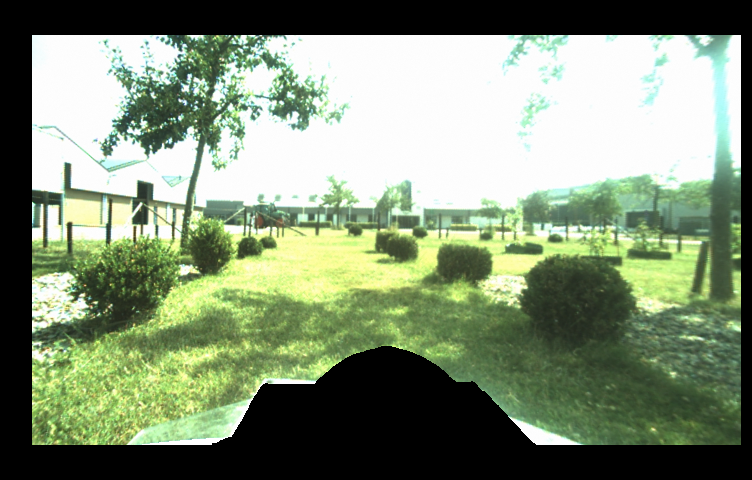}}
	\end{center}
	\begin{center}
		%\flushright
		\subfloat[Reconstructed Model 3]{\includegraphics[ width=0.75\linewidth]{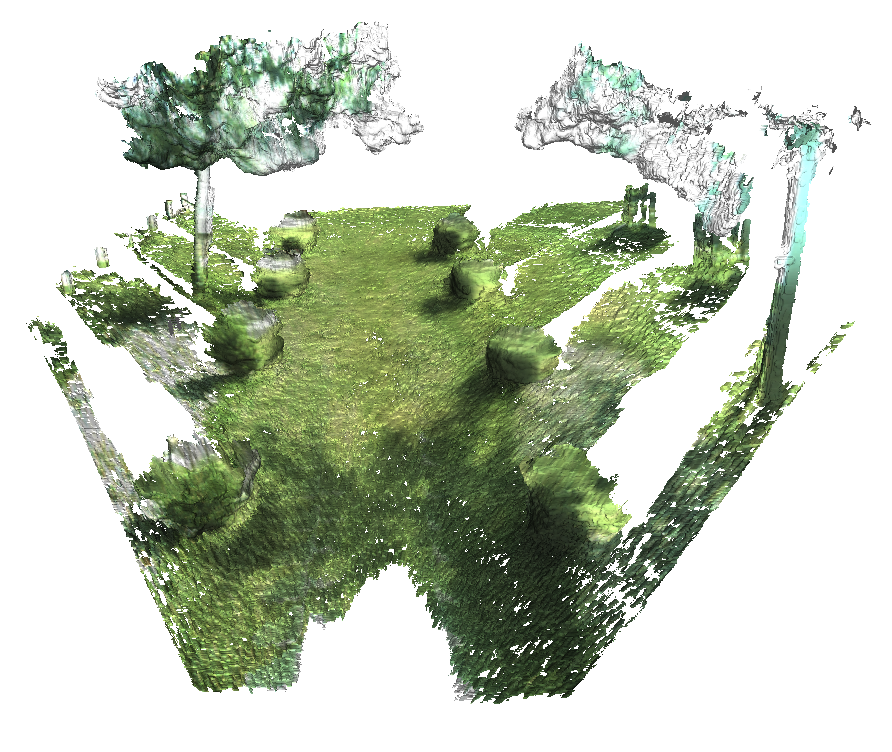}}
	\end{center}
\end{multicols}

	\caption{\textcolor{black}{Figure (a) (c) (e) shows the scenes and Figure (b) (d) (f) shows the reconstructed models. }}
	\label{fig:robustness-test-appendix}
\end{figure*}

\textcolor{black}{
Figure~\ref{fig:robustness-test-appendix} shows the three scenes and the related 3D reconstruction results in surface mesh format by the proposed framework.
As acquiring ground truth requires considerable  expense and effort by a big team, we only show the corresponding qualitative results. 
Compare the shadows, trees, posts and bushes in the example images (Figure~\ref{fig:robustness-test-appendix} a c e) with the corresponding shapes in the reconstructed 3D meshes (Figure~\ref{fig:robustness-test-appendix} b d f).
The high quality and believable shapes, even with different lighting and viewpoint, demonstrate the high quality of the 3D reconstruction by our proposed framework.
%From all the experiments in this paper, we could see the proposed framework is robust to different scene's appearance, the sunlight's angle, and illumination intensity. 
Further robustness tests against other factors (e.g. temperature, wind speed, long operating time, humidity, foggy or snowy weather) are future work.} 

\begin{table*}[H]
	\centering
	\caption{\textcolor{black}{Scene Definition with Different Condition Setting}.} \label{table:robustness-appendix}
	\begin{tabular}{>{\color{black}}c>{\color{black}}c>{\color{black}}c>{\color{black}}c>{\color{black}}c>{\color{black}}c>{\color{black}}c>{\color{black}}c>{\color{black}}c>{\color{black}}c}
		\toprule & \multicolumn{2}{>{\color{black}}c}{\bf Controlled Variable } & \multicolumn{7}{>{\color{black}}c}{\bf Uncontrolled Random Factors } \\ \midrule
		\textbf{Name} & \textbf{route} & \textbf{sunlight} & \textbf{time} & \textbf{date} & \textbf{temperature }& \textbf{wind speed} & \textbf{humidity}  & \textbf{dew point} & \textbf{pressure} \\
		\midrule 
		\textbf{Scene 1} & A & A & $\sim 15:30$ & 2017-05-17 & $22.5^\circ C$ & 12.7 km/h & 76\% & $15.7 ^\circ C$ & 995.6Mb \\
		%\hline 
		\textbf{Scene 2} & B & A & $\sim 15:30$ & 2017-05-17 & $22.5^\circ C$ & 12.7 km/h & 76\% & $15.7 ^\circ C$ & 995.6Mb \\
		%\hline 
		\textbf{Scene 3} & A & B & $\sim 11:00$ & 2018-06-27 & $20.5^\circ C$ & 12.7 km/h & 75\%  & $12.7 ^\circ C$ & 1003.4Mb \\
		\bottomrule 
	\end{tabular} 
\end{table*}

\subsection{Fair Comparison with More Open-source Frameworks} \label{Appendix-fair-comparison}

\textcolor{black}{Orbslam3~\citep{orbslam3} and Open3D~\citep{Open3d} needed hints in Section~\ref{section: comparison_pose_fusion_module} to make them work. 
In this subsection, we will compare all the frameworks when using only one stereo vision camera (which consists of the image sensor Cam0 and Cam1) without any hints provided to any framework. 
We use the initial parameter setting of Open3D for a test (denoted by Open3D-initial). 
For Orbslam3, we test their RGBD SLAM algorithm (denoted by Orbslam3-rgbd), their stereo SLAM algorithm (denoted by Orbslam3-stereo), and their monocular SLAM algorithm (denoted by Orbslam3-monocular). 
For the parameter setting of each SLAM algorithm from Orbslam3, please refer to the video at \url{https://youtu.be/4luADtHNbuA}. 
In this video\footnote{\textcolor{black}{There are many false extracted features near the edge of the robot in the above video. We refined the data input of Orbslam3 by masking more areas near the robot and tuned the parameters to maximize Orbslam3's performance. The modified video (URL: \url{https://youtu.be/9TaayXl4mJ4}) shows the corresponding performance.}}, we also show the performance of each algorithm from Orbslam3 on our Trimbot Wageningen SLAM dataset when having a random test\footnote{Due to the stochastic property of the algorithm RANSAC~\citep{Ransac1981} in the framework Orbslam3, the results from Orbslam3 are different each time.}. 
}

\begin{figure*}
	\centering
	\includegraphics[width=1\linewidth]{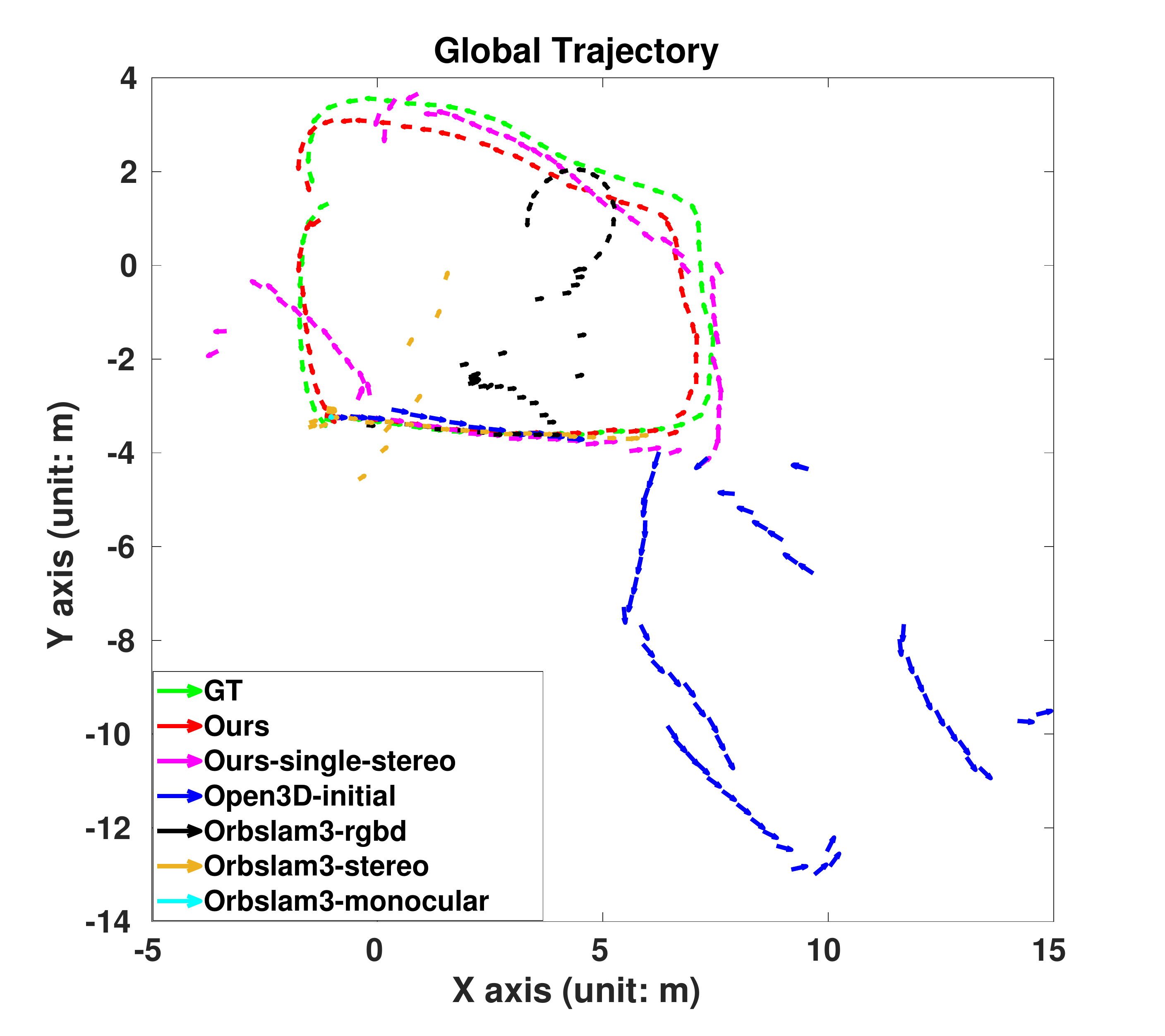}
	\caption{\textcolor{black}{The global pose trajectory from each algorithm}}
	\label{fig:trajectory-fair-comparison-appendix}
\end{figure*}

\textcolor{black}{
Figure~\ref{fig:trajectory-fair-comparison-appendix} shows the global pose trajectory of each framework. 
The results of Orbslam3 are from the best trials among tens of random trials. 
Method "Ours" is the method  proposed in this paper with panoramic stereo images as input. 
Method "Ours-single-stereo" denotes a method which is derived from the "Ours" method, but with the input from only one stereo camera. 
Figure~\ref{fig:trajectory-fair-comparison-appendix} shows that all of the algorithms sucessfully estimate poses initially, but eventually fail, except "Ours" when traversing the whole dataset. 
"Ours-single-stereo" seems to work better compared with the external counterparts, although it  still fails at the latter part of the global pose trajectory. 
The trajectory of Orbslam3-monocular (cyan trajectory) has only one dot to show at the starting point, near the coordinate (0 m, -1 m ), because Orbslam3-monocular almost lost tracking every frame.
Given the bad performance of each external algorithm, it is meaningless to calculate the corresponding quantitative results. 
It is also the reason why we needed to provide "some extra help" to the external counterparts to get results to compare with in Section~\ref{section: comparison_pose_fusion_module}.}

\subsection{Comparison with Software `ContextCapture'}  \label{Appendix-contextcapture}
\begin{figure*}
	
	\begin{multicols}{2}
		
		\begin{center}
			%\flushleft
			\subfloat[One View]{\includegraphics[ width=1\linewidth]{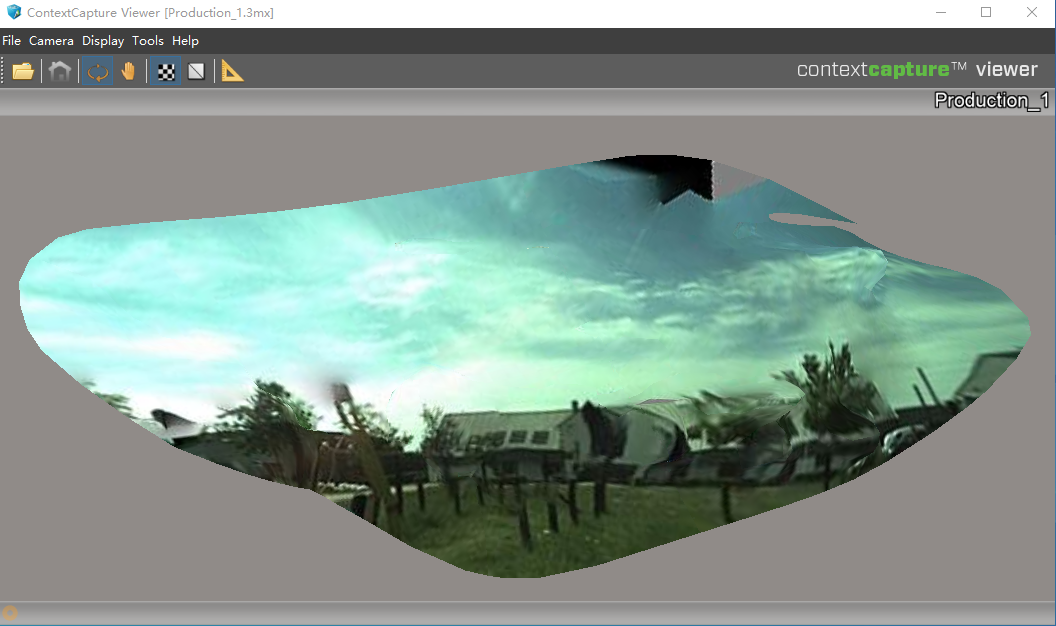}}
		\end{center}
		
		\begin{center}
			%\flushright
			\subfloat[Another View]{\includegraphics[ width=1\linewidth]{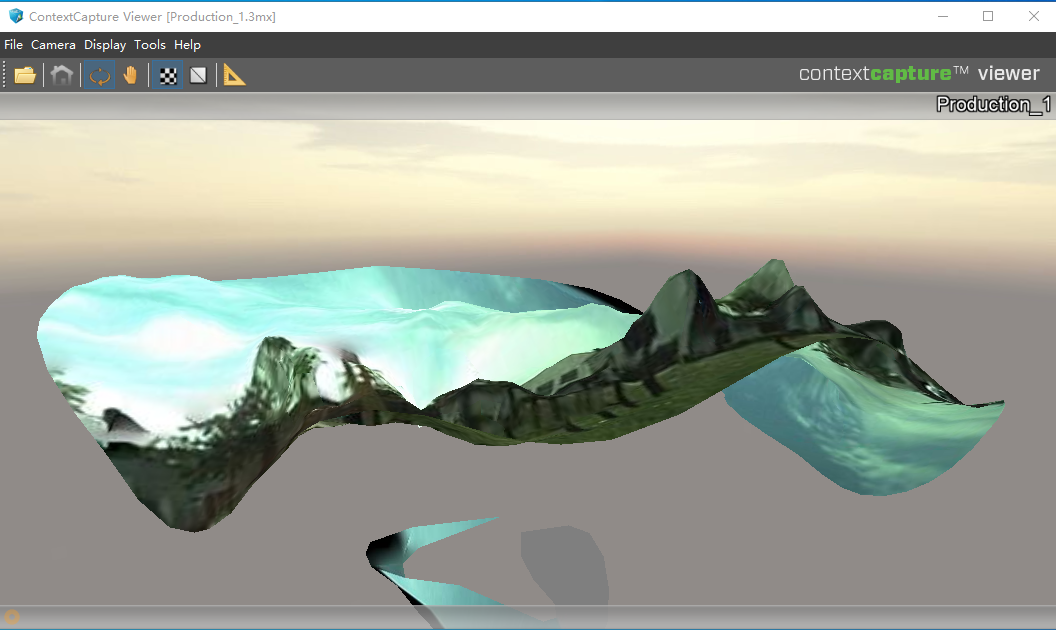}}
		\end{center}
	\end{multicols}

	\caption{\textcolor{black}{Figures (a) and (b) show the reconstructed models from the professional 3D reconstruction Software `ContextCapture'.} }
	\label{fig:contextcapture-appendix}
\end{figure*}

\textcolor{black}{Above, we compared our proposed framework with the latest popular frameworks Orbslam3~\citep{orbslam3} and Open3D~\citep{Open3d}. 
The experimental results have shown our framework's performance superiority over the two open-source frameworks. 
Here, we compare the proposed framework with the commercial software called "ContextCapture Center" from Bentley\footnote{\url{https://www.bentley.com/software/contextcapture/}}, which is known for its image-based 3D reconstruction and aerial photogrammetry. 
Because the commercial software ContextCapture's 3D reconstruction framework does not support  $360^\circ$ image-based 3D reconstruction, we could only input the RGB images from our image sensor Cam0 in our Wageningen SLAM Dataset. 
Figure~\ref{fig:contextcapture-appendix} shows the reconstructed meshes from two views. 
From the strange reconstruction results, we could easily see that the ContextCapture reconstruction system failed at our task. 
The reason is that the real robot's fast movement and large transformation between frames when navigating in the real outdoor garden lead ContextCapture to lose tracking (please read the acquisition report\footnote{ContextCapture's acquisition report: \url{https://drive.google.com/file/d/1k_oUyolomh2LmJv3ATyrUnxA5sc--bxE/view?usp=sharing}} from ContextCapture). The quality report\footnote{ContextCapture's quality report: \url{https://drive.google.com/file/d/1Q5DqU8XOqrO10RWt9dWxitlhvU81vRjQ/view?usp=sharing}} from the software ContextCapture reveals more reconstruction details about its failure.} 

\color{black}
%\newpage
%% Loading bibliography style file
%\bibliographystyle{model1-num-names}
\bibliographystyle{cas-model2-names}

%% Specify the bibliography file. Default is thesis.bib.
\bibliography{visual-pipeline-refs}

\end{document}